%% file: main.tex
\PassOptionsToPackage{table}{xcolor}
\documentclass[10pt,twocolumn,letterpaper]{article}

\usepackage[pagenumbers]{iccv} % To force page numbers, e.g. for an arXiv version

\input{preamble}

\definecolor{iccvblue}{rgb}{0.21,0.49,0.74}
\usepackage[pagebackref,breaklinks,colorlinks,allcolors=iccvblue]{hyperref}

\usepackage[capitalize]{cleveref}
\crefname{section}{Sec.}{Secs.}
\Crefname{section}{Section}{Sections}
\crefname{table}{Tab.}{Tabs.}
\Crefname{table}{Table}{Tables}
\crefname{observation}{Observation}{Observations}
\crefname{conclusion}{Conclusion}{Conclusions}

\title{What Makes for Text to 360-degree Panorama Generation with Stable Diffusion?}

\author{
Jinhong Ni$^1$\thanks{Work partially done at The University of Hong Kong.}
\qquad
Chang-Bin Zhang$^2$
\qquad
Qiang Zhang$^{3,4}$
\qquad
Jing Zhang$^1$ \\
$^1$Australian National University 
\quad
$^2$The University of Hong Kong 
\\
$^3$Beijing Innovation Center of Humanoid Robotics \\
$^4$Hong Kong University of Science and Technology (Guangzhou) \\
\texttt{\small \{jinhong.ni,jing.zhang\}@anu.edu.au}
\quad
\texttt{\small cbzhang@connect.hku.hk}
\quad
\texttt{\small jony.zhang@x-humanoid.com}
}

\begin{document}

\maketitle

%%%%%%%%% ABSTRACT
\begin{abstract}
Recent prosperity of text-to-image diffusion models, \eg Stable Diffusion, has stimulated research to adapt them to 360-degree panorama generation. Prior work has demonstrated the feasibility of using conventional low-rank adaptation techniques on pre-trained diffusion models to generate panoramic images. However, the substantial domain gap between perspective and panoramic images raises questions about the underlying mechanisms enabling this empirical success. We hypothesize and examine that the trainable counterparts exhibit distinct behaviors when fine-tuned on panoramic data, and such an adaptation conceals some intrinsic mechanism to leverage the prior knowledge within the pre-trained diffusion models. Our analysis reveals the following: 1) the query and key matrices in the attention modules are responsible for common information that can be shared between the panoramic and perspective domains, thus are less relevant to panorama generation; and 2) the value and output weight matrices specialize in adapting pre-trained knowledge to the panoramic domain, playing a more critical role during fine-tuning for panorama generation. We empirically verify these insights by introducing a simple framework called UniPano, with the objective of establishing an elegant baseline for future research. UniPano not only outperforms existing methods but also significantly reduces memory usage and training time compared to prior dual-branch approaches, making it scalable for end-to-end panorama generation with higher resolution. The code will be released\footnote{\url{https://github.com/jinhong-ni/UniPano}}.
\end{abstract}

\begin{figure*}[ht]
    \centering
    \setlength{\tabcolsep}{1pt}
    \def\arraystretch{0.42}
    \begin{tabular}{cccc}
        \multicolumn{4}{c}{\emph{``Amidst the ruins of an ancient civilization, deciphering hieroglyphics that tell the story of a lost world.''}}\\
        % \multirow{2}{*}[32mm]{\rotatebox[origin=c]{90}{\emph{a living room with bookshelves}}} 
        \multicolumn{4}{c}{\includegraphics[width=0.983\textwidth]{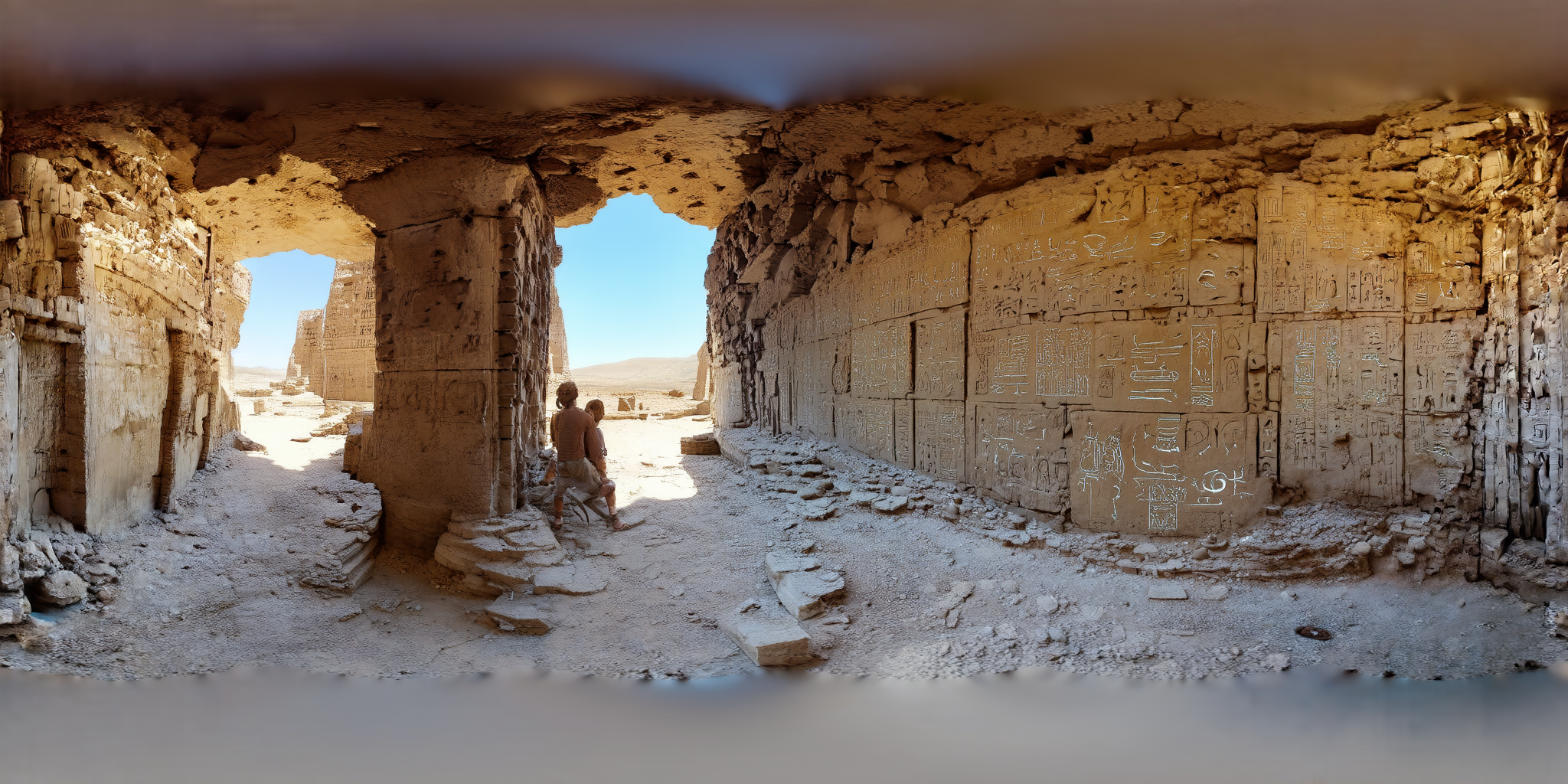}} \\
        \multicolumn{2}{c}{\emph{``a living room with a fireplace.''}} & \multicolumn{2}{c}{\emph{``a home with pool and patio.''}} \\
        \multicolumn{2}{c}{\includegraphics[width=0.49\textwidth]{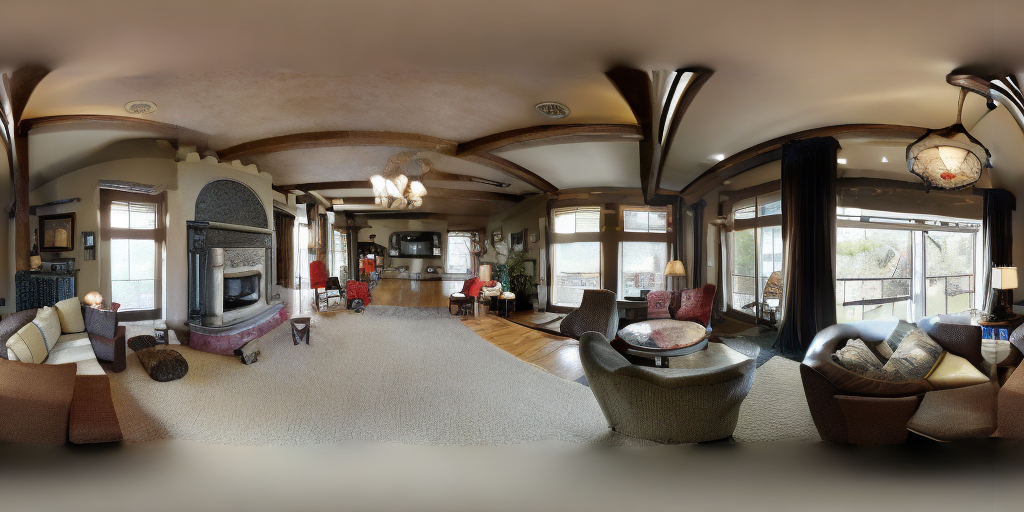}} &
        \multicolumn{2}{c}{\includegraphics[width=0.49\textwidth]{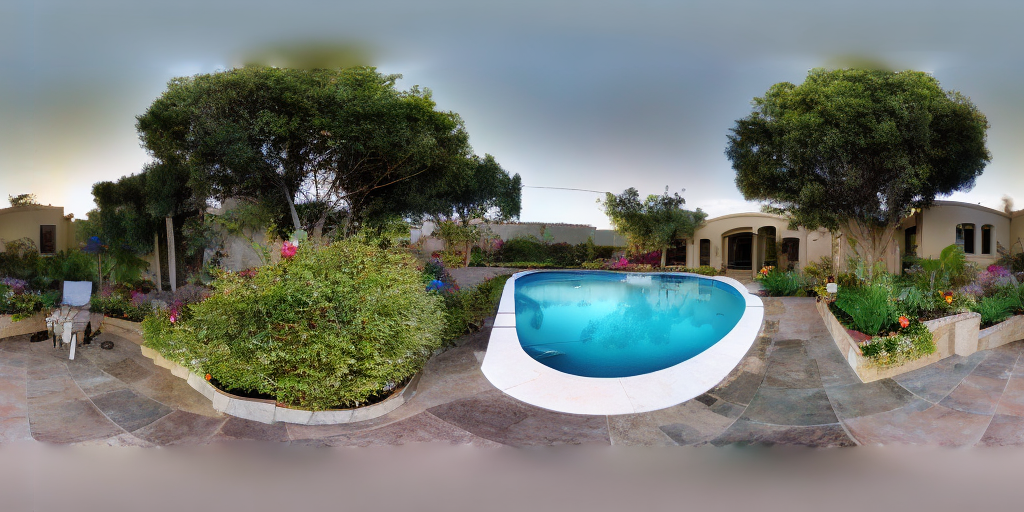}}
    \end{tabular}
    \caption{Our UniPano can synthesize realistic 360-degree panoramic images by fine-tuning Stable Diffusion. (Top) $1024\times 2048$ panoramic images generated by UniPano. (Bottom) $512\times 1024$ panoramic images generated by UniPano.}
    \label{fig:pre-res}
\end{figure*}

%%%%%%%%% BODY TEXT
\section{Introduction}
Creating 360-degree panoramic images has gained substantial attention due to its significant potential~\citep{yang2023dreamspace,li20244k4dgen}. Despite the considerable advancement in text-to-image synthesis recently~\citep{ldm,dalle,dalle2}, generating panoramas from text prompts remains challenging from the following aspect. Panoramic images encompass the entire surrounding view with a 360-degree horizontal and 180-degree vertical field of view, typically represented using equirectangular projection geometry. This results in distinctive features such as a $2:1$ aspect ratio and spherical distortion, setting them apart from standard square perspective images. On top of this, due to the high cost of capturing panoramic images in practice, the panoramic datasets are often relatively scarce, \eg Matterport3D~\citep{Matterport3D} contains 10,800 panoramic images. The lack of data complicates the training of generative models, as conventional perspective diffusion models~\citep{ldm} generally require billions of text-image pairs for training~\citep{laion}.

To mitigate data scarcity, the typical strategy is to fine-tune pre-trained diffusion models for downstream applications~\citep{dreambooth,controlnet,marigold}.
However, as stated in~\cite{cubediff}, the fundamental structural differences between panoramic and perspective images intuitively suggest that the embedded perspective knowledge within the pre-trained diffusion models may not be readily transferable. Aligning with this intuition, prior work~\citep{panogen,mvdiff,cubediff} has proposed generating multiple perspective images according to predefined camera poses and stitching them into a panorama. 
Contrary to the aforementioned intuition, another line of work~\citep{panfusion} has demonstrated that fine-tuning pre-trained diffusion models on limited panoramic data using conventional low-rank adaptation (LoRA)~\citep{lora} still yields effective text-to-panorama generation results. This empirical success suggests the presence of some intrinsic mechanism that enables LoRA to effectively leverage prior knowledge from the pre-trained perspective diffusion models, thereby circumventing the structural differences.
This motivates us to explore the following question: \emph{What exactly makes for fine-tuning Stable Diffusion for text-to-panorama generation?}

\begin{figure}[!t]
    \centering
    \definecolor{mygreen}{RGB}{0,210,0}
    \includegraphics[width=0.88\linewidth]{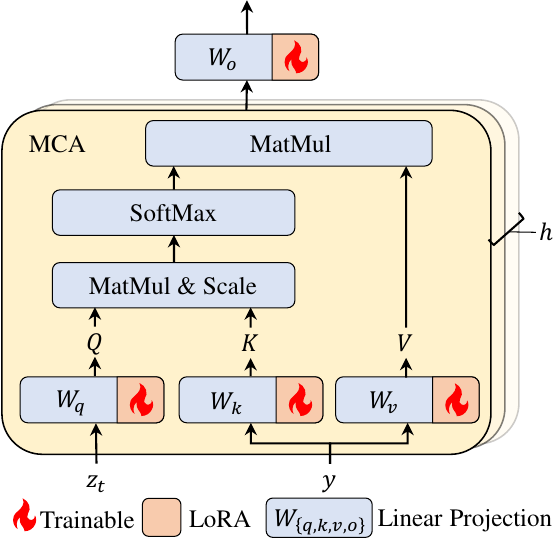}
    \caption{\textbf{Illustration of fine-tuning cross-attention blocks within diffusion models with low-rank adaptation (LoRA)}. MCA, MatMul, $h$ denotes the multi-head cross attention, matrix multiplication, and the number of attention heads respectively.}
    \label{fig:attn-lora}
\end{figure}

We base our analysis on the LoRA fine-tuning paradigm to study the behaviors and ideally functionalities of all trainable counterparts, particularly their impact on panorama generation, with the ultimate goal of elucidating the mechanism that thrives in adapting perspective diffusion models for panorama generation. Our launching point is to isolate the trainable components within LoRA fine-tuning (\ie, $W_{\{q,k,v,o\}}$, \cf \cref{fig:attn-lora}) and examine their relevance for learning panoramic structures. 
Subsequently, we identify the underlying behaviors of each trainable component when they are tuned jointly. 
We draw two major empirical findings (\cf \cref{sec:analysis} for details):
\begin{itemize}
    \item $W_{\{q,k\}}$ in the attention blocks fail to learn the panoramic structures when they are trained in isolation, whereas $W_{\{v,o\}}$ both succeed in capturing such information.
    \item When $W_{\{q,k,v,o\}}$ are jointly trained, $W_{\{v,o\}}$ are responsible for learning panoramic-specific information (\ie, equirectangular structure), whereas $W_{\{q,k\}}$ learn shared knowledge across panoramic and perspective domains that are irrelevant to the panoramic structure.
\end{itemize}
Our analysis reveals the following: After fine-tuning with panoramic images, we discover that the query and key within the cross-attention blocks capture less panoramic-specific information, namely, they function to `preserve' or `enhance' the pre-trained perspective knowledge; In contrast, the value and output weight matrices are responsible for adapting such perspective information into the panoramic domain.
Based on the analysis, we believe that fine-tuning the query and key matrices is less relevant to panorama generation, whereas the representational capability of the value and output matrices should be emphasized.
This yields our straightforward yet efficacious uni-branch solution, dubbed \emph{UniPano}, targeting to serve as a simple baseline to foster future research. UniPano achieves state-of-the-art results on $512\times 1024$ text-to-panorama generation while requiring notably less memory and training time compared to the current SoTA~\citep{panfusion}. Thanks to this computational efficiency, UniPano can be scaled to generate panoramic images with even higher resolution in an end-to-end manner, as shown in~\cref{fig:pre-res}.

\section{Related Work}

\paragraph{Diffusion Models.}
The recent breakthrough in diffusion models~\citep{ddim,ddpm,scoresde} has accelerated the inference process~\citep{dpmv3,unipc,rf,cm} and significantly boosted the generation quality~\citep{sdxl,sd3}. The prosperity of large-scale pre-trained diffusion models~\citep{dalle,dalle2,ldm} has prompted various applications, including text-to-3D generation~\citep{dreamfusion,chen2024text}, personalized customization~\citep{textualinv,dreambooth}, image inpainting~\citep{lugmayr2022repaint,xie2023smartbrush}, depth estimation~\citep{marigold}, perception~\citep{xu2023open,diffusiondet}, \etc. The core of most of these works is to exploit pre-trained text-to-image diffusion models as a priori thus circumventing the data scarcity which is common in downstream applications. Such an adaptation is usually achieved by parameter-efficient fine-tuning techniques such as low-rank adaptation (LoRA)~\citep{lora}, or via distillation~\citep{meng2023on}. This paper targets the former approach and attempts to demystify what makes for panorama generation by fine-tuning pre-trained diffusion models with LoRA.

\paragraph{Panorama Generation.}
Existing works can be roughly divided into two categories, namely panorama outpainting and text-to-panorama generation. The former approach~\citep{faed,panodiffusion,akimoto2022diverse,wang2022stylelight,lu2024autoregressive,cubediff} aims to complete a panoramic image based on a partial input image, exemplified by CubeDiff~\citep{cubediff} which proposes to jointly generate six faces of cubemap for panorama generation. 
Aligning with the prosperity of text-conditioned generation as in the perspective domain, text-to-panorama generation~\citep{yu2023long,panfusion,stitchdiffusion,panogen,mvdiff,multidiffusion,panofree,diffpano} has gained attention recently. Among these works, \cite{panogen,mvdiff,multidiffusion,yu2023long} generate a sequence of consistent perspective images and stitch them into a panorama. A separate branch of works fine-tunes the pre-trained text-to-image diffusion models to generate an equirectangular panoramic image in an end-to-end manner~\citep{panfusion,stitchdiffusion,diffpano}. 
StitchDiffusion~\citep{stitchdiffusion} fine-tunes pre-trained diffusion models with techniques ensuring panoramic continuity.
DiffPano~\citep{diffpano} includes multi-view panoramic awareness into the framework.
PanoFree~\citep{panofree} stands apart from the aforementioned methods by employing a tuning-free approach to generate panoramas.
PanFusion~\citep{panfusion} introduces a dual-branch framework by simultaneously generating perspective and panoramic images and ensuring consistency through a cross-branch attention mechanism.
Of particular importance to our study, \cite{panfusion} shows that LoRA fine-tuning on Stable Diffusion reports reasonable performance,
and our work aims to investigate and elucidate the underlying factors contributing to this empirical success.
As a side product of our analysis, we present an efficient and effective uni-branch panorama generation framework, serving as a baseline method for future research.

\begin{figure*}[t]
    \centering
    \definecolor{mygreen}{RGB}{0, 210, 0}
    \setlength{\tabcolsep}{1pt}
    \begin{tabular}{cc}
        \includegraphics[width=0.48\linewidth]{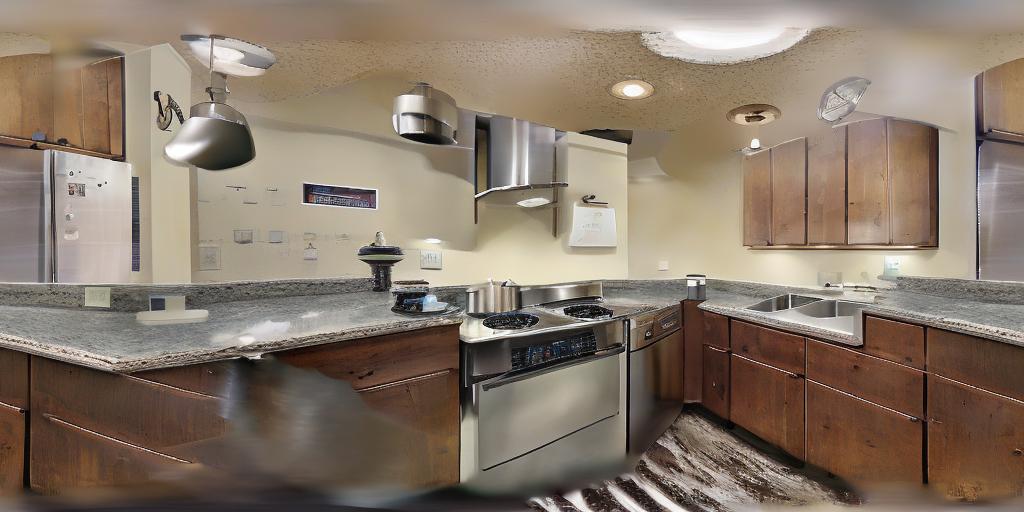} & \includegraphics[width=0.48\linewidth]{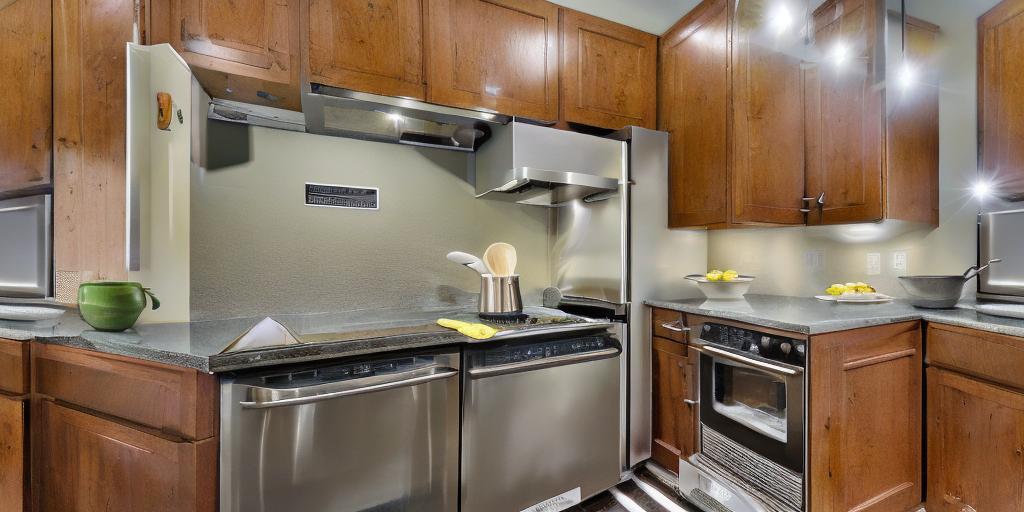} \\
        (a) Training $W_q$ in isolation \textcolor{red}{\XSolidBrush} & (b) Training $W_k$ in isolation \textcolor{red}{\XSolidBrush} \\
        \includegraphics[width=0.48\linewidth]{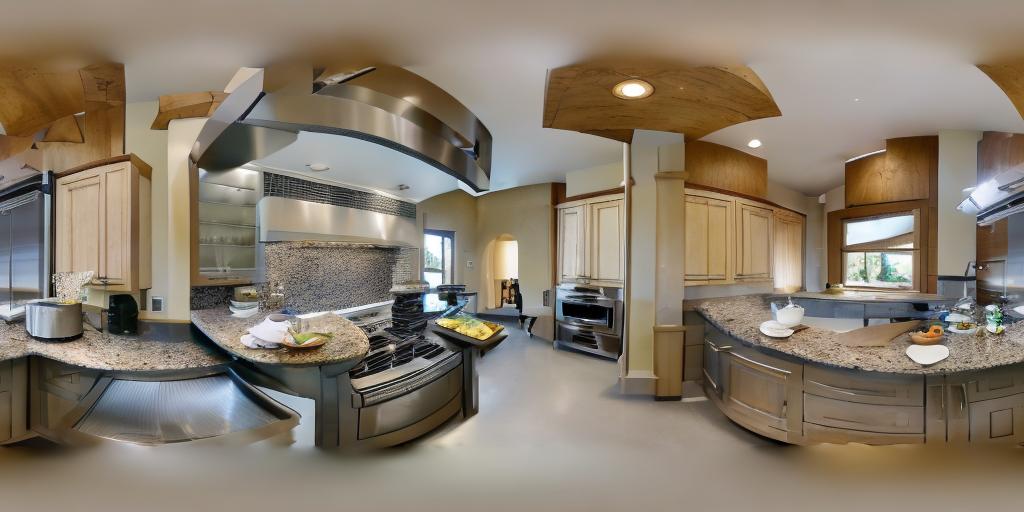} & \includegraphics[width=0.48\linewidth]{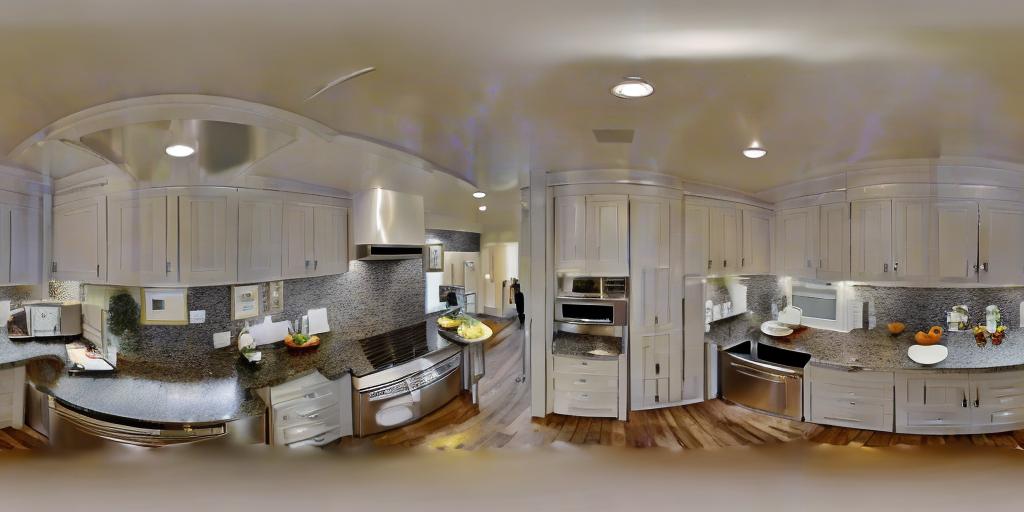} \\
        (c) Training $W_v$ in isolation \textcolor{mygreen}{\CheckmarkBold} & (d) Training $W_o$ in isolation \textcolor{mygreen}{\CheckmarkBold}
    \end{tabular}
    \caption{
    \textbf{Qualitative comparison for training $W_{\{q,k,v,o\}}$ in isolation separately}. Training $W_q$ or $W_k$ in isolation fails to capture the spherical structure, as in (a) and (b); whereas training $W_v$ or $W_o$ in isolation successfully captures the spherical distortion of the panoramic images, as in (c) and (d). 
    All visualizations are generated with the text prompt ``a kitchen with stainless steel appliances''.
    }
    \label{fig:qkvo-iso}
\end{figure*}

\section{What Makes for Panorama Generation?}

\subsection{Preliminary}\label{sec:preliminary}
Diffusion models~\citep{ddpm,ddim,scoresde} generate images by iteratively transforming the noise sampled from the prior distribution into the target data distribution, where each sampling step involves predicting the noise from the input noisy image. We defer details on diffusion models to the supplementary materials.
Of particular relevance to our study, the conditioning in diffusion models is often accomplished by cross-attention. Formally, given an input latent $z_t$ and the corresponding condition $y$, the attention computes:
\begin{equation*}
    \mrm{MHCA}(z_t,y)W_o,
\end{equation*}
where $\mrm{MHCA}$ denotes the multi-head cross-attention, and $W_o$ represents the output weight; for notational simplicity, we write $\mrm{MHCA}$ in the single-head form
\begin{equation*}
    \mrm{softmax}\rb{\frac{QK^\top}{\sqrt{d_h}}}V,
\end{equation*}
with head channel dimension $d_h$, $Q=z_t W_q$, $K=y W_k$, and $V=y W_v$, where $W_{\{q,k,v,o\}}$\footnote{Throughout the paper, we denote $W_{\{q,k,v,o\}}$ as the set of trainable weights $\{W_q,W_k,W_v,W_o\}$.} are the set of trainable weights. The community has shown that training these attention modules within text-to-image diffusion models using parameter efficient fine-tuning techniques (\eg LoRA~\citep{lora}) suffices to surrogate fine-tuning the entire models, exemplified by~\cite{textualinv,dreambooth}. We illustrate fine-tuning diffusion models with LoRA in~\cref{fig:attn-lora}.

\subsection{Motivation and Insights}\label{sec:analysis}

360$^\circ$ panoramic images capture a complete spherical view of the surroundings, involving a full 360-degree horizontal field of view and a vertical field of view of 180 degrees, typically stored in equirectangular format. These unique characteristics in view structures make legitimate panoramic images fundamentally different from the perspective ones, which at first glance, implies that perspective-related knowledge within the pre-trained diffusion models may not be immediately relevant. Contrary to this intuition, previous works~\citep[]{panfusion} have demonstrated the feasibility of directly fine-tuning pre-trained perspective diffusion models (\eg, Stable Diffusion) with LoRA (\cf, \cref{sec:preliminary}) for text-to-panorama generation, given a relatively scarce set of panoramic data. Evinced by this empirical success, we speculate that such an adaptation to the panoramic domain has to conceal some intrinsic mechanism to leverage perspective knowledge within the pre-trained diffusion models. 

\input{tabs/qkvo_exp}

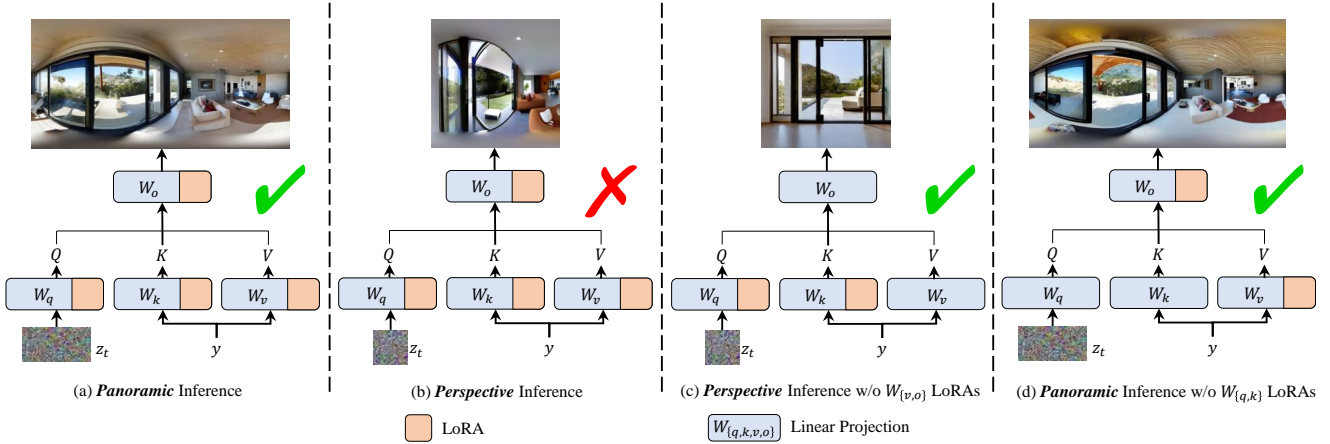
\begin{figure*}[!t]
    \centering
    \definecolor{mygreen}{RGB}{0,210,0}
    \begin{overpic}[width=\textwidth]{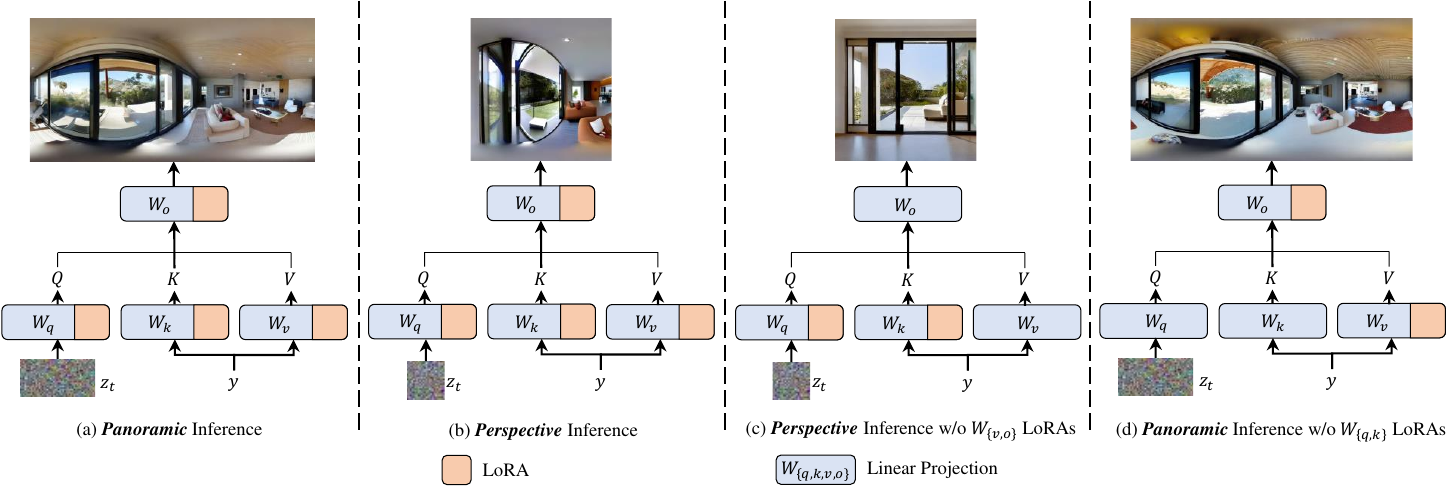}
    \put(19,17){\huge \textcolor{mygreen}{\CheckmarkBold}}
    \put(44,17){\huge \textcolor{red}{\XSolidBrush}}
    \put(70,17){\huge \textcolor{mygreen}{\CheckmarkBold}}
    \put(95,17){\huge \textcolor{mygreen}{\CheckmarkBold}}
    \end{overpic}
    \caption{\textbf{Illustration of roles of $W_{\{q,k,v,o\}}$ when jointly fine-tuned}. We first fine-tune $W_{\{q,k,v,o\}}$ jointly with LoRA on panoramic data, then optionally deactivate some LoRAs for inference with different purposes. (a) the panoramic image generated by fine-tuning $W_{\{q,k,v,o\}}$ jointly with LoRA, where we simplify the LoRA architecture (\cf \cref{fig:attn-lora}); (b) diffusion models with panorama fine-tuned LoRAs can only generate distorted, panoramic-like images, are thus no longer capable of generating perspective images; (c) by excluding $W_v$ and $W_o$ LoRAs, diffusion models fine-tuned on panoramic data recover the ability to generate perspective images; (d) excluding $W_q$ and $W_k$ LoRAs does not affect the model's ability to generate panoramic images.}
    \label{fig:qkvo-joint}
\end{figure*}

To reveal such a mechanism and elucidate what makes such perspective-based diffusion adaptation succeed in panorama generation, we start by decomposing the trainable components and training them in isolation to identify the behaviors -- and ideally the functionalities -- of each counterpart. Specifically, we fine-tune $W_{\{q,k,v,o\}}$ (\cf, \cref{fig:attn-lora}) separately with LoRA for panorama generation, and showcase the comparison both qualitatively in \cref{fig:qkvo-iso} and quantitatively in \cref{tab:exp-qkvo}. 
We highlight the following.

\begin{observation}\label{obs:qkvo-iso}
    As shown in~\cref{fig:qkvo-iso} (a) and (b), it is evident that training $W_q$ or $W_k$ in isolation notably fails to capture the spherical structure within panoramic images, while both $W_v$ and $W_o$ are capable of achieving such goals, as in~\cref{fig:qkvo-iso} (c) and (d). The quantitative results in~\cref{tab:exp-qkvo} also align with the qualitative observation, as training $W_v$ or $W_o$ in isolation leads to considerably better FAED and FID than $W_q$ or $W_k$. 
\end{observation}

\begin{conclusion}\label{conclusion:qkvo-iso}
    The four trainable components $W_{\{q,k,v,o\}}$ within the cross-attention modules exhibit varying abilities to learn the spherical structures for successful adaptation to panorama generation. In particular, $W_q$ and $W_k$ fail to capture such distortions even when fitted purely on panoramic data, whereas $W_v$ and $W_o$ both successfully adapt the pre-trained diffusion model to panorama generation when trained in isolation.
\end{conclusion}

\begin{remark}\label{remark:qkvo-iso}
\cref{conclusion:qkvo-iso} suggests that $W_{\{q,k\}}$ have limited capability to be adapted for panorama generation. As such, excluding them during fine-tuning shall not impact the model's capability to learn the panoramic structure.
\end{remark}

Our subsequent step is to investigate how each component contributes to adaptation when fine-tuned collectively. Knowing that $W_q$ and $W_k$ are unable to capture the spherical structure characteristic of panoramic images, we hypothesize that their role is limited to learning common knowledge across both the perspective and panoramic domains. To validate this argument, we design the following experiment: 
we first fine-tune $W_{\{q,k,v,o\}}$ with LoRA jointly on panoramic data, then optionally deactivate some LoRAs during inference to examine their relevance to panorama generation.
The illustrations are provided in~\cref{fig:qkvo-joint}, from which we draw the observation and conclusion below.

\begin{observation}
    After fine-tuning the diffusion models with panoramic data (\cref{fig:qkvo-joint} (a)), the model when attempted to generate perspective images with all LoRAs applied can only generate panoramic-like images, as shown in \cref{fig:qkvo-joint} (b). If the LoRAs associated with $W_{\{v,o\}}$ are excluded during inference, the models with only the remaining $W_{\{q,k\}}$ LoRAs successfully recover the ability to generate valid perspective images, evidenced in \cref{fig:qkvo-joint} (c). On top of this, excluding $W_{\{q,k\}}$ LoRAs does not affect the model's ability to generate panoramic images, as shown in~\cref{fig:qkvo-joint} (d).
\end{observation}

\begin{conclusion}\label{conclusion:vo-pano}
    When all weights $W_{\{q,k,v,o\}}$ are trained jointly, LoRAs associated with $W_q$ and $W_k$ learn shared knowledge across panoramic and perspective domains that are irrelevant to panoramic structures, whereas LoRAs associated with $W_v$ and $W_o$ are responsible for learning the spherical structures of the panoramic images.
\end{conclusion}

\begin{remark}\label{remark:qkvo-joint}
\cref{conclusion:vo-pano} differentiates the trainable components based on their roles. In particular, $W_{\{q,k\}}$ even when trained on panoramic images do not apply any spherical distortion to the generated images, such a finding also aligns with \cref{conclusion:qkvo-iso} as they also struggle to learn spherical information when trained isolatedly. Contrarily, as $W_{\{v,o\}}$ are responsible for learning the panoramic structure, their capacity needs to be emphasized more during fine-tuning.
\end{remark}

To this end, we have explicated what exactly makes for panorama generation with pre-trained diffusion models. Specifically, after fine-tuning with panoramic images, we find that $W_q$ and $W_k$ within the cross-attention modules do not capture the spherical distortion within the panorama at all. They act as if they were tuned using perspective images, and their roles are likely to be `preserving' or `refining' the pre-trained perspective knowledge. In stark contrast, $W_v$ and $W_o$ are responsible for adapting the information -- captured both in $W_{\{q,k\}}$ and within the pre-trained model itself -- into the panoramic domain. Their roles are thus learning the spherical structure of the panorama, which is much more instrumental to the task.
% \end{remark}

\subsection{UniPano}
As a byproduct of our insights, we present a memory-efficient uni-branch fine-tuning framework for adapting pre-trained text-to-image diffusion models to panorama generation, dubbed \emph{UniPano}. 
The core idea of UniPano is based on \cref{remark:qkvo-iso,remark:qkvo-joint}: in essence, it freezes $W_{\{q,k\}}$ as they are associated with non-parametric-specific information and emphasizes $W_{\{v,o\}}$ as they are critical to capture equirectangular distortion within the panoramas. 
We highlight that the purpose of UniPano is to empirically verify our insights in~\cref{sec:analysis}, and with the aim to provide a simple yet effective baseline for future research.

\input{tabs/ablation_wo_arch}

\input{tabs/main_exp}

\paragraph{Design Choices of $W_o$.}
We propose to increase the representational capacity of the corresponding modules to enhance model's learning ability for the spherical structure of panoramic images. 
We differentiate $W_v$ and $W_o$ with the intuition that $W_v$ is associated with head-wise projection while $W_o$ interacts directly with the entire set of representations.
Due to this reason, although both components are capable of learning the spherical structure, we choose to increase the capacity of $W_o$ because of its directness. We adopt and compare several common strategies to enhance the representational capability: 
\begin{itemize}
    \item \emph{Larger LoRA ranks} (LoRA $r=8$)~\citep{lora} is the most straightforward way to increase capability. This approach simply doubles the LoRA ranks of $W_o$ from $4$ to $8$.
    \item \emph{Local Window Attention} (LA)~\citep{localattn,nat} constrains the receptive field of the attention operation to the neighboring pixels. We insert such an attention block before each $W_o$ LoRA.
    \item \emph{Deformable Attention} (DA)~\citep{deformableattn} introduces a learnable offset and computes the attention based on the sampled features. We attempt to insert a deformable attention block prior to each $W_o$ LoRA.
    \item \emph{Squeeze and Excitation} (SE)~\citep{senet} adaptively recalibrates channel-wise feature responses to strengthen the representational power. We similarly insert a SE block before each $W_o$ LoRA.
    \item \emph{Mixture of Experts} (MoE)~\citep{ditmoe,mole} involves computing weighted sum over several expert networks, where the weights are learned via a routing network. We replace each $W_o$ LoRA with a MoE module. We adopt the same MoE architecture as~\citep{ditmoe}, except that each expert network is a LoRA, similar to the design in~\citep{mole}. We apply the same auxiliary loss as~\citep{ditmoe} for routing load balancing.
\end{itemize}

The comparison is detailed in~\cref{tab:arch-ab}, where we also provide a reference to the panorama branch only (Pano Only) baseline. To ensure a fair comparison with this baseline, we also fine-tune $W_{\{q,k\}}$ LoRAs in all compared strategies. 
We simply opt to use MoE to enhance the capacity of $W_o$ because of its superiority on FAED, while we highlight that such a choice may not be optimal and encourage future work to investigate further.

\begin{figure*}
    \centering
    \setlength{\tabcolsep}{1pt}
    \def\arraystretch{0.5}
    \begin{tabular}{lcccccccc}
        \multirow{2}{*}[32mm]{\rotatebox[origin=c]{90}{\emph{``a house with a pool and patio''}}} & \multicolumn{4}{c}{\includegraphics[width=0.48\linewidth]{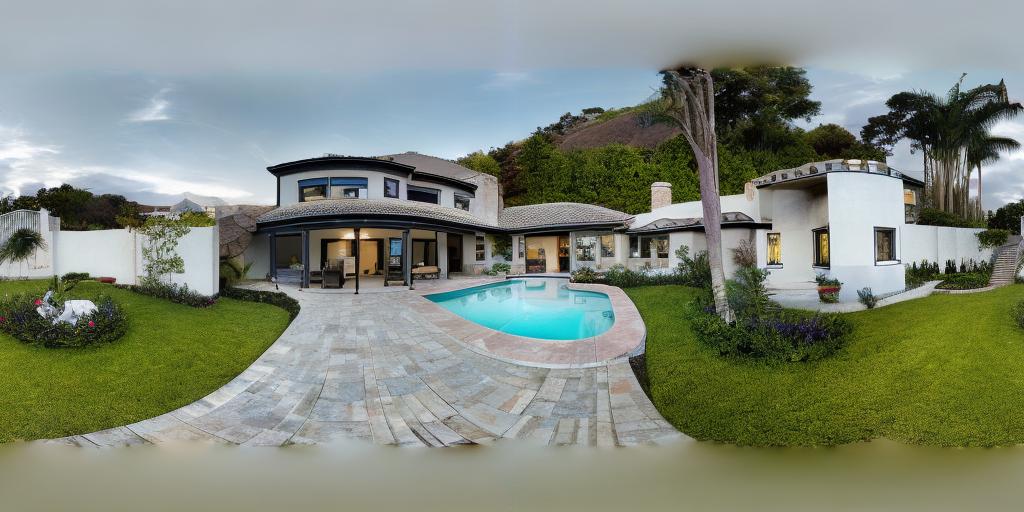}} & \multicolumn{4}{c}{\includegraphics[width=0.48\linewidth]{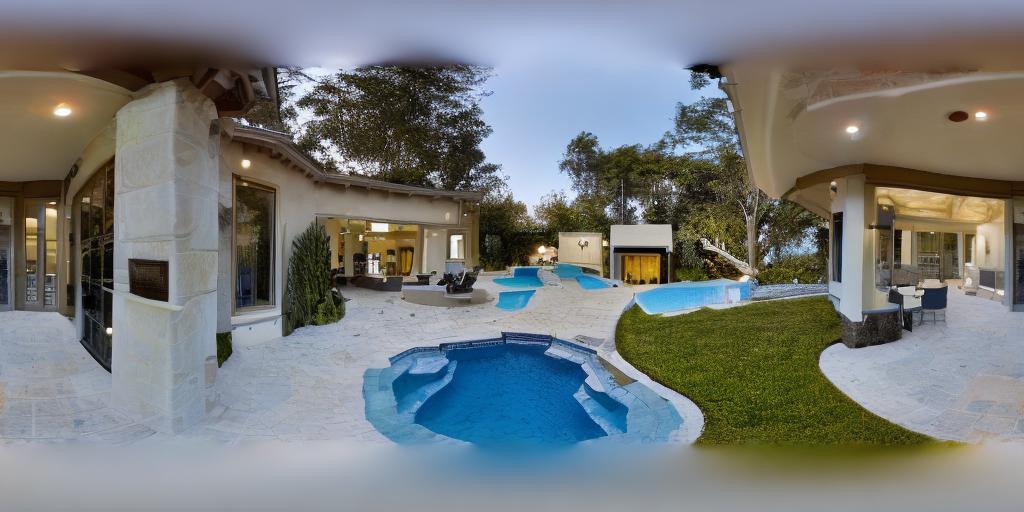}} \\
         & \includegraphics[width=0.117\linewidth]{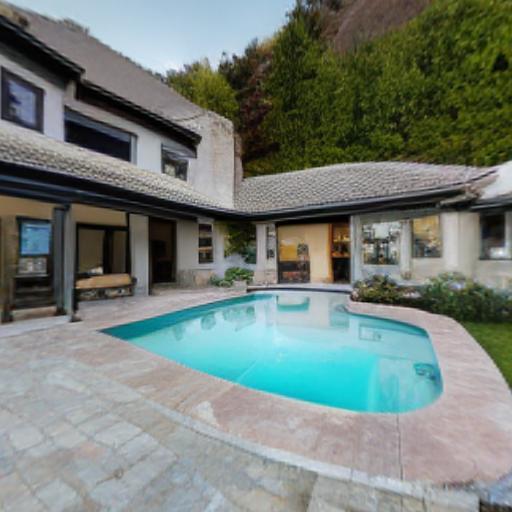} & \includegraphics[width=0.117\linewidth]{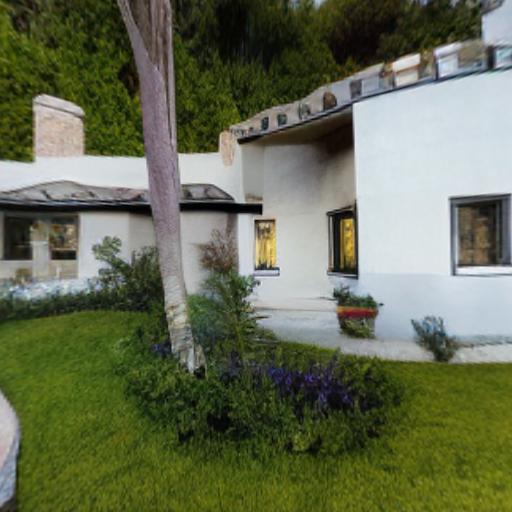} & \includegraphics[width=0.117\linewidth]{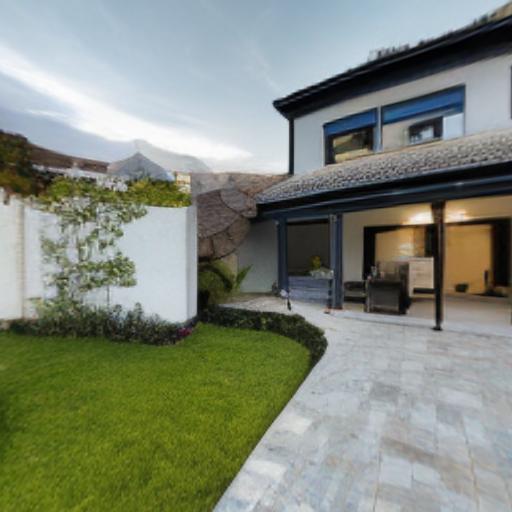} & \includegraphics[width=0.117\linewidth]{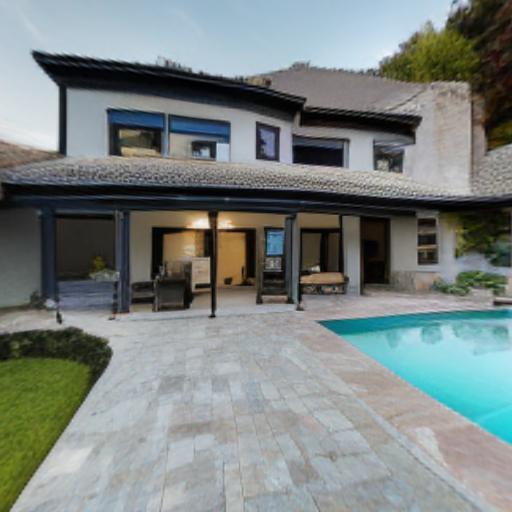} & \includegraphics[width=0.117\linewidth]{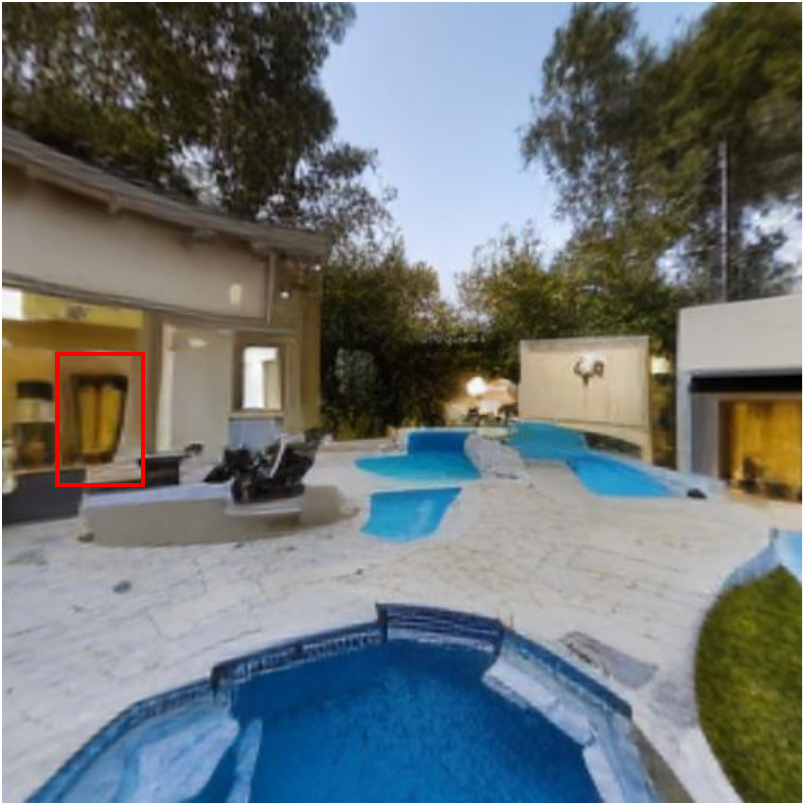} & \includegraphics[width=0.117\linewidth]{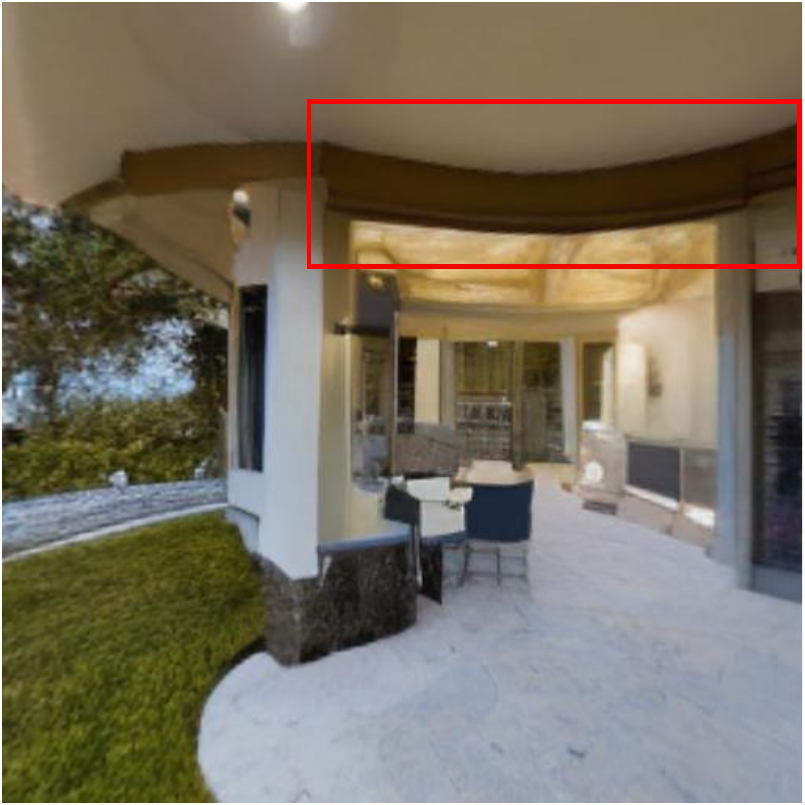} & \includegraphics[width=0.117\linewidth]{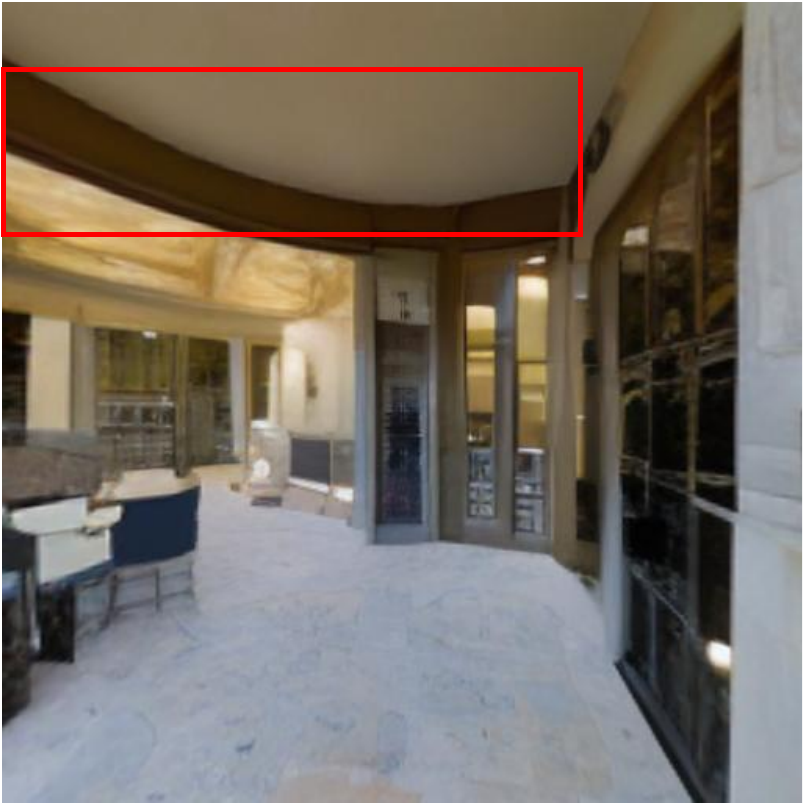} & 
         \includegraphics[width=0.117\linewidth]{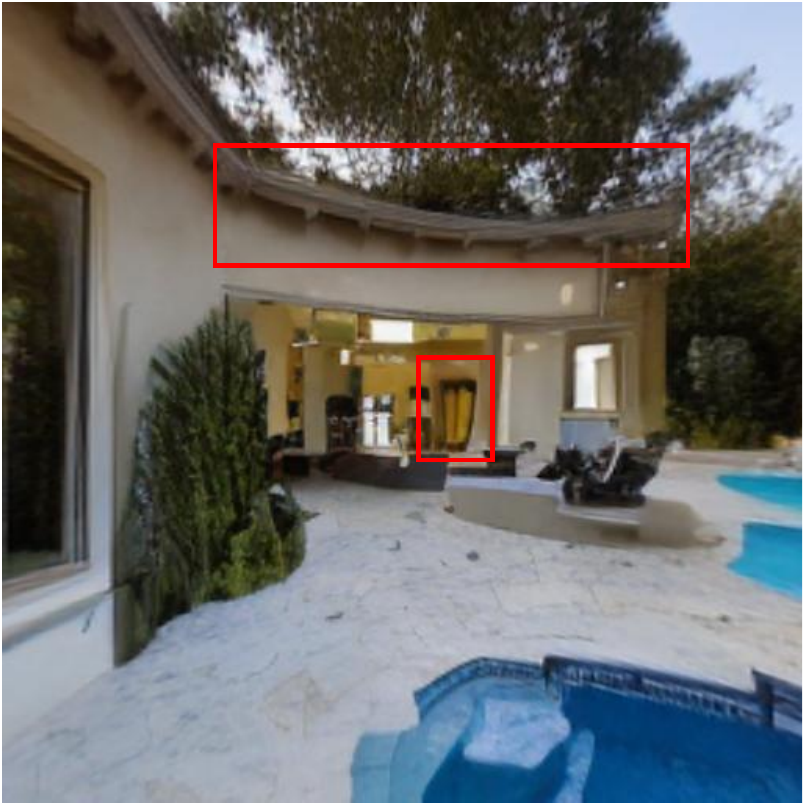} \\
         % \\
         % \multirow{2}{*}[23mm]{\rotatebox[origin=c]{90}{\emph{a large living room}}} & \multicolumn{4}{c}{\includegraphics[width=0.48\linewidth]{figs/qualitative_comparison/a large living room/unipano.jpg}} & \multicolumn{4}{c}{\includegraphics[width=0.48\linewidth]{figs/qualitative_comparison/a large living room/panfusion.jpg}} \\
         % & \includegraphics[width=0.117\linewidth]{figs/qualitative_comparison/a large living room/unipano_pers_0.jpg} & \includegraphics[width=0.117\linewidth]{figs/qualitative_comparison/a large living room/unipano_pers_1.jpg} & \includegraphics[width=0.117\linewidth]{figs/qualitative_comparison/a large living room/unipano_pers_2.jpg} & \includegraphics[width=0.117\linewidth]{figs/qualitative_comparison/a large living room/unipano_pers_3.jpg} & \includegraphics[width=0.117\linewidth]{figs/qualitative_comparison/a large living room/panfusion_pers_0.jpg} & \includegraphics[width=0.117\linewidth]{figs/qualitative_comparison/a large living room/panfusion_pers_1.jpg} & \includegraphics[width=0.117\linewidth]{figs/qualitative_comparison/a large living room/panfusion_pers_2.jpg} & 
         % \includegraphics[width=0.117\linewidth]{figs/qualitative_comparison/a large living room/panfusion_pers_3.jpg} \\
         % \\
         & \multicolumn{4}{c}{{UniPano (Ours)}} & \multicolumn{4}{c}{{PanFusion}~\citep{panfusion}}
    \end{tabular}
    \caption{\textbf{Selected qualitative comparisons between UniPano (Ours) and PanFusion}. We show the generated panoramic image for each text prompt and 4 randomly sampled horizontal perspective views below. We highlight notable artifacts such as non-perspective lines with \textcolor{red}{red boxes}. More qualitative comparisons can be found in the supplementary material.
    }
    \label{fig:qualitative}
\end{figure*}

\section{Experiments}

\subsection{Experimental Setup}\label{sec:exp-setup}

\paragraph{Dataset.}
Matterport3D dataset~\citep{Matterport3D} is a scene understanding dataset with 10,800 panoramic images. We use the same captions as~\citep{panfusion}, which are generated by BLIP-2~\citep{blip2} with a prompt of ``a 360 - degree view of''. We adopt the same data split as~\citep{panfusion,mvdiff}, containing 9,820 and 1,092 pairs for training and evaluation respectively.

\paragraph{Implementation Details.}
To facilitate a fair comparison, we strictly follow~\cite{panfusion,mvdiff} to train our model using AdamW optimizer~\citep{adamw} with a batch size of 4 and a learning rate of $2\times 10^{-4}$ for 10 epochs, with identical cosine annealing learning rate scheduler. Following \citep{panfusion}, we base our model on Stable Diffusion 2 base version. 

\paragraph{Evaluation Metrics.}
We follow previous works to evaluate the generated panoramic images in the panorama~\cite{panfusion,chen2022text2light} and perspective~\cite{mvdiff} domain. 
\begin{itemize}
    \item \emph{Panorama.} Following~\cite{panfusion,chen2022text2light}, we report Fr\'echet Inception Distance (FID) and Inception Score (IS) to measure the quality and realism of the generated panoramas. In addition, we report the CLIP Score (CS) to evaluate text-image consistency. Since both FID and IS are based on InceptionNet~\citep{inceptionnet} which is trained using perspective images only, we follow~\cite{panfusion} to report a panoramic-customized metric Fr\'echet Auto-Encoder Distance (FAED)~\citep{faed} for panorama evaluation.
    \item \emph{Perspective.} We follow~\cite{panfusion} to randomly sample 20 perspective views to simulate practical navigation on panoramas, and these views are evaluated based on FID and IS. Following~\cite{mvdiff}, we also report FID, IS, and CS on 8 horizontally evenly spaced views.
\end{itemize}

As stated in~\cite{panfusion}, IS evaluates the diversity of objects within the generated image, as such, lower IS does not necessarily reflect the quality and realism of images in case models do not tend to generate unexpected objects. Similarly, as in~\cite{panfusion}, higher CS may be due to the repetition of objects to strengthen text-image alignment. On top of these, we also note that the 20 randomly sampled views may capture the top or the bottom of the panoramas, which are often blurred even on real panoramic images, impacting the evaluation quality. 
\emph{For the above reasons, we emphasize FAED and FID while caring horizontal FID more than 20-view FID among all evaluation metrics.}

\subsection{Main Results}\label{sec:main-res}

\paragraph{Compared Methods.}
We compare our uni-branch approach with several baseline methods.
\begin{itemize}
\item \emph{MVDiffusion}~\citep{mvdiff} trains a multi-view diffusion model to simultaneously generate 8 horizontal views, which can be stitched into a panorama.
\item \emph{SD+LoRA}~\citep{lora,ldm} is the baseline method which fine-tunes the Stable Diffusion model~\citep{ldm} with LoRA~\citep{lora} on panoramic images. 
\item \emph{Pano Only}~\citep{panfusion} is another baseline method introduced in~\citep{panfusion} which additionally includes circular padding on top of SD+LoRA to ensure loop consistency. 
\item \emph{PanFusion}~\citep{panfusion} is the SoTA solution to date, which adopts a dual-branch approach and adds a cross-attention module between panoramic and perspective branches to ensure consistency.
\end{itemize}

\paragraph{Quantitative Comparison.}
We present the quantitative comparison in~\cref{tab:main-exp}. 
Our UniPano achieves state-of-the-art FAED and horizontal FID, while being on par with PanFusion on FID and 20-view FID. 
We highlight that UniPano introduces minimal computational overhead, with an additional 2.8\% allocated GPU memory and about 1 additional hour of training compared to SD+LoRA baseline. In comparison, training PanFusion~\citep{panfusion} almost double the allocated GPU memory (+89.7\%) and almost triple the time required for training compared to SD+LoRA baseline.

\paragraph{Qualitative Comparison.}
We showcase the qualitative comparison in~\cref{fig:qualitative}. PanFusion may sometimes generate panoramic images with invalid equirectangular projection, evidenced by the notable artifacts (curvy and panoramic-like lines) in the regions of the perspective views highlighted with \textcolor{red}{red boxes}. With the boosted capacity of the panoramic-specific modules, UniPano faithfully generates panoramic images that follow equirectangular projection in the illustrated cases. 
More qualitative comparisons are deferred to the supplementary material.

\input{tabs/ablation}

\subsection{Ablation Study}

\paragraph{Different fine-tuning strategies.}
We compare several different fine-tuning strategies for trainable weights $W_{\{q,k,v,o\}}$, in~\cref{tab:ab}, with different combinations of the following: freezing, fine-tuning with LoRA~\citep{lora}, and with mixture of experts (MoE)~\citep{moe,mole,ditmoe}. The panorama branch only baseline (highlighted in gray) fine-tunes all trainable components with LoRA and is thereby considered as the baseline. Switching to MoE for fine-tuning $W_o$ contributes to a notable improvement in FAED and FID metrics, demonstrating the benefits of increasing the capacity for panoramic-specific components. Based on the baseline, we further experiment with keeping $W_{\{q,k\}}$ frozen throughout fine-tuning, resulting in improved FID scores while the FAED remained comparable to the baseline. This result is in line with our analysis that $W_{\{q,k\}}$ relate to non-parametric-specific information. Merging these two strategies -- \ie, freezing $W_{\{q,k\}}$ and fine-tuning $W_o$ with MoE -- yields our state-of-the-art UniPano.

\input{tabs/expert_ablation}

\paragraph{Different settings for mixture of experts.}
We additionally ablate different sets of hyperparameters for the mixture of experts (MoE) in~\cref{tab:exp-ab}, namely the number of experts $n$ and the number of top-$k$ selected experts per token. Our first observation is that unlike many other MoE applications~\citep{ditmoe,mole,moe}, the performance in our context saturates for a relatively small amount of experts ($n=4$), evidenced by the notable deterioration in FID-related metrics when scaling $n$ from $4$ to $8$. This is likely due to the relative simplicity of fine-tuning a pre-trained diffusion model, in comparison to the typical use cases of MoE \ie training the entire model from scratch. Additionally, aligning with the previous work~\citep{moe,mole,ditmoe}, we find that sparsity is critical, as increasing the number of used experts per token $k$ from $2$ to $4$ reduces the FAED and FID metrics rapidly.

\subsection{Scaling to Higher Resolution}
Panoramic images store the entire 360-degree surrounding scenes within one equirectangular image, cropping perspective views from a panoramic image up to $512\times 1024$ thus still leading to somewhat low-resolution images. This motivates the importance of scaling panorama generation to higher resolution, which is typically achieved with a separate super-resolution stage~\citep{chen2022text2light}.
As a direct benefit of lowering the memory burden of PanFusion, UniPano can be readily scaled for higher-resolution panorama generation in an end-to-end manner.
As Stable Diffusion 2 base is optimized for generating images up to $512\times 512$, directly adapting it to generate higher-resolution images leads to suboptimal results. We thus adopt the state-of-the-art Stable Diffusion 3~\citep{sd3} which natively supports $1024\times 1024$ image generation. We show UniPano based on Stable Diffusion 3 can generate realistic $1024\times 2048$ panoramic images, and is robust to out-of-distribution prompts and extremely complex prompts, in~\cref{fig:pre-res}. More high-resolution results and experimental details are deferred to the supplementary material.

\begin{figure}[!t]
    \centering
    \setlength{\tabcolsep}{1pt}
    \def\arraystretch{0.3}
    \begin{tabular}{cc}
        \includegraphics[width=0.49\linewidth]{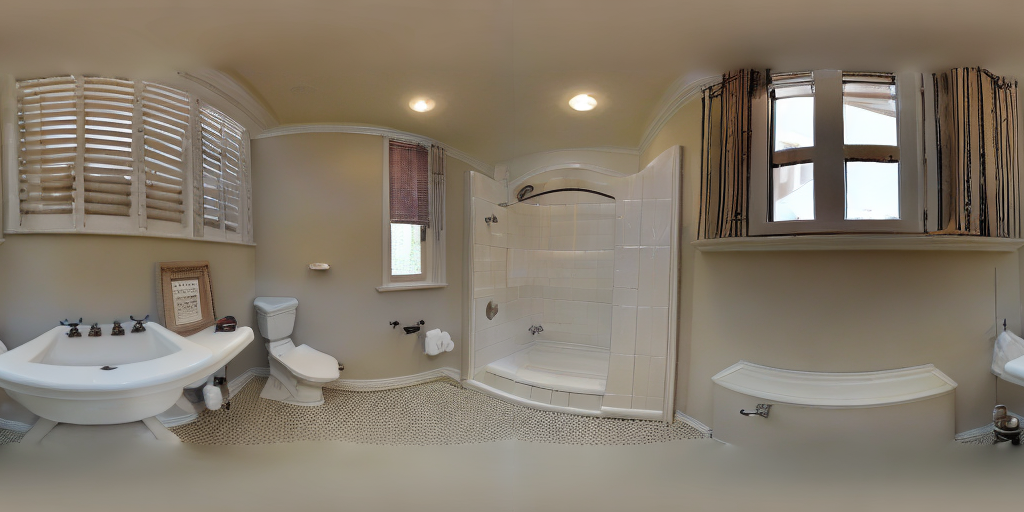} & \includegraphics[width=0.49\linewidth]{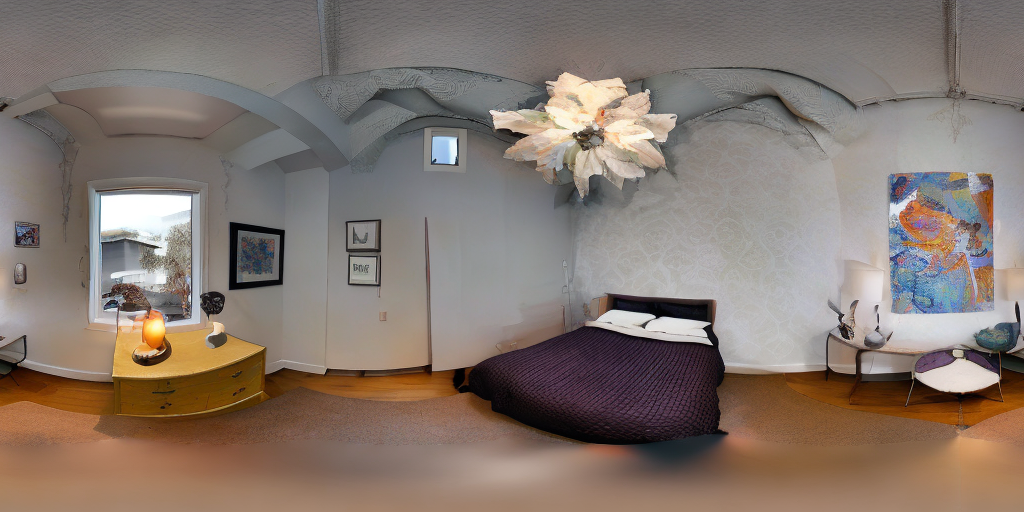}
    \end{tabular}
    \caption{\textbf{Failure cases of UniPano.} Similar to PanFusion~\citep{panfusion}, UniPano sometimes generates scenes with invalid layouts, such as rooms without entrances.}
    \label{fig:failure-case}
\end{figure}

\section{Conclusion}
We have elucidated the underlying mechanism that facilitates low-rank adaptation of pre-trained perspective diffusion models to panorama generation. Particularly, our analysis reveals that the query and key matrices ($W_{\{q,k\}}$) learn common semantic information that can be shared between the panoramic and perspective domains, whereas the value and output matrices ($W_{\{v,o\}}$) specialize in capturing the equirectangular structure of panoramic images. Based on these insights, we propose \emph{UniPano}, which outperforms and reduces the memory and training burden compared to the previous dual-branch approach.

\paragraph{Limitations.}
The primary focus of this paper is to investigate the underlying behaviors of the trainable components within LoRAs when adapting pre-trained perspective diffusion models to panorama generation. While our UniPano reports the state-of-the-art results, given the abundance of hyperparameters, the performance is still possibly far from optimal. 
% Future works may optimize the performance further. 
Additionally, similar to the drawbacks of PanFusion~\citep{panfusion}, we find that UniPano sometimes generates scenes with invalid layouts, such as rooms without entrances, as shown in~\cref{fig:failure-case}.

{
    \small
    \bibliographystyle{ieeenat_fullname}
    \bibliography{main}
}

\include{appendix}

\end{document}

%% file: tabs/qkvo_exp.tex
\begin{table}[!t]
    \centering
    \setlength{\tabcolsep}{10pt}
    \begin{tabular}{lcccc}
    \toprule
         & \multicolumn{2}{c}{Panorama} & \multicolumn{1}{c}{20 Views} & \multicolumn{1}{c}{8 Views} \\
        \cmidrule(lr){2-3} \cmidrule(lr){4-4} \cmidrule(lr){5-5}
         & FAED$\downarrow$ & FID$\downarrow$ & FID$\downarrow$ & FID$\downarrow$ \\
    % \hline
    \midrule
        $W_q$ & 10.86 & 81.09 & 30.66 & 27.44 \\
        {$W_k$} & 13.27 & 67.63 & 27.01 & 24.47 \\
        \rowcolor{gray!20}{$W_v$} & \textbf{8.66} & \underline{52.60} & \textbf{17.01} & \textbf{19.35} \\
        \rowcolor{gray!20}{$W_o$} & \underline{9.38} & \textbf{52.17} & \underline{20.32} & \underline{20.24} \\
    \bottomrule
    \end{tabular}
    \caption{\textbf{Quantitative comparison for training $W_{\{q,k,v,o\}}$ in isolation separately}. Training only $W_v$ or $W_o$ reports considerably better FAED and FID than $W_q$ or $W_k$. Details of reported metrics are in~\cref{sec:exp-setup}.}
    \label{tab:exp-qkvo}
\end{table}

%% file: tabs/ablation_wo_arch.tex
\begin{table}[!t]
    \centering
    \setlength{\tabcolsep}{3.5pt}
    \begin{tabular}{lcccc}
    \toprule
         & \multicolumn{2}{c}{Panorama} & \multicolumn{1}{c}{20 Views} & \multicolumn{1}{c}{8 Views} \\
        \cmidrule(lr){2-3} \cmidrule(lr){4-4} \cmidrule(lr){5-5}
         & FAED$\downarrow$ & FID$\downarrow$ & FID$\downarrow$ & FID$\downarrow$ \\
    \midrule
        Pano Only~\citep{panfusion} & 7.90 & 50.40 & 20.10 & 20.56 \\
    \midrule
        DA~\citep{deformableattn} & 9.78 & \textbf{46.41} & \textbf{16.56} & \textbf{18.77} \\
        SE~\citep{senet} & 8.72 & 50.67 & 17.71 & 19.61 \\
        % SA & 8.35 & 51.37 & 17.64 & \underline{19.36} \\
        LoRA ($r=8$)~\citep{lora} & 8.34 & \underline{48.58} & \underline{16.94} & {19.42} \\
        LA~\citep{nat} & \underline{7.65} & 50.39 & 18.24 & \underline{19.41} \\
        \rowcolor{gray!20}MoE~\citep{mole,ditmoe} & \textbf{7.21} & 48.83 & 19.50 & 20.05 \\
    \bottomrule
    \end{tabular}
    \caption{\textbf{Comparison across several designs for increasing the capacity of $W_o$}. Based on the panorama branch only baseline (Pano Only), we compare several plausible attempts to increase the capacity of the output layer ($W_o$). Details of reported metrics are in~\cref{sec:exp-setup}.}
    \label{tab:arch-ab}
\end{table}

%% file: tabs/main_exp.tex
\begin{table*}[!t]
    \centering
    \setlength{\tabcolsep}{4.6pt}
    \begin{tabular}{llcccccccccc}
    \toprule
        \multirow{2}{*}{Methods} & \multirow{2}{*}{\begin{tabular}{c}
                   Peak Mem.$^{\dagger}$\\
                   (GB)\\
                   \end{tabular}} & \multirow{2}{*}{\begin{tabular}{c}
                   Dur.$^{\dagger}$ \\
                   (hrs)\\
                   \end{tabular}} & \multicolumn{4}{c}{Panorama} & \multicolumn{2}{c}{20 Views} & \multicolumn{3}{c}{Horizontal 8 Views} \\
        \cmidrule(lr){4-7} \cmidrule(lr){8-9} \cmidrule(lr){10-12}
         & & & \textbf{FAED}$\downarrow$ & \textbf{FID}$\downarrow$ & IS$\uparrow$ & CS$\uparrow$ & \textbf{FID}$\downarrow$ & IS$\uparrow$ & \textbf{FID}$\downarrow$ & IS$\uparrow$ & CS$\uparrow$ \\
    \midrule
        SD+LoRA~\citep{panfusion,lora,ldm} & 31.69 & 2.26 & 7.19 & 51.69 & \underline{4.40} & \textbf{28.83} & 19.32 & \underline{6.90} & 20.68 & 6.48 & 24.77 \\
        MVDiffusion~\citep{mvdiff} & 26.66 (-15.9\%) & 9.86 & - & - & - & - & - & - & 25.27 & \textbf{6.90} & \textbf{26.34} \\
        Pano Only~\citep{panfusion} & 31.81 (+0.4\%) & 2.33 & 7.90 & \underline{50.40} & \textbf{4.54} & \underline{28.67} & 20.10 & \textbf{7.06} & 20.56 & 6.37 & 24.85 \\
        PanFusion~\citep{panfusion} & 60.12 (+89.7\%) & 6.61 & \underline{6.04} & \textbf{46.47} & 4.36 & 28.58 & \textbf{17.04} & 6.85 & \underline{19.88} & \underline{6.50} & \underline{24.98} \\
        \rowcolor{gray!20}UniPano (\textbf{Ours}) & 32.59 (+2.8\%) & 3.43 & \textbf{5.90} & \textbf{46.47} & 4.16 & 28.37 & \underline{17.09} & 6.74 & \textbf{17.74} & 6.00 & 24.82 \\
    \bottomrule
    \end{tabular}
    \caption{\textbf{Comparison between SoTA methods on 512$\times$1024 panorama generation}. We quantitatively evaluate the panorama images based on Fr\'echet Auto-Encoder Distance (FAED) Fr\'echet Inception Distance (FID), Inception Score (IS), and CLIP Score (CS). We follow \cite{panfusion} to randomly sample 20 views from a panoramic image and \cite{mvdiff} to horizontally sample 8 evenly spaced views to evaluate the quality of cropped perspective images. We report the peak allocated GPU memory (Peak Mem.) and time duration (Dur.) for 10-epoch training. All evaluated results are based on Stable Diffusion 2 base. $\dagger$: results are reproduced with \texttt{FP32} precision.}
    \label{tab:main-exp}
\end{table*}

%% file: tabs/ablation.tex
\begin{table}[!t]
    \centering
    % \scriptsize
    % \renewcommand{\arraystretch}{1.1}
    \setlength{\tabcolsep}{2.5pt}
    \begin{tabular}{ccccccc}
    \toprule
        & & & \multicolumn{2}{c}{Panorama} & \multicolumn{1}{c}{20 Views} & \multicolumn{1}{c}{8 Views} \\
        \cmidrule(lr){4-5} \cmidrule(lr){6-6} \cmidrule(lr){7-7}
        $W_{\{q,k\}}$ & $W_v$ & $W_o$ & FAED$\downarrow$ & FID$\downarrow$ & FID$\downarrow$ & FID$\downarrow$ \\
    \midrule
        LoRA & LoRA & LoRA & 7.90 & 50.40 & 20.10 & 20.56 \\
        LoRA & LoRA & MoE & \underline{7.21} & 48.83 & 19.50 & 20.05 \\
        \textcolor{cyan}{\footnotesize\SnowflakeChevron} & LoRA & LoRA & 7.99 & \underline{48.62} & \underline{19.27} & \underline{19.30} \\
        % \textcolor{cyan}{\footnotesize\SnowflakeChevron} & \textcolor{cyan}{\footnotesize\SnowflakeChevron} & MoE & 7.25 & 47.02 & 4.08 & 28.35 & 17.44 & 6.58 & 17.34 & 5.68 & 24.79 \\
        \rowcolor{gray!20}\textcolor{cyan}{\footnotesize\SnowflakeChevron} & LoRA & MoE & \textbf{5.90} & \textbf{46.47} & \textbf{17.09} & \textbf{17.74} \\
    \bottomrule
    \end{tabular}
    \caption{\textbf{Ablation study on fine-tuning strategies}. We compare several fine-tuning strategies for weights $W_{\{q,k,v,o\}}$ in attention modules, with various combinations of freezing (\textcolor{cyan}{\footnotesize\SnowflakeChevron}), LoRA fine-tuning~\citep{lora}, and MoE~\citep{moe,mole}. 
    % Our best approach is to freeze $W_{\{q,k\}}$, fine-tune $W_v$ with LoRA, and increase the capacity of $W_o$ with MoE.
    }
    \label{tab:ab}
\end{table}

%% file: tabs/expert_ablation.tex
% \begin{table*}[!t]
%     \centering
%     \begin{tabular}{lccccccccc}
%     \toprule
%          & \multicolumn{4}{c}{Panorama} & \multicolumn{2}{c}{20 Views} & \multicolumn{3}{c}{Horizontal 8 Views} \\
%         \cmidrule(lr){2-5} \cmidrule(lr){6-7} \cmidrule(lr){8-10}
%          & FAED$\downarrow$ & FID$\downarrow$ & IS$\uparrow$ & CS$\uparrow$ & FID$\downarrow$ & IS$\uparrow$ & FID$\downarrow$ & IS$\uparrow$ & CS$\uparrow$ \\
%     \midrule
%         % $n=2$, $k=1$ & 7.71 & 48.87 & 4.32 & 28.16 & 18.46 & 7.02 & 18.96 & 6.24 & 24.74 \\
%         $n=2$, $k=2$ & 6.75 & 51.01 & 4.32 & 28.49 & 18.27 & 6.90 & 19.71 & 6.36 & 24.73 \\
%         $n=4$, $k=2$ & 5.90 & 46.47 & 4.16 & 28.37 & 17.09 & 6.74 & 17.74 & 6.00 & 24.82 \\
%         % $n=6$, $k=2$ & 7.14 & 45.38 & 4.06 & 28.18 & 19.18 & 6.75 & 19.68 & 6.05 & 24.64 \\
%         $n=8$, $k=2$ & 6.17 & 47.10 & 4.02 & 28.28 & 20.19 & 7.01 & 19.51 & 6.29 & 24.67 \\
%         $n=8$, $k=4$ & 7.31 & 47.61 & 4.04 & 28.03 & 22.05 & 7.10 & 21.12 & 6.15 & 24.64 \\
%     \bottomrule
%     \end{tabular}
%     \caption{\textbf{Ablation study on the mixture of experts in $W_o$}. We experiment with different sets of hyperparameters for the mixture of experts (MoE) by adjusting the number of experts $n$, and selecting top-$k$ experts with the highest weighting. Selecting the top-2 experts out of 4 total experts gives the best results.}
%     \label{tab:exp-ab}
% \end{table*}

\begin{table}[!t]
    \centering
    \setlength{\tabcolsep}{8.5pt}
    \begin{tabular}{cccccc}
    \toprule
         & & \multicolumn{2}{c}{Panorama} & \multicolumn{1}{c}{20 Views} & \multicolumn{1}{c}{8 Views} \\
        \cmidrule(lr){3-4} \cmidrule(lr){5-5} \cmidrule(lr){6-6}
        $n$ & $k$ & FAED$\downarrow$ & FID$\downarrow$ & FID$\downarrow$ & FID$\downarrow$ \\
    \midrule
        % $n=2$, $k=1$ & 7.71 & 48.87 & 4.32 & 28.16 & 18.46 & 7.02 & 18.96 & 6.24 & 24.74 \\
        2 & 2 & 6.75 & 51.01 & \underline{18.27} & 19.71 \\
        \rowcolor{gray!20}4 & 2 & \textbf{5.90} & \textbf{46.47} & \textbf{17.09} & \textbf{17.74} \\
        % $n=6$, $k=2$ & 7.14 & 45.38 & 4.06 & 28.18 & 19.18 & 6.75 & 19.68 & 6.05 & 24.64 \\
        8 & 2 & \underline{6.17} & \underline{47.10} & 20.19 & \underline{19.51} \\
        8 & 4 & 7.31 & 47.61 & 22.05 & 21.12 \\
    \bottomrule
    \end{tabular}
    \caption{\textbf{Ablation study on the mixture of experts in $W_o$}. We experiment with different hyperparameters for the mixture of experts (MoE) by adjusting the number of experts $n$, and selecting top-$k$ experts with the highest weighting. 
    % Selecting the top-2 experts out of 4 total experts gives the best results.
    }
    \label{tab:exp-ab}
\end{table}

%% file: appendix.tex
\appendix

% \twocolumn[{
% \begin{center}
% \setlength{\baselineskip}{1.6\baselineskip}
% {\Large\bf What Makes for Text to 360-degree Panorama Generation with Stable Diffusion? \\
% \textit{-- Supplementary Material --}}
% \end{center}
% }]

\section{Preliminary on Diffusion Models}
For completeness sake, we provide preliminary on diffusion models, particularly latent diffusion models, below.

Diffusion models involve iteratively transforming the noise into the target data. The sampling step thus requires learning a time-conditioned noise (or equivalently the score function) prediction network $\epsilon_\theta$, often in the form of U-Net~\citep{unet} or transformers~\citep{dit}. In practice, to optimize efficiency and performance, the denoising process is generally performed in the latent space of a pre-trained encoder $\mcal{E}$, which leads to the training objective:
\begin{equation*}
    \min_\theta \mbb{E}_{t\sim\mcal{U}(0,T),(x,y)\sim p_{\mrm{data}},\epsilon\sim \mcal{N}(0,\mrm{Id})}\left[\|\epsilon_\theta(z_t,t,y)-\epsilon\|^2\right],
\end{equation*}
where $\mcal{U}$ is the uniform distribution, $p_\mrm{data}$ denotes the data distribution for which each sample contains an input image $x$ and an input condition $y$ (which is text in our context), $z_t=\alpha(t)\mcal{E}(x)+\beta(t)\epsilon$ is the noisy latent at a timestep $t$ with $\alpha(t)$ and $\beta(t)$ defining the diffusion trajectory, and $T$ is the largest timestep such that $z_T\sim\mcal{N}(0,\mrm{Id})$. During sampling, a random noise $z_T$ is first drawn from the prior distribution, and gradually denoised to the clean latent $z_0$ by the learned denoising network $\epsilon_\theta$ following a pre-defined noise schedule. The clean latent is finally converted into the image space using the pre-trained decoder $\mcal{D}$.

\section{Experimental Details}
We provide more details on the experimental setup of $512\times 1024$ panorama generation below.

\paragraph{Implementation Details.}
Our implementation is based on Stable Diffusion from \texttt{diffusers}~\citep{diffusers}. In addition to the implementation details listed in the main article, we strictly follow MVDiffusion~\citep{mvdiff} and PanFusion~\citep{panfusion} using the DDIM sampler~\citep{ddim} with 50 sampling steps and classifier-free guidance scale~\citep{cfg} of 9 for inference.

\paragraph{Compared Methods.}
We provide more details on the compared methods and the reported results in~\cref{tab:main-exp} below.

\begin{itemize}
    \item MVDiffusion~\citep{mvdiff} trains diffusion models with multi-view awareness, which generate 8 horizontal perspective views simultaneously. These images can then be stitched into a panorama, however, we note that the panoramas are incomplete due to the missing top and bottom regions. The reported results are directly taken from~\cite{panfusion}, where the only difference with the original MVDiffusion paper is to downsample to $256\times 256$ for evaluation to match the resolution of the ground truth images. 
    \item SD+LoRA~\citep{lora,ldm} directly fine-tunes Stable Diffusion with LoRA~\citep{lora} on panoramic images, which is a standard technique for adapting pre-trained diffusion models for downstream tasks. The reported metrics are directly taken from~\cite{panfusion}. 
    \item Pano Only~\citep{panfusion} is a baseline method proposed in~\cite{panfusion} which adopts circular padding on top of SD+LoRA to ensure loop consistency. The reported metrics are again taken from~\cite{panfusion}. 
    \item PanFusion~\citep{panfusion} is the state-of-the-art solution to date and our most important baseline method. It adopts a dual-branch approach, consisting of a panoramic and a perspective branch. It proposes an equirectangular-perspective projection attention module to establish a correspondence between these two branches to ensure consistency. The reported metrics are directly taken from the original PanFusion paper~\citep{panfusion}. 
\end{itemize}

\begin{figure*}[p]
    \centering
    \setlength{\tabcolsep}{1pt}
    \def\arraystretch{0.3}
    \begin{tabular}{cccccccc}

         \multicolumn{8}{c}{\emph{``a bedroom with a ceiling fan''}} \\ \multicolumn{4}{c}{\includegraphics[width=0.48\linewidth]{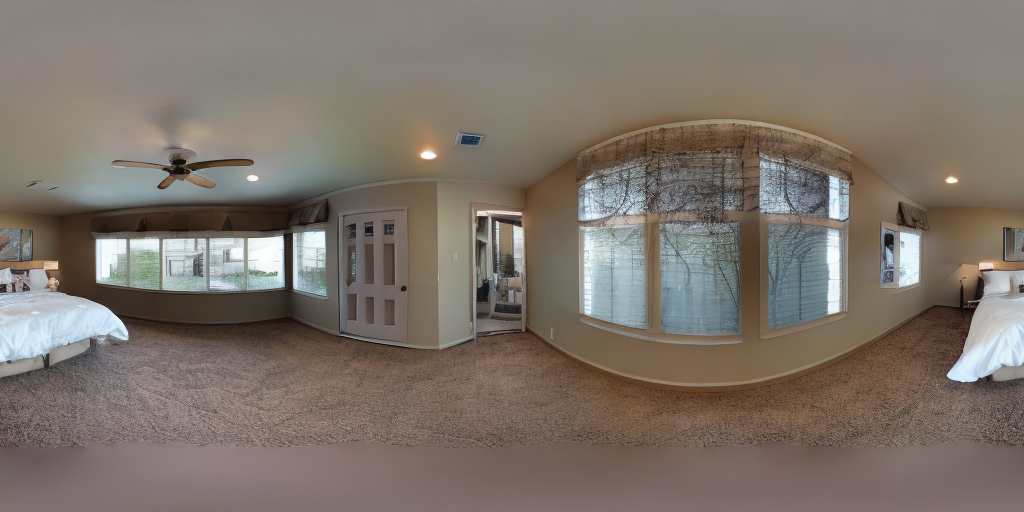}} & \multicolumn{4}{c}{\includegraphics[width=0.48\linewidth]{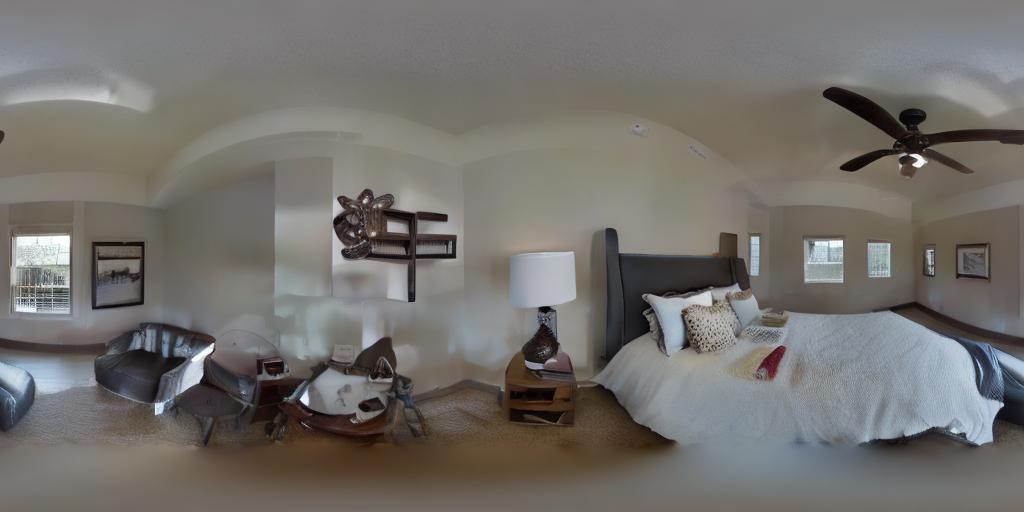}} \\
         \includegraphics[width=0.117\linewidth]{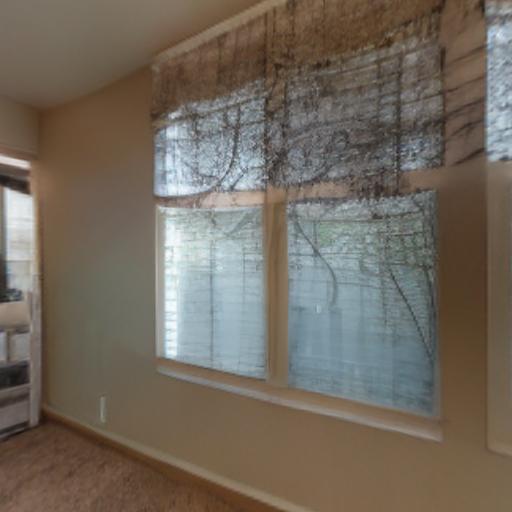} & \includegraphics[width=0.117\linewidth]{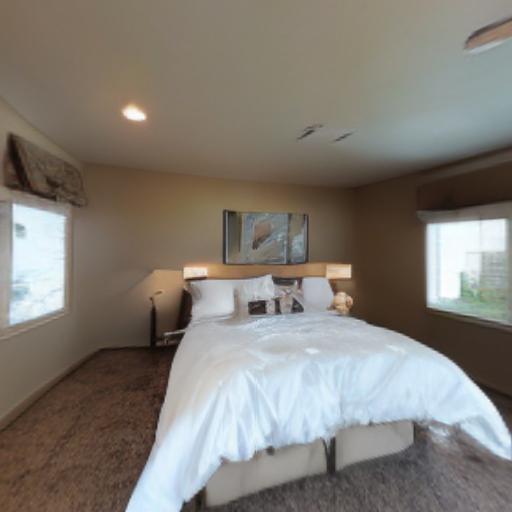} & \includegraphics[width=0.117\linewidth]{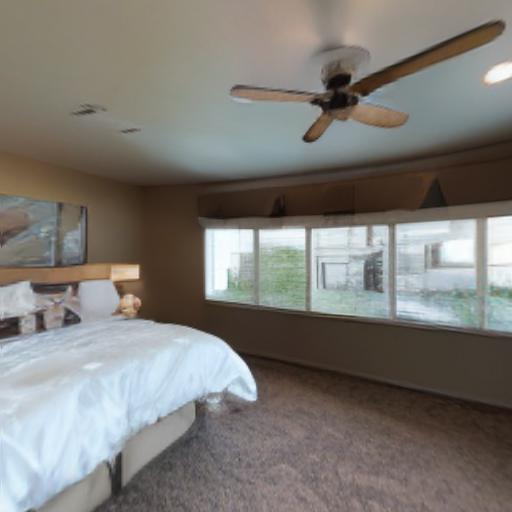} & \includegraphics[width=0.117\linewidth]{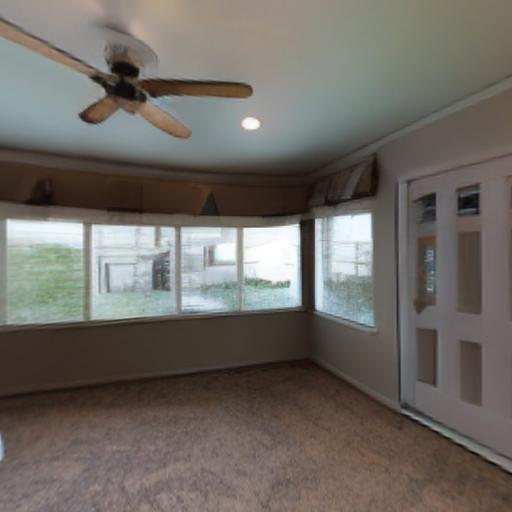} & \includegraphics[width=0.117\linewidth]{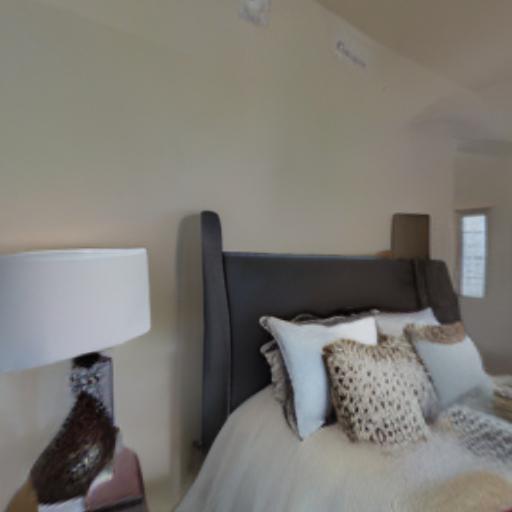} & \includegraphics[width=0.117\linewidth]{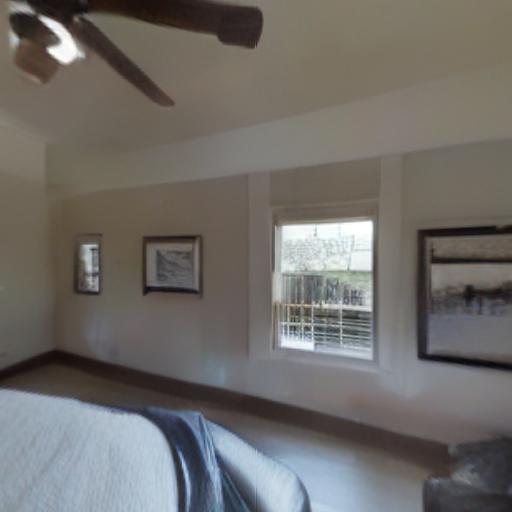} & \includegraphics[width=0.117\linewidth]{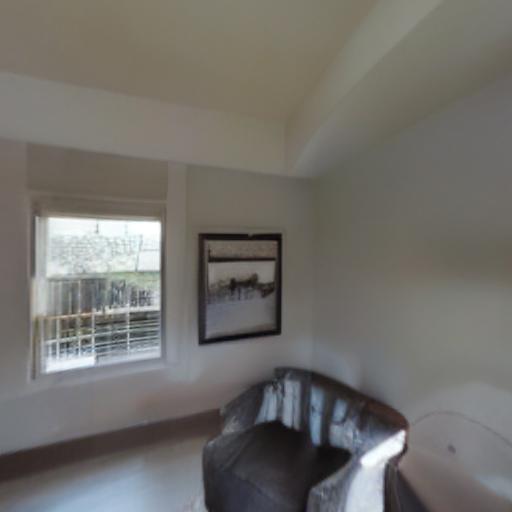} & 
         \includegraphics[width=0.117\linewidth]{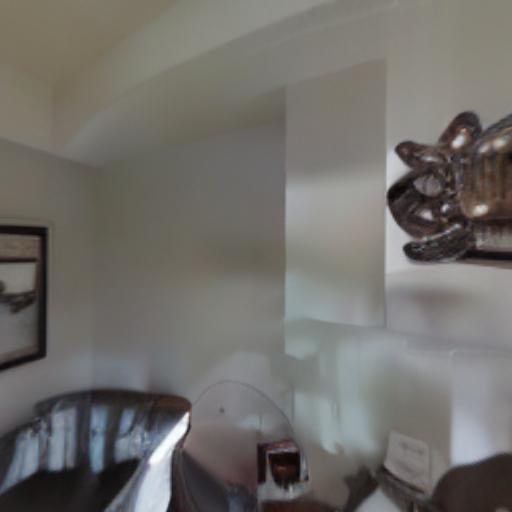} \\
         \\

         \multicolumn{8}{c}{\emph{``the interior of a store''}} \\
         \multicolumn{4}{c}{\includegraphics[width=0.48\linewidth]{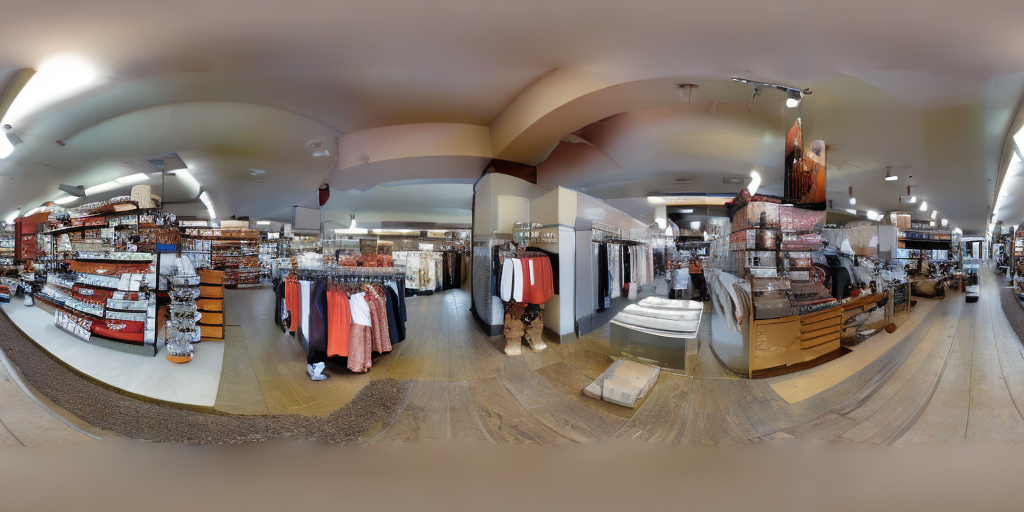}} & \multicolumn{4}{c}{\includegraphics[width=0.48\linewidth]{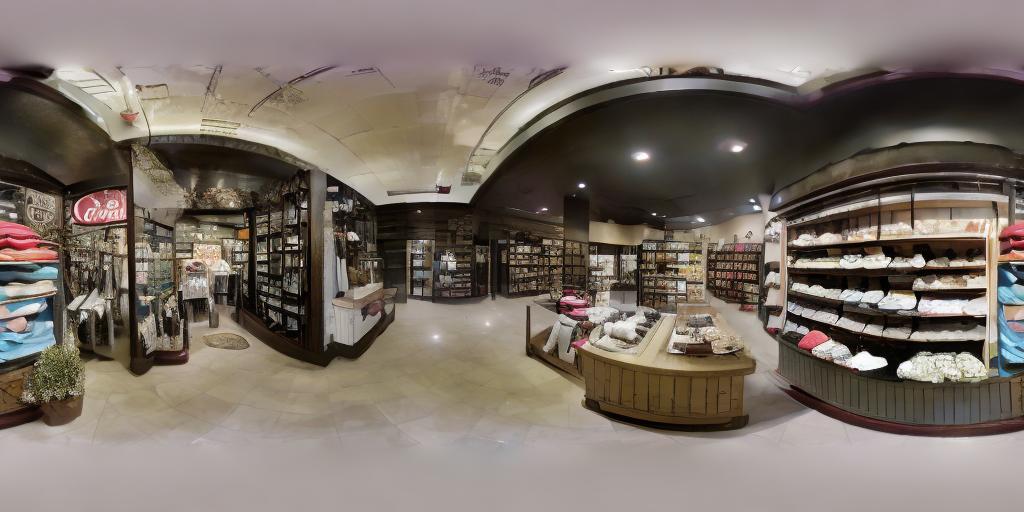}} \\
         \includegraphics[width=0.117\linewidth]{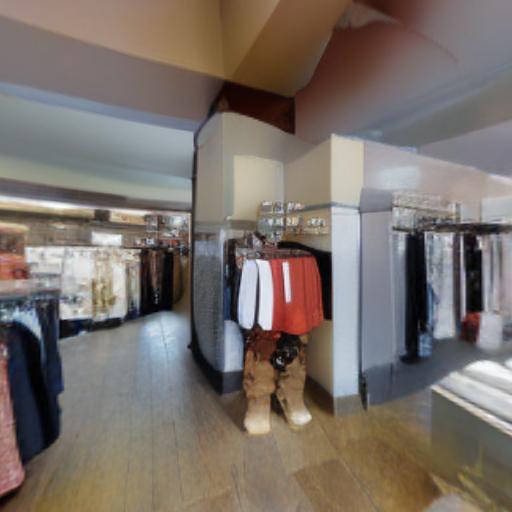} & \includegraphics[width=0.117\linewidth]{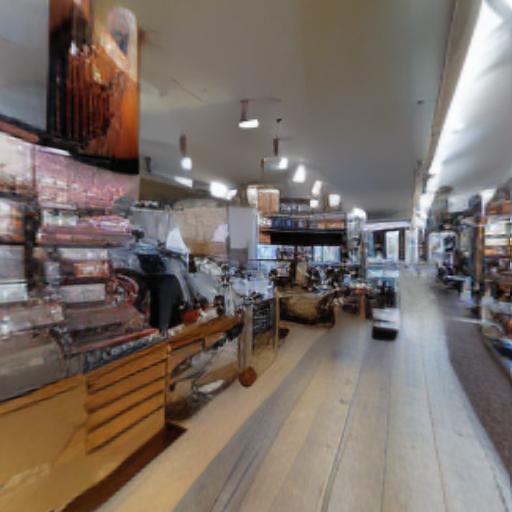} & \includegraphics[width=0.117\linewidth]{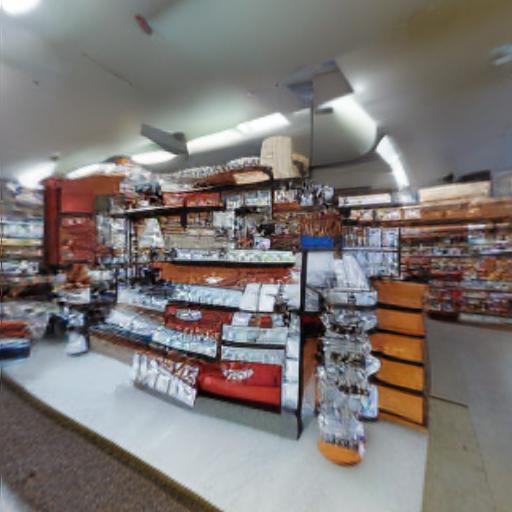} & \includegraphics[width=0.117\linewidth]{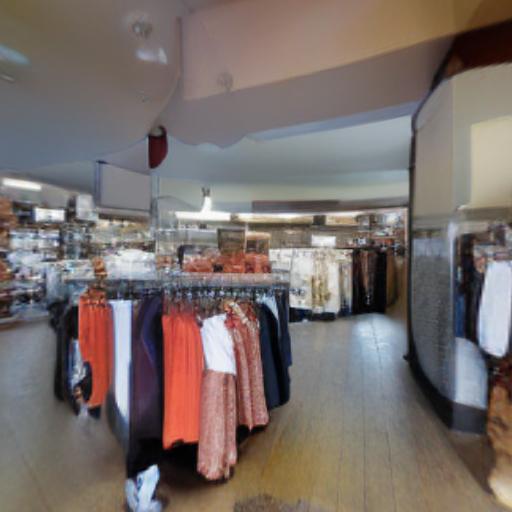} & \includegraphics[width=0.117\linewidth]{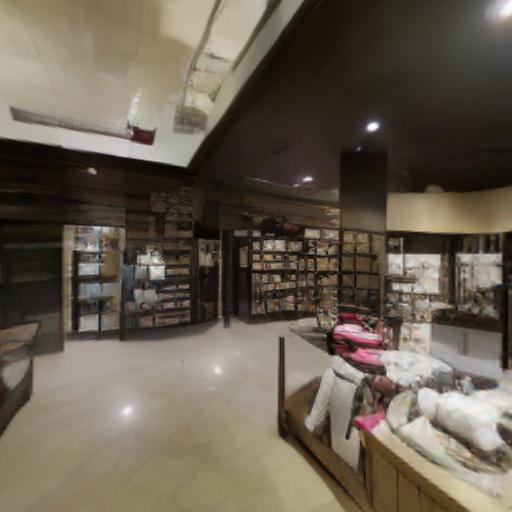} & \includegraphics[width=0.117\linewidth]{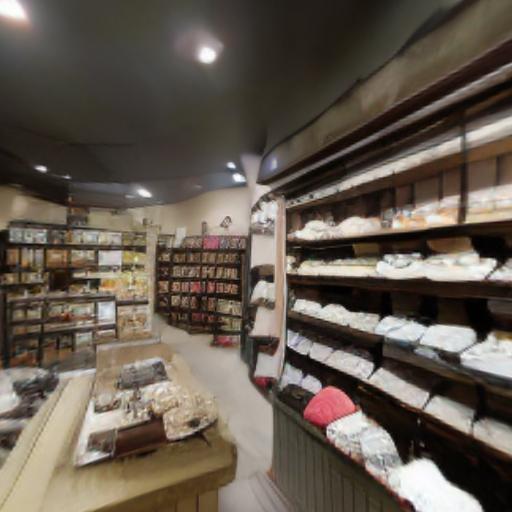} & \includegraphics[width=0.117\linewidth]{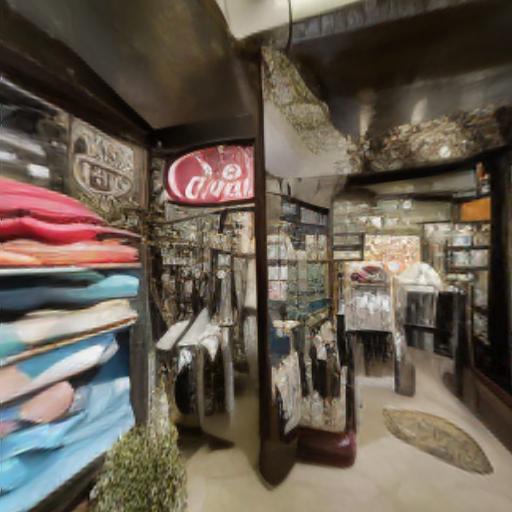} & 
         \includegraphics[width=0.117\linewidth]{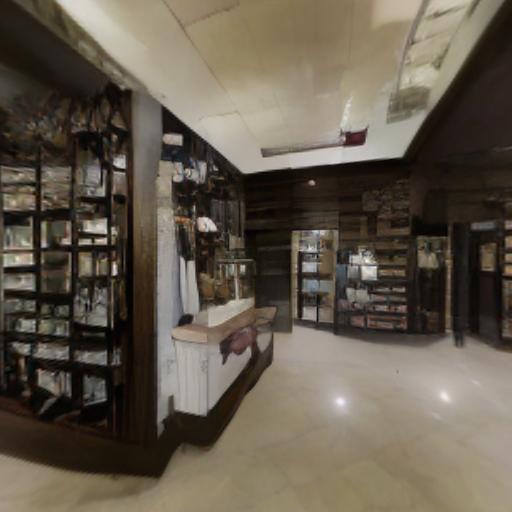} \\
         \\

         \multicolumn{8}{c}{{\emph{``a living room with bookshelves''}}} \\ \multicolumn{4}{c}{\includegraphics[width=0.48\linewidth]{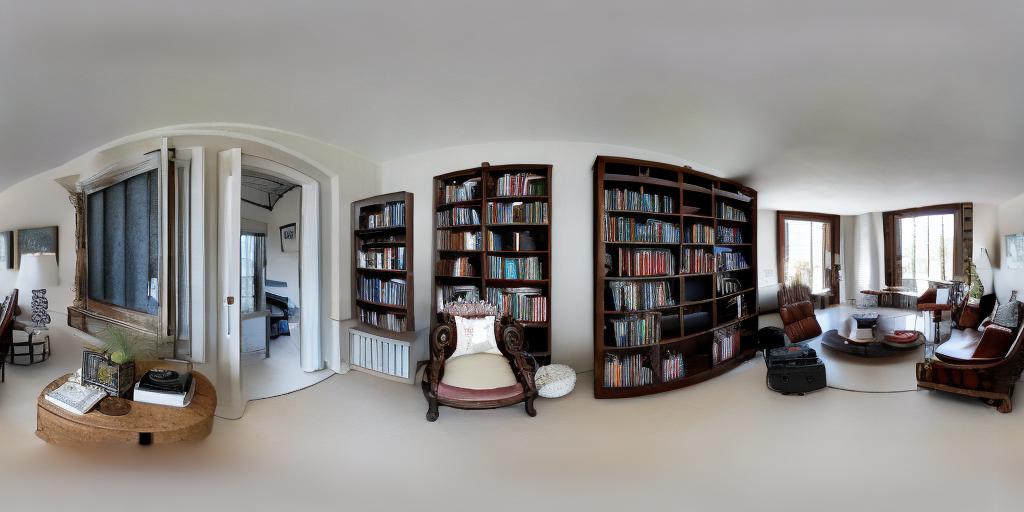}} & \multicolumn{4}{c}{\includegraphics[width=0.48\linewidth]{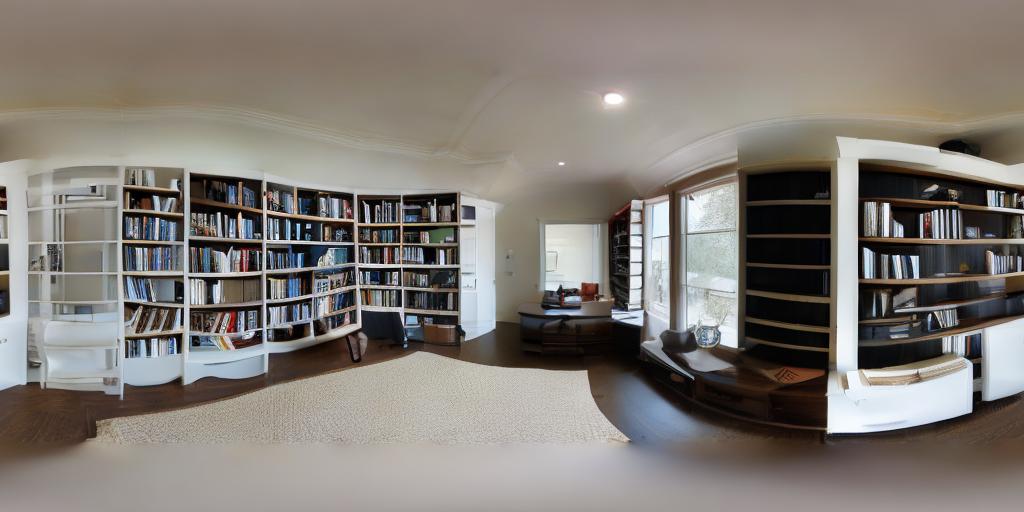}} \\
         \includegraphics[width=0.117\linewidth]{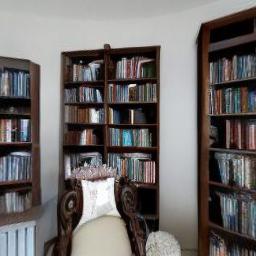} & \includegraphics[width=0.117\linewidth]{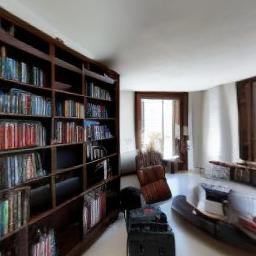} & \includegraphics[width=0.117\linewidth]{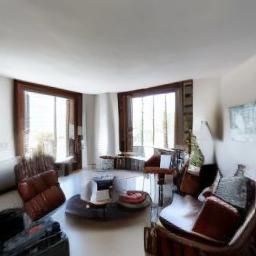} & \includegraphics[width=0.117\linewidth]{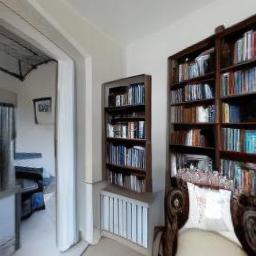} & \includegraphics[width=0.117\linewidth]{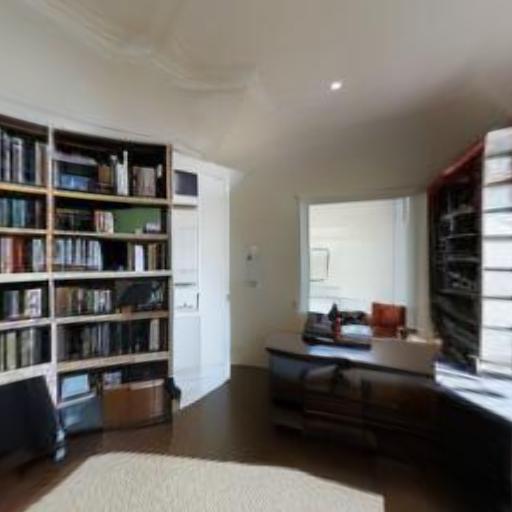} & \includegraphics[width=0.117\linewidth]{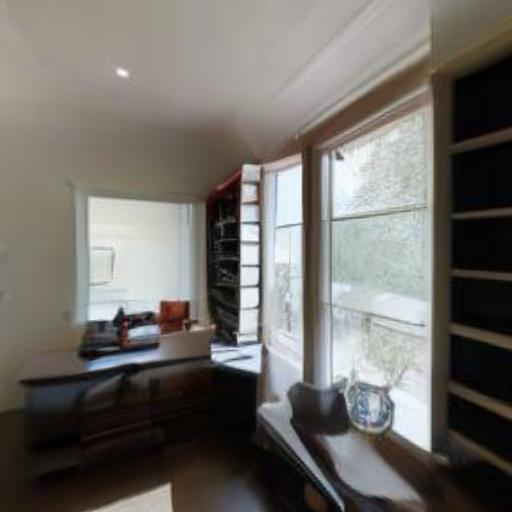} & \includegraphics[width=0.117\linewidth]{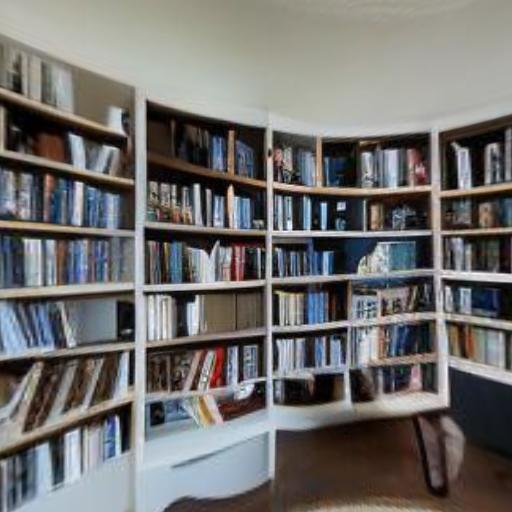} & 
         \includegraphics[width=0.117\linewidth]{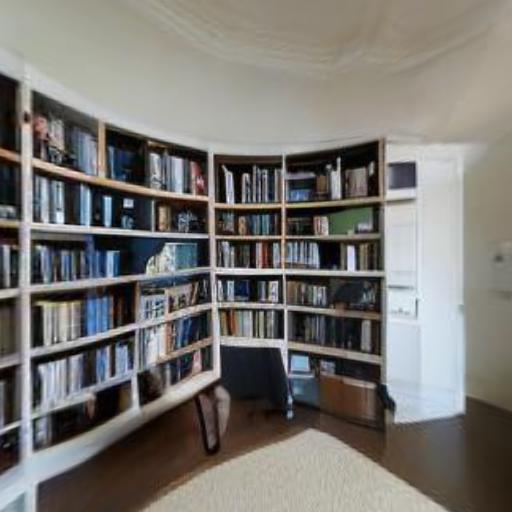} \\
         \\
         
         \multicolumn{4}{c}{{UniPano (Ours)}} & \multicolumn{4}{c}{{PanFusion}~\citep{panfusion}}
    \end{tabular}
    \caption{Additional qualitative comparisons.}
    \label{fig:add-qualitative}
\end{figure*}

\begin{figure*}[p]
    \centering
    \setlength{\tabcolsep}{1pt}
    \def\arraystretch{0.3}
    \begin{tabular}{cccccccc}
         
         \multicolumn{8}{c}{{\emph{``a house with a pool and mountains in the background''}}} \\ \multicolumn{4}{c}{\includegraphics[width=0.48\linewidth]{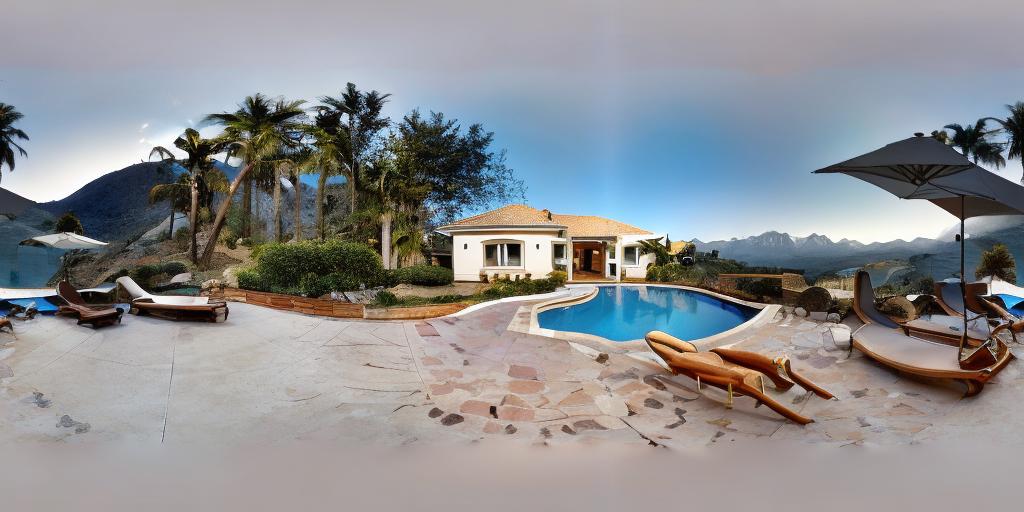}} & \multicolumn{4}{c}{\includegraphics[width=0.48\linewidth]{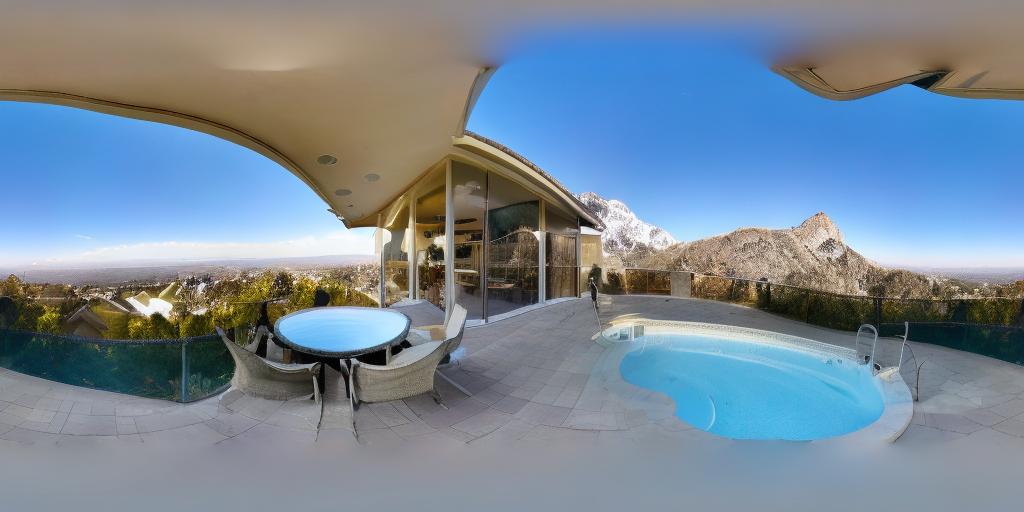}} \\
         \includegraphics[width=0.117\linewidth]{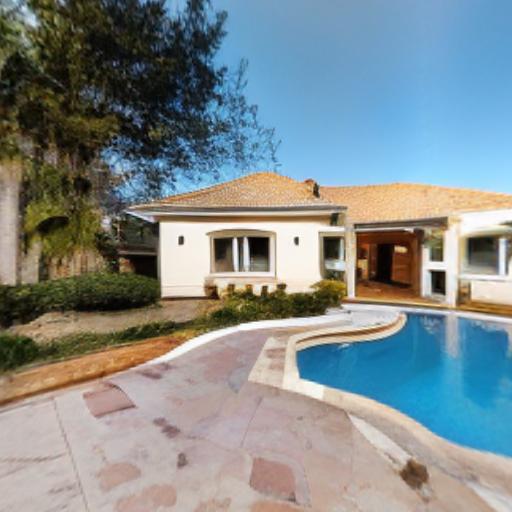} & \includegraphics[width=0.117\linewidth]{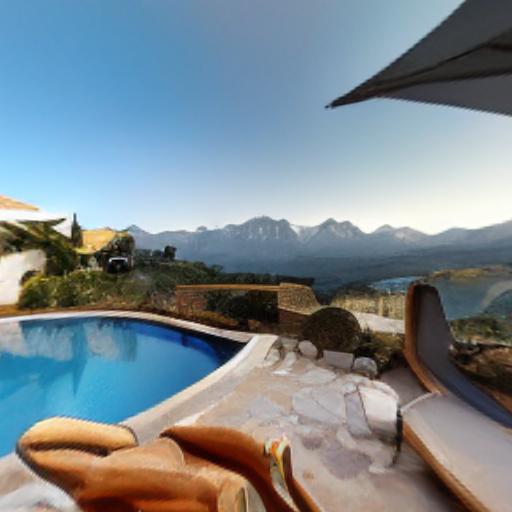} & \includegraphics[width=0.117\linewidth]{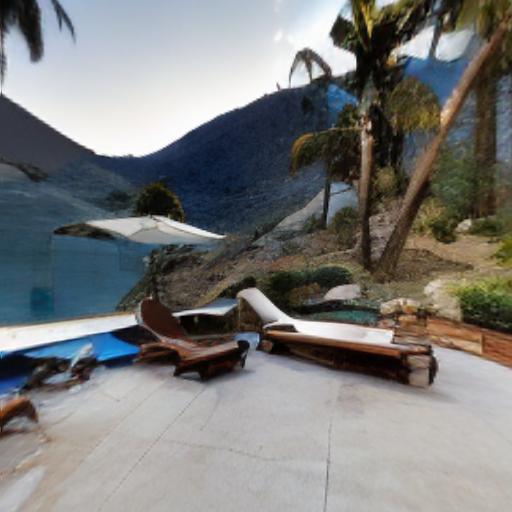} & \includegraphics[width=0.117\linewidth]{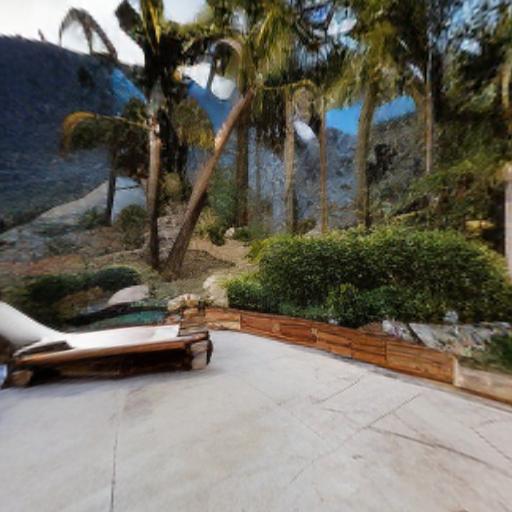} & \includegraphics[width=0.117\linewidth]{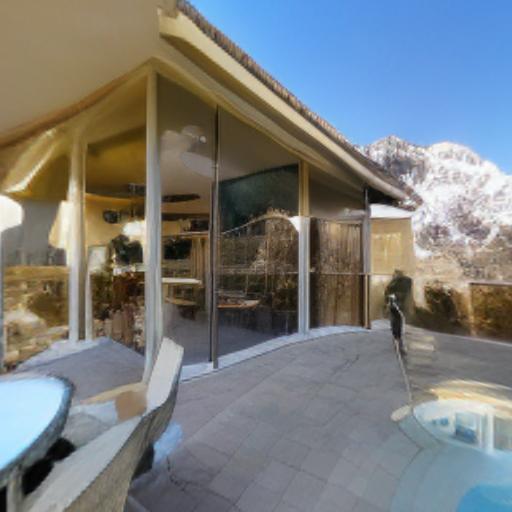} & \includegraphics[width=0.117\linewidth]{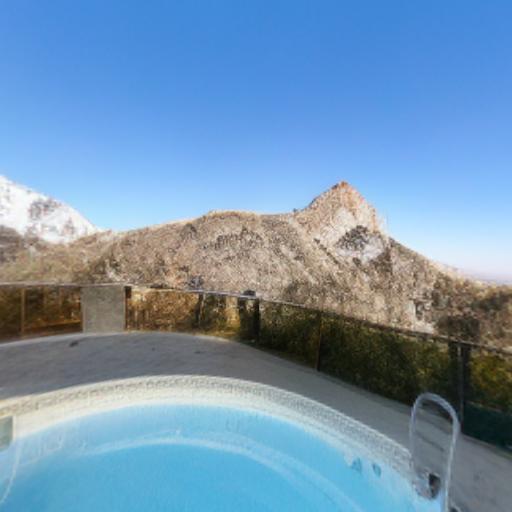} & \includegraphics[width=0.117\linewidth]{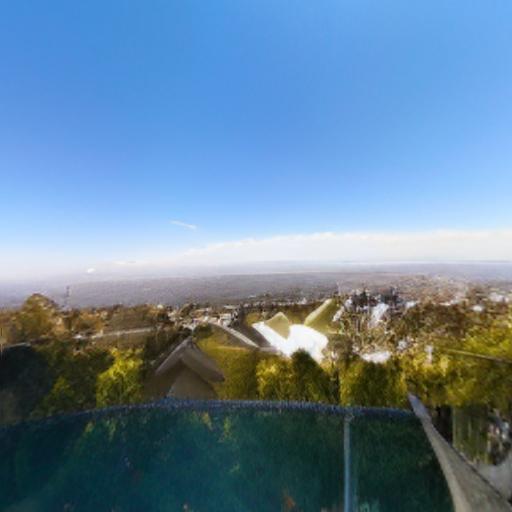} & 
         \includegraphics[width=0.117\linewidth]{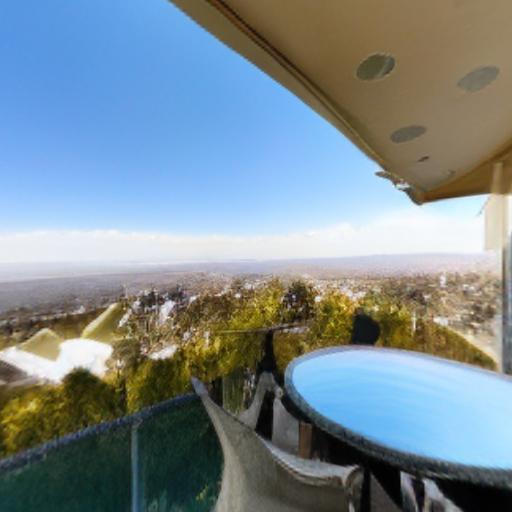} \\
         \\

         \multicolumn{8}{c}{\emph{``the inside of a garage''}} \\ \multicolumn{4}{c}{\includegraphics[width=0.48\linewidth]{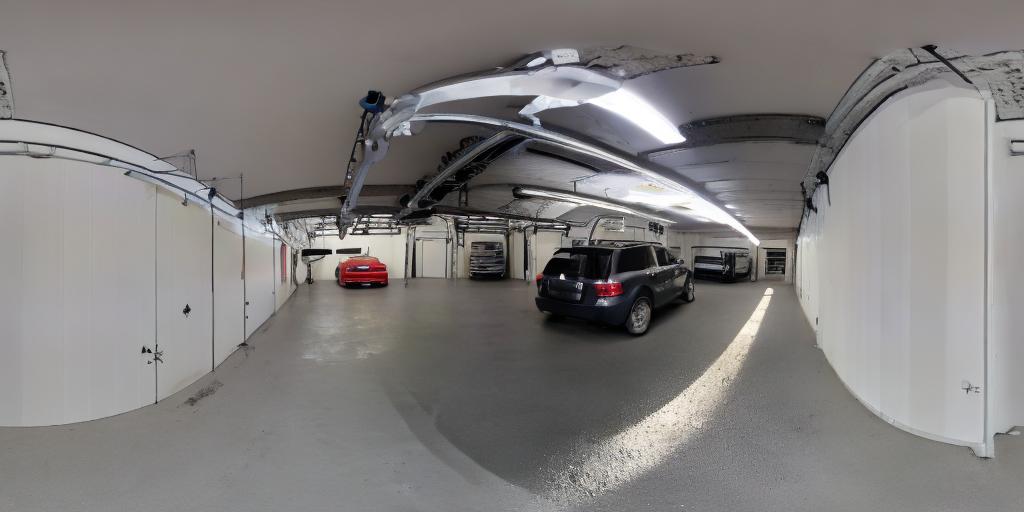}} & \multicolumn{4}{c}{\includegraphics[width=0.48\linewidth]{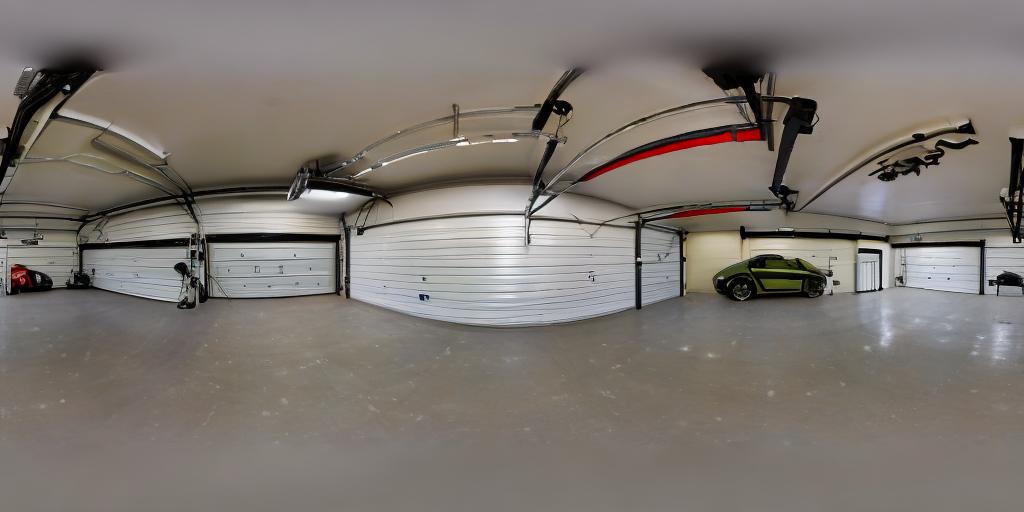}} \\
         \includegraphics[width=0.117\linewidth]{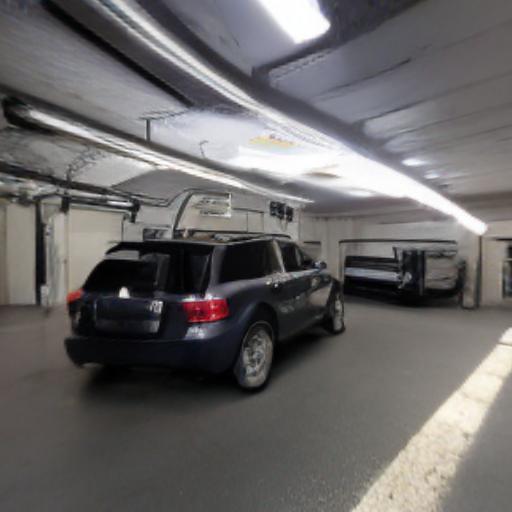} & \includegraphics[width=0.117\linewidth]{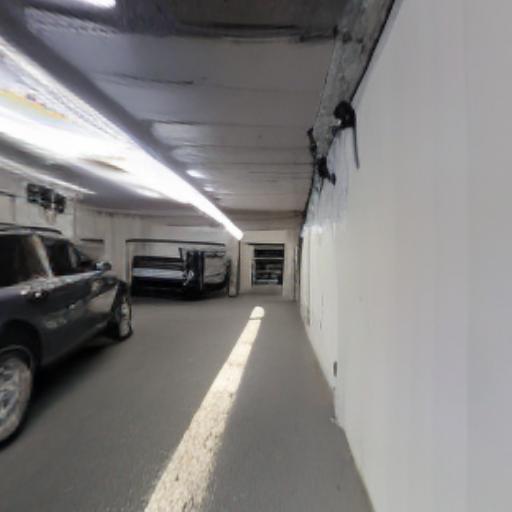} & \includegraphics[width=0.117\linewidth]{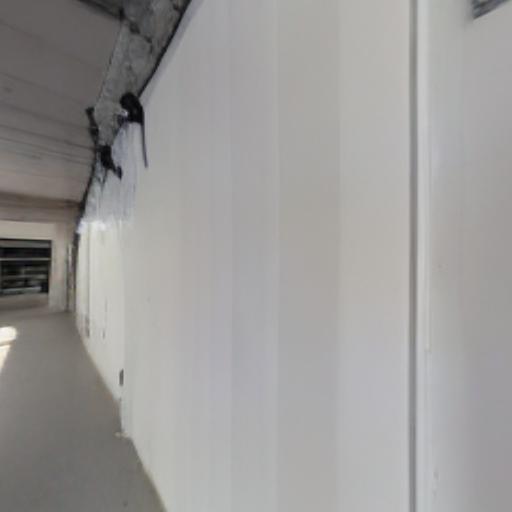} & \includegraphics[width=0.117\linewidth]{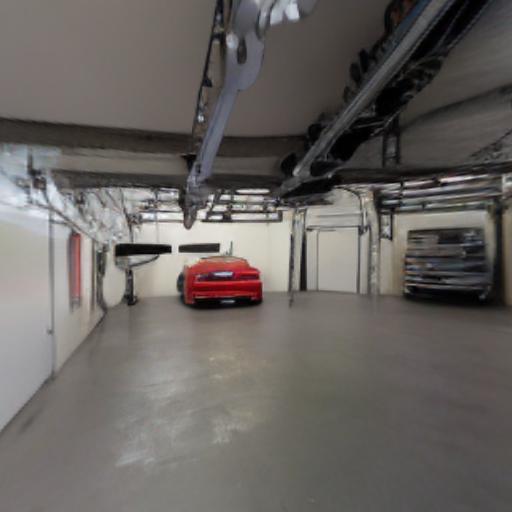} & \includegraphics[width=0.117\linewidth]{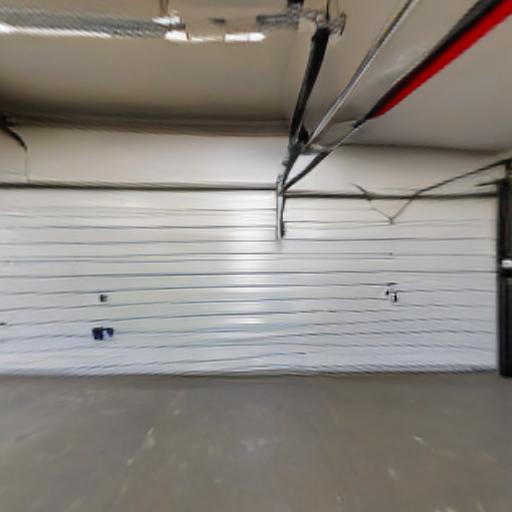} & \includegraphics[width=0.117\linewidth]{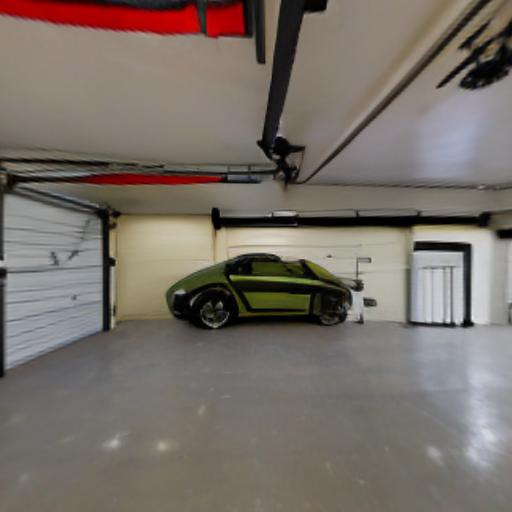} & \includegraphics[width=0.117\linewidth]{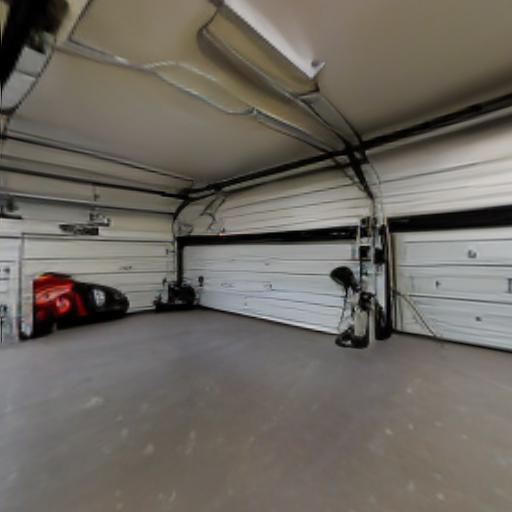} & 
         \includegraphics[width=0.117\linewidth]{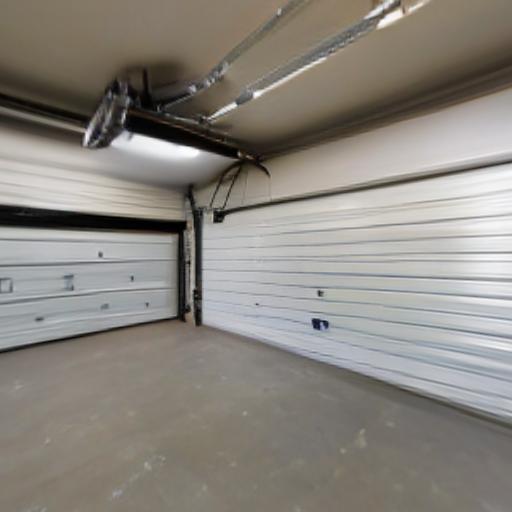} \\
         \\

         \multicolumn{8}{c}{\emph{``a hallway in a luxury home''}} \\
         \multicolumn{4}{c}{\includegraphics[width=0.48\linewidth]{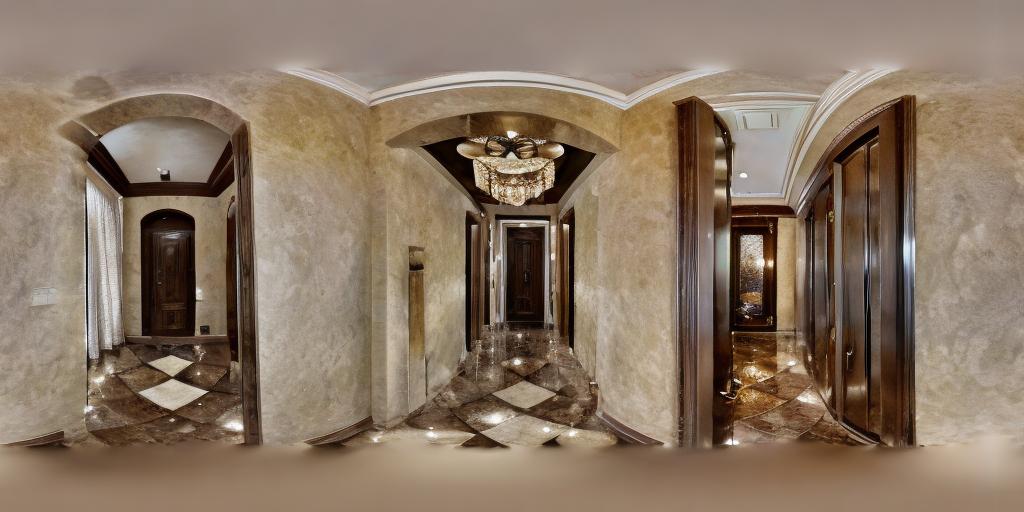}} & \multicolumn{4}{c}{\includegraphics[width=0.48\linewidth]{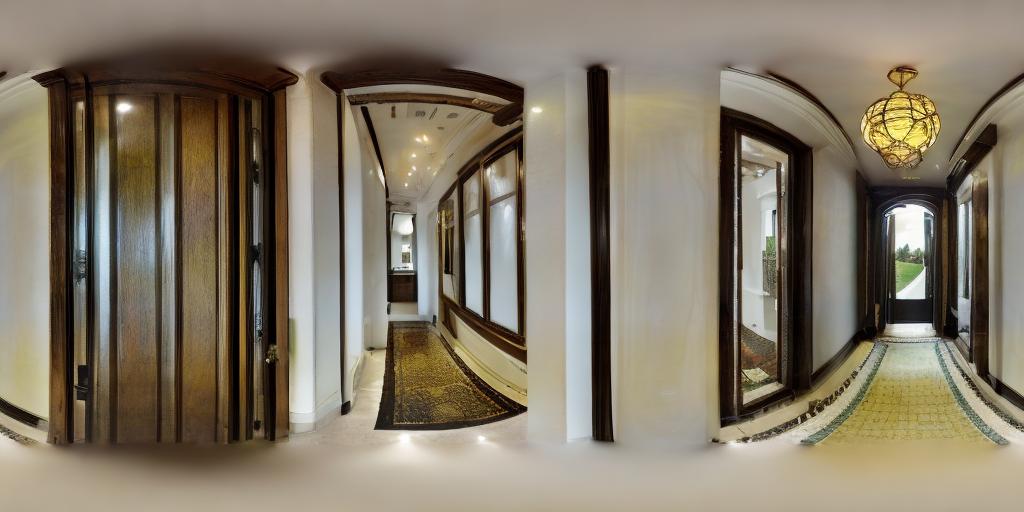}} \\
         \includegraphics[width=0.117\linewidth]{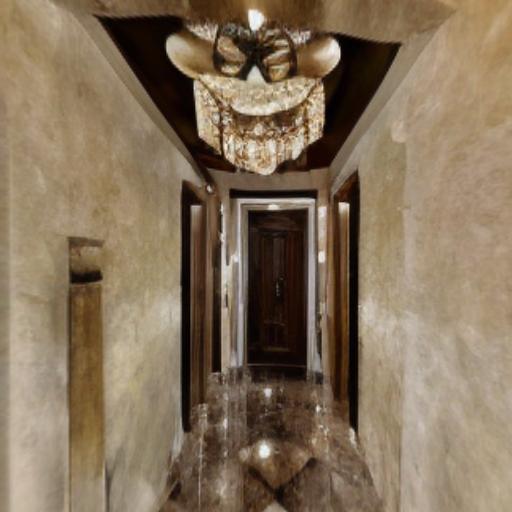} & \includegraphics[width=0.117\linewidth]{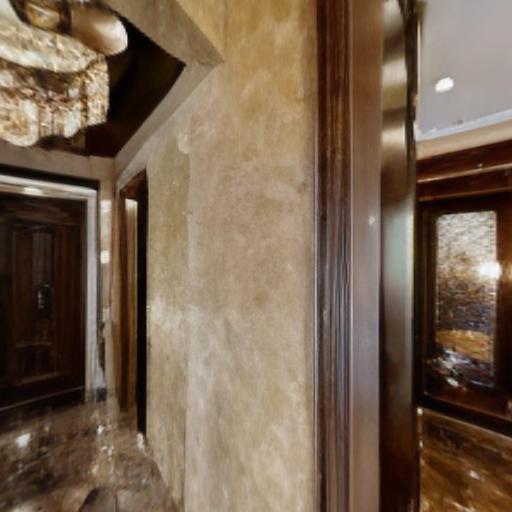} & \includegraphics[width=0.117\linewidth]{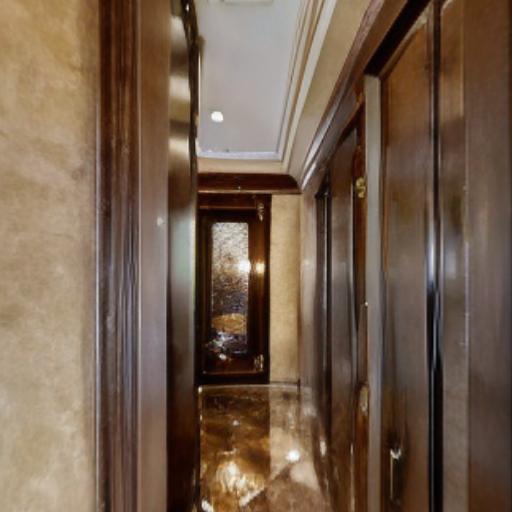} & \includegraphics[width=0.117\linewidth]{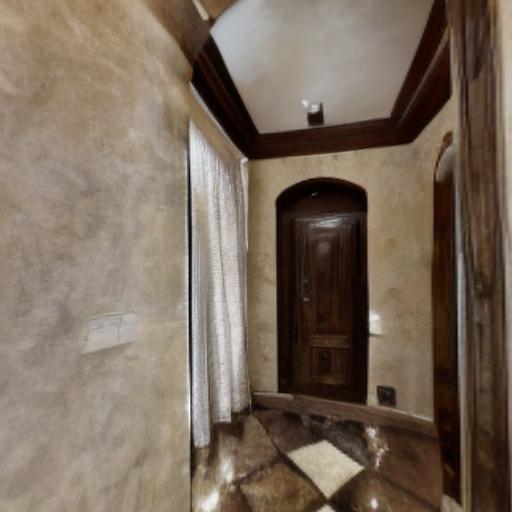} & \includegraphics[width=0.117\linewidth]{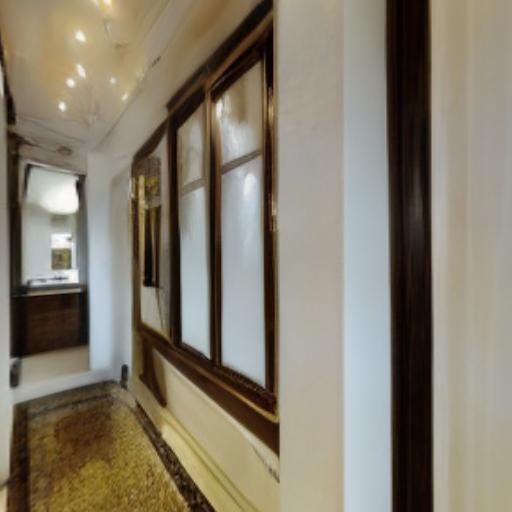} & \includegraphics[width=0.117\linewidth]{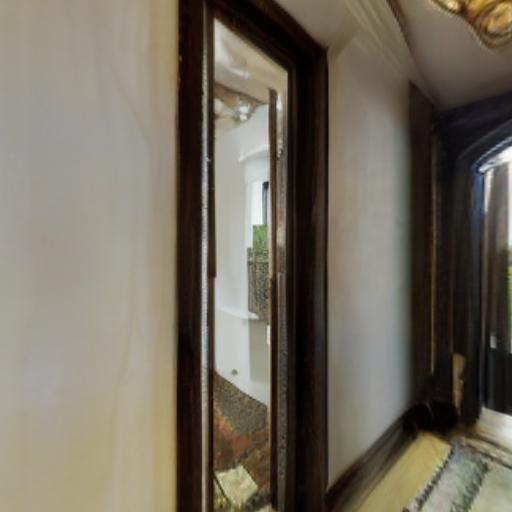} & \includegraphics[width=0.117\linewidth]{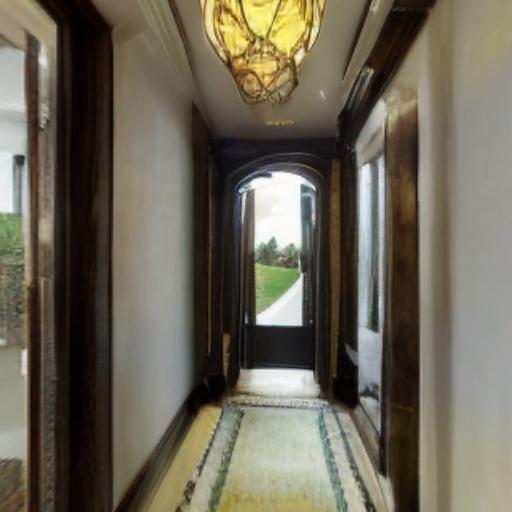} & 
         \includegraphics[width=0.117\linewidth]{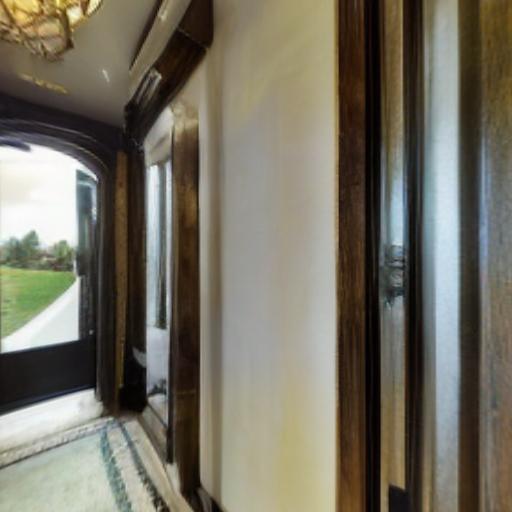} \\
         \\
         
         \multicolumn{4}{c}{{UniPano (Ours)}} & \multicolumn{4}{c}{{PanFusion}~\citep{panfusion}}
    \end{tabular}
    \caption{Additional qualitative comparisons.}
    \label{fig:add-qualitative-1}
\end{figure*}

\section{Additional Qualitative Comparisons}\label{sec:add-qualitative-exp}
In addition to the qualitative results in~\cref{sec:main-res}, we showcase more qualitative comparisons in~\cref{fig:add-qualitative,fig:add-qualitative-1}. We randomly sample 4 horizontal perspective views below each generated panoramic image.
One may also use panorama viewer (\eg \citep{renderer,carenderer}) to freely navigate the panoramas.

\section{Higher-resolution Panorama Generation}\label{sec:add-high-res}

\subsection{Implementation Details}
The setup for our higher-resolution generation experiments besides the base model is identical to~\cref{sec:exp-setup}. As the current SoTA PanFusion is not capable of generating $1024\times 2048$ panoramic images, we emphasize that our experiments serve primarily as illustrations rather than comparisons with current baseline models.
Another special note is that since Stable Diffusion 3 is based on transformer architectures, for which circular padding cannot be trivially applied and thus has been left out for our implementation.

\begin{figure*}[p]
    \centering
    \setlength{\tabcolsep}{1pt}
    \def\arraystretch{0.3}
    \begin{tabular}{cccccccc}
         
         \multicolumn{4}{c}{\emph{``a room with a swimming pool''}} & \multicolumn{4}{c}{\emph{``a room with marble floors''}} \\ \multicolumn{4}{c}{\includegraphics[width=0.48\linewidth]{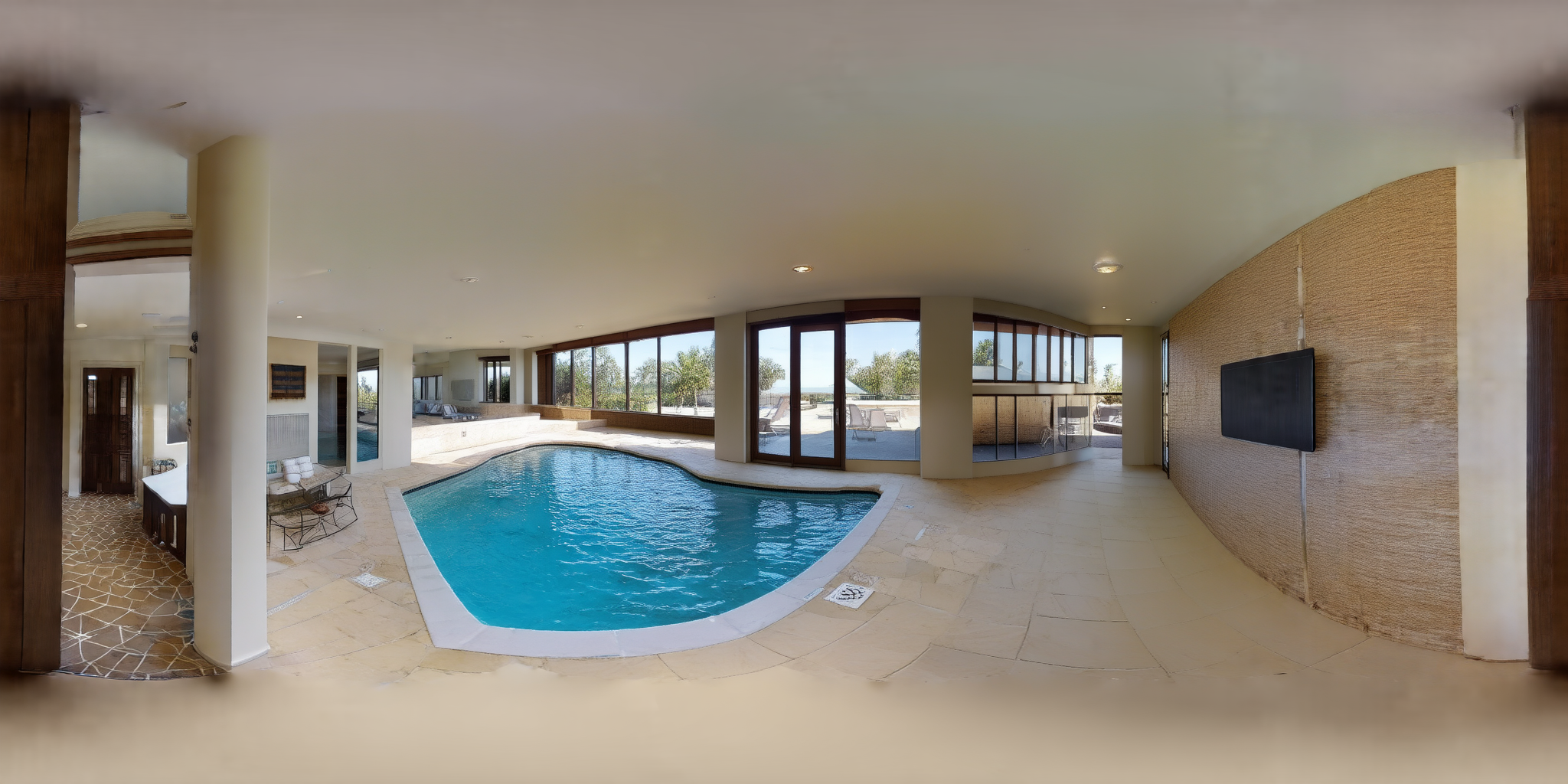}} & \multicolumn{4}{c}{\includegraphics[width=0.48\linewidth]{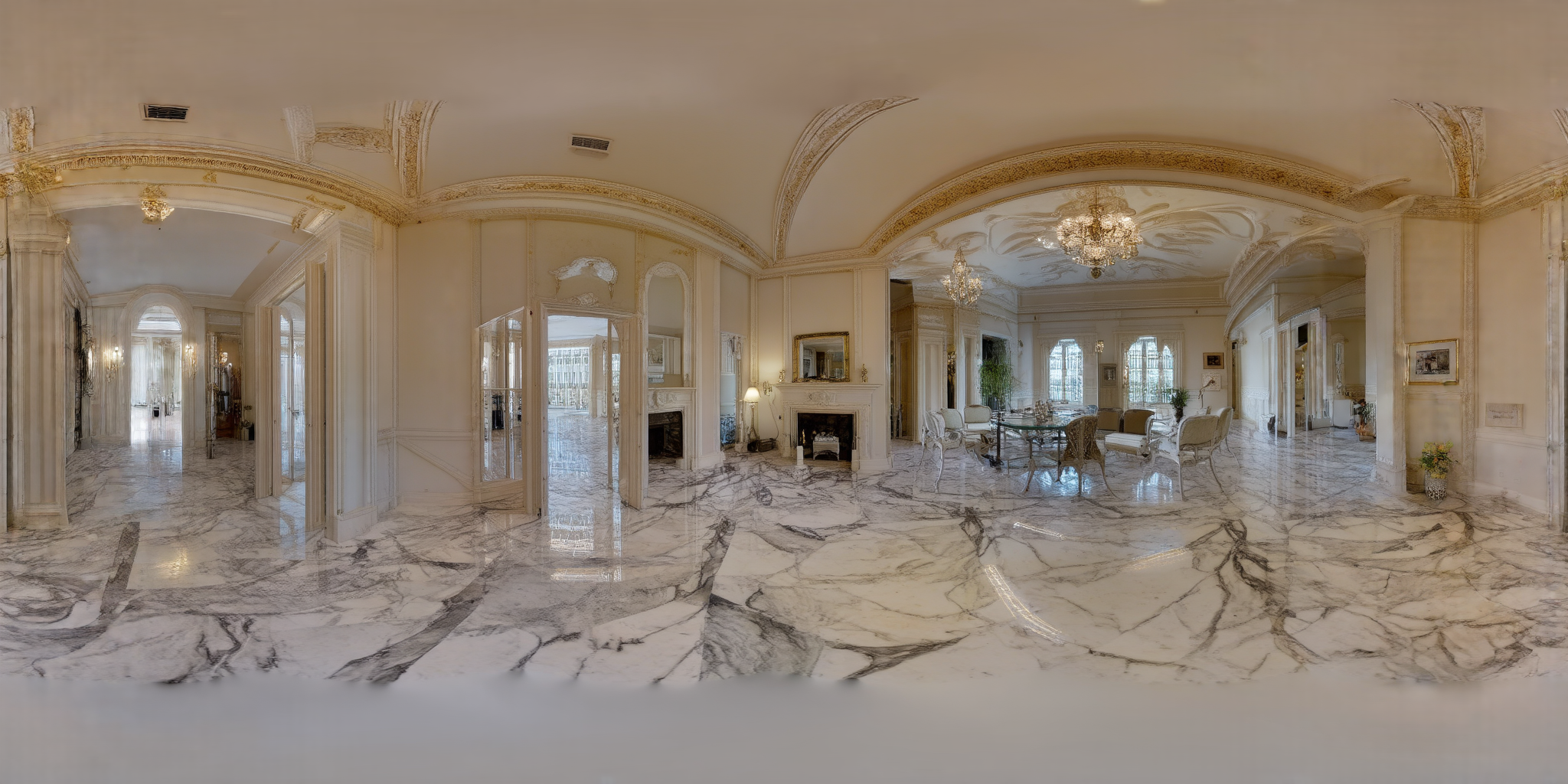}} \\
         \includegraphics[width=0.117\linewidth]{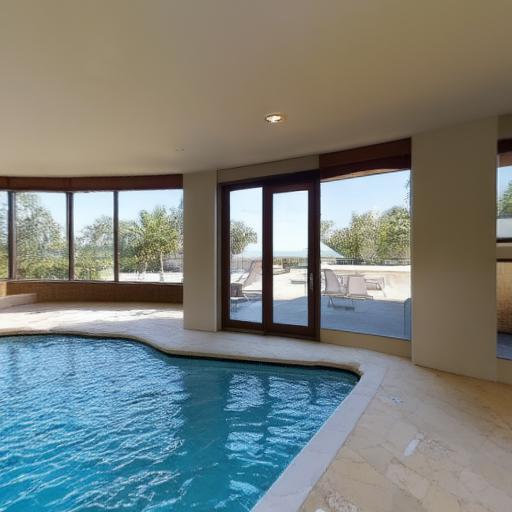} & \includegraphics[width=0.117\linewidth]{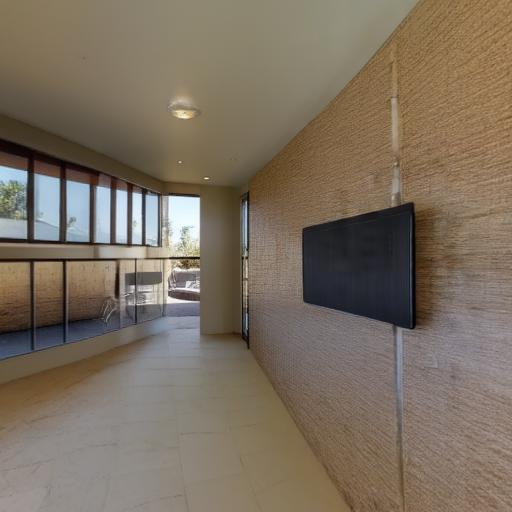} & \includegraphics[width=0.117\linewidth]{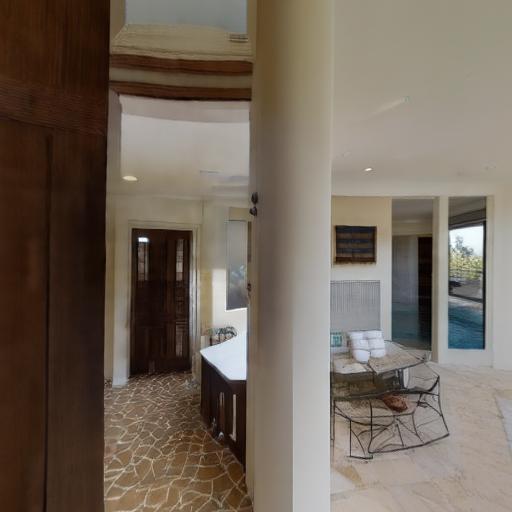} & \includegraphics[width=0.117\linewidth]{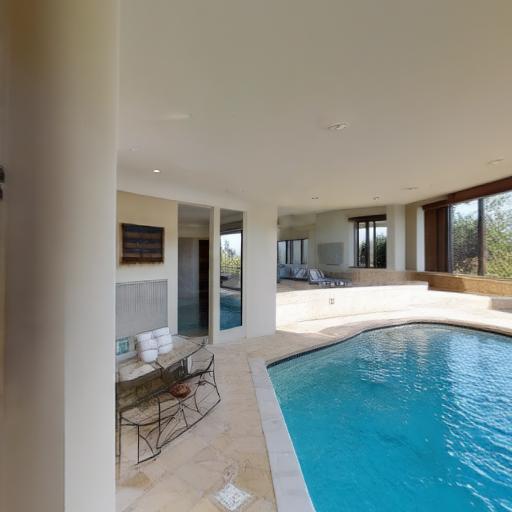} & \includegraphics[width=0.117\linewidth]{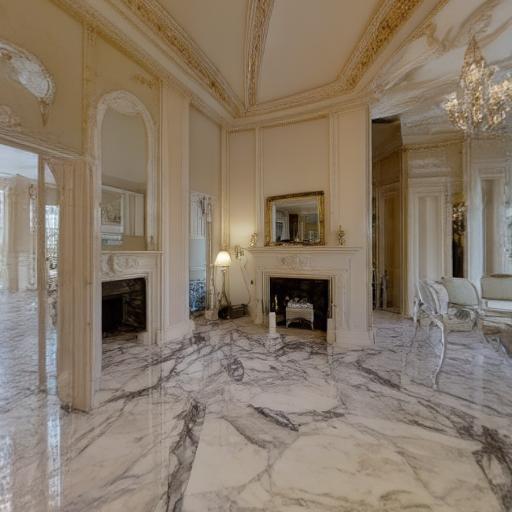} & \includegraphics[width=0.117\linewidth]{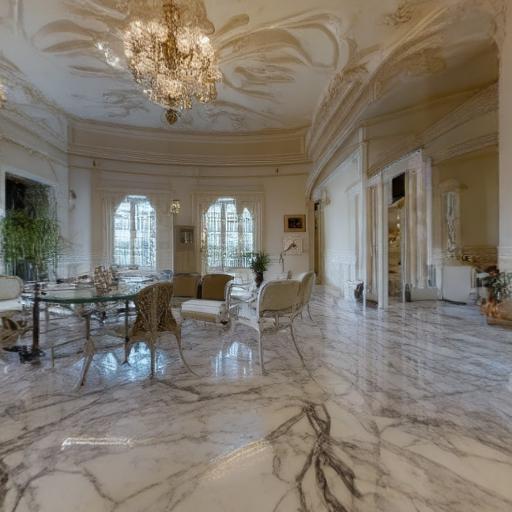} & \includegraphics[width=0.117\linewidth]{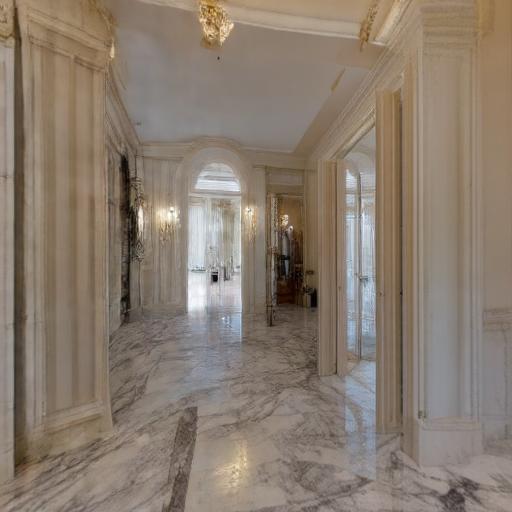} & 
         \includegraphics[width=0.117\linewidth]{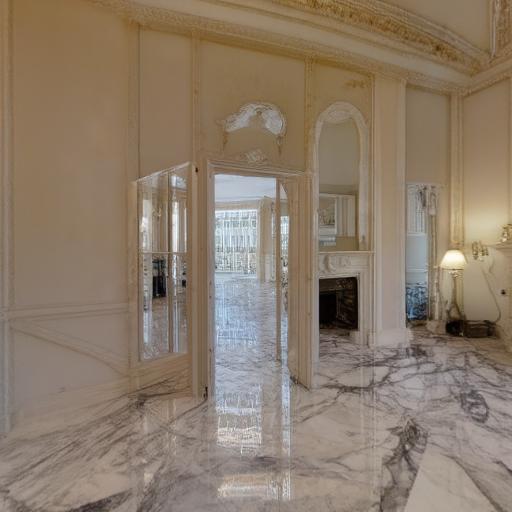} \\
         \\

         \multicolumn{4}{c}{\emph{``the inside of a home''}} & \multicolumn{4}{c}{\emph{``a garage with a car in it''}} \\ \multicolumn{4}{c}{\includegraphics[width=0.48\linewidth]{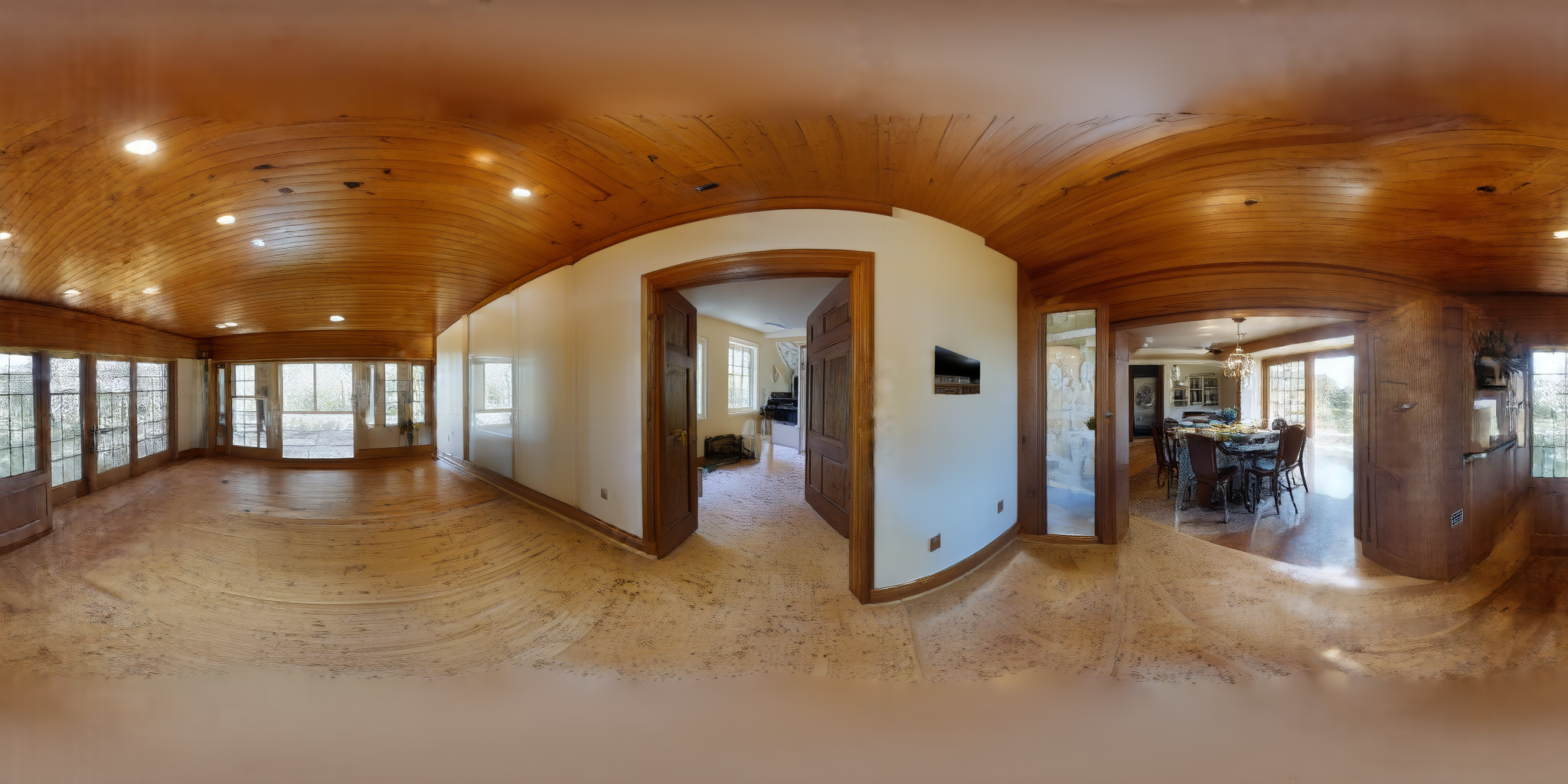}} & \multicolumn{4}{c}{\includegraphics[width=0.48\linewidth]{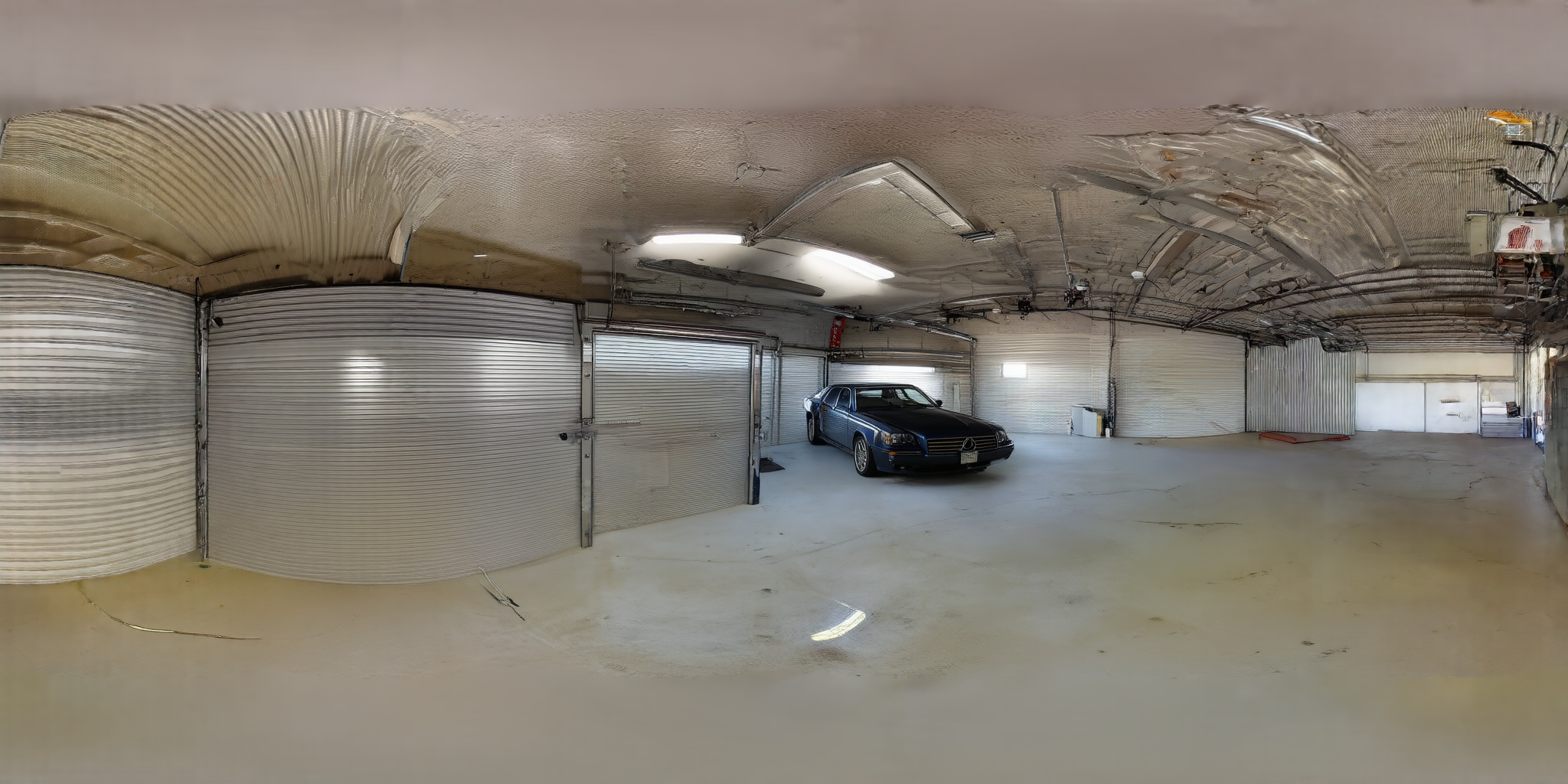}} \\
         \includegraphics[width=0.117\linewidth]{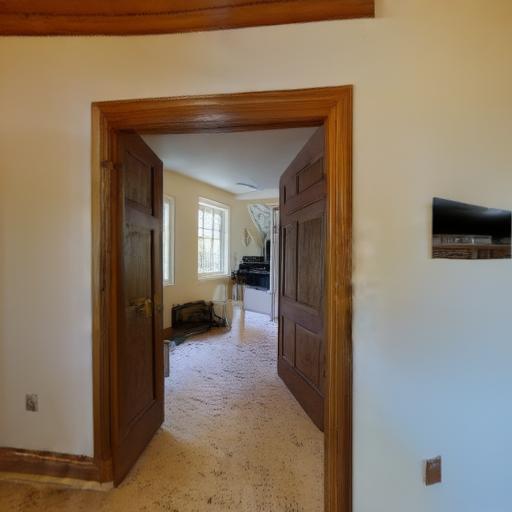} & \includegraphics[width=0.117\linewidth]{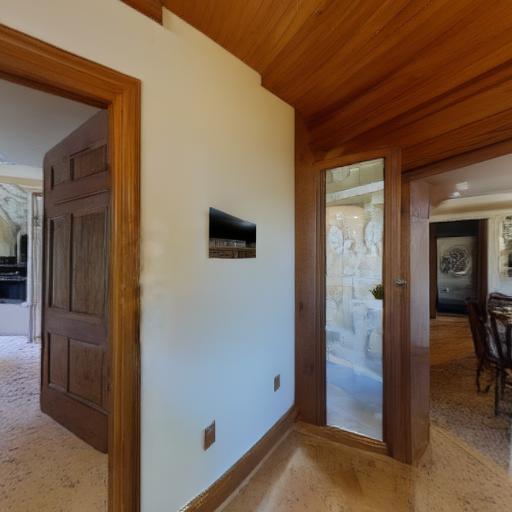} & \includegraphics[width=0.117\linewidth]{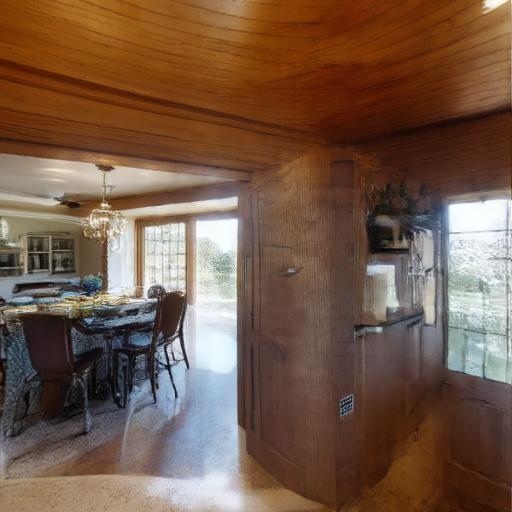} & 
         \includegraphics[width=0.117\linewidth]{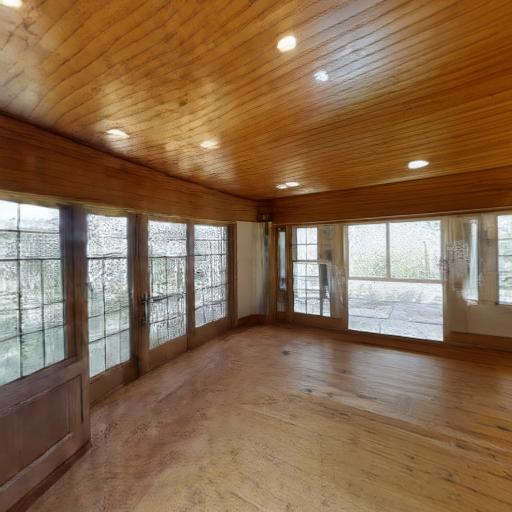} & \includegraphics[width=0.117\linewidth]{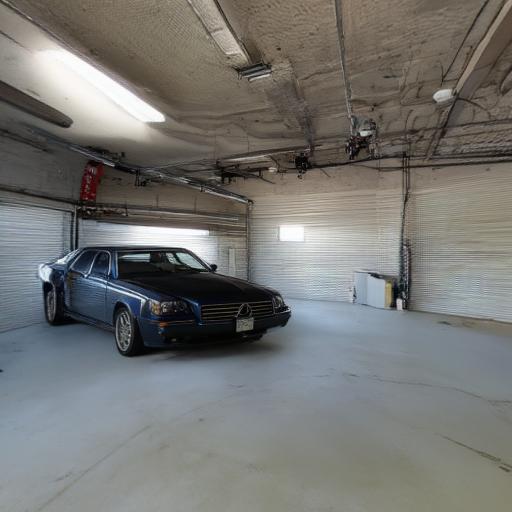} & \includegraphics[width=0.117\linewidth]{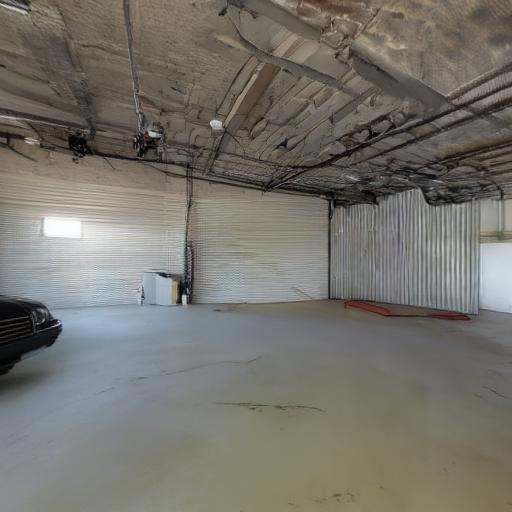} & \includegraphics[width=0.117\linewidth]{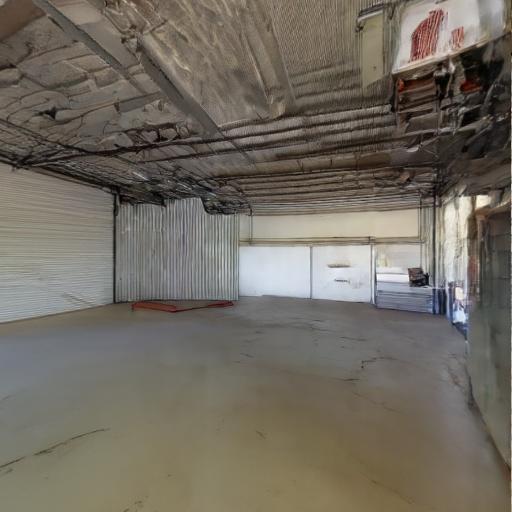} & 
         \includegraphics[width=0.117\linewidth]{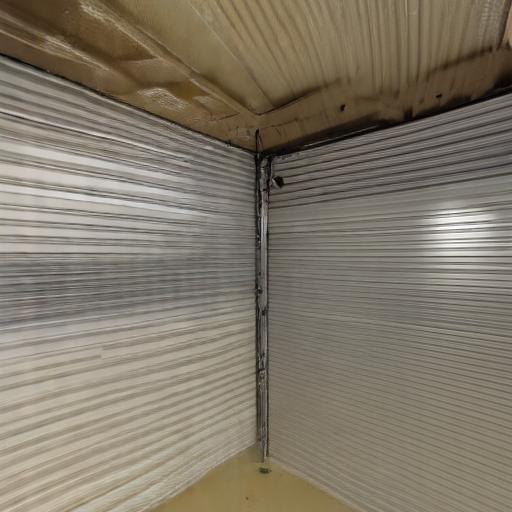} \\
         % \\

         % \multicolumn{4}{c}{\emph{``a large pantry with shelves''}} & \multicolumn{4}{c}{\emph{``a living room with a fireplace''}} \\ \multicolumn{4}{c}{\includegraphics[width=0.48\linewidth]{figs/high_resolution/a_large_pantry_with_shelves/unipano.png}} & \multicolumn{4}{c}{\includegraphics[width=0.48\linewidth]{figs/high_resolution/a_living_room_with_a_fireplace/unipano.png}} \\
         % \includegraphics[width=0.117\linewidth]{figs/high_resolution/a_large_pantry_with_shelves/unipano_pers_0.jpg} & \includegraphics[width=0.117\linewidth]{figs/high_resolution/a_large_pantry_with_shelves/unipano_pers_1.jpg} & \includegraphics[width=0.117\linewidth]{figs/high_resolution/a_large_pantry_with_shelves/unipano_pers_2.jpg} & 
         % \includegraphics[width=0.117\linewidth]{figs/high_resolution/a_large_pantry_with_shelves/unipano_pers_3.jpg} & \includegraphics[width=0.117\linewidth]{figs/high_resolution/a_living_room_with_a_fireplace/unipano_pers_0.jpg} & \includegraphics[width=0.117\linewidth]{figs/high_resolution/a_living_room_with_a_fireplace/unipano_pers_1.jpg} & \includegraphics[width=0.117\linewidth]{figs/high_resolution/a_living_room_with_a_fireplace/unipano_pers_2.jpg} & 
         % \includegraphics[width=0.117\linewidth]{figs/high_resolution/a_living_room_with_a_fireplace/unipano_pers_3.jpg}
    \end{tabular}
    \caption{Additional high-resolution ($1024\times 2024$) results. Note that all results are generated using UniPano based on Stable Diffusion 3.}
    \label{fig:highres-in-dist}
\end{figure*}

% \begin{figure*}[!th]
%     \centering
%     \setlength{\tabcolsep}{1pt}
%     \def\arraystretch{0.3}
%     \begin{tabular}{c}
%            \emph{``A peaceful coastal village at sunrise, with fishing boats docked along the quiet harbor.''} \\
%         \includegraphics[width=0.99\linewidth]{figs/high_resolution/highres_example1.png} \\
%         \\
%         \\

%         \emph{``A bustling tech conference in Silicon Valley, with innovators discussing the latest advancements in technology.''} \\
%         \includegraphics[width=0.99\linewidth]{figs/high_resolution/highres_example2.png}
%     \end{tabular}
%     \caption{{Additional high-resolution results for out-of-distribution prompts}.}
%     \label{fig:highres1-ood}
% \end{figure*}

\begin{figure*}[!th]
    \centering
    \setlength{\tabcolsep}{1pt}
    \def\arraystretch{0.3}
    \begin{tabular}{c}
           \emph{``Exploring the historic streets of Prague, with its charming architecture, cobblestone alleys, and medieval ambiance.''} \\
        \includegraphics[width=0.99\linewidth]{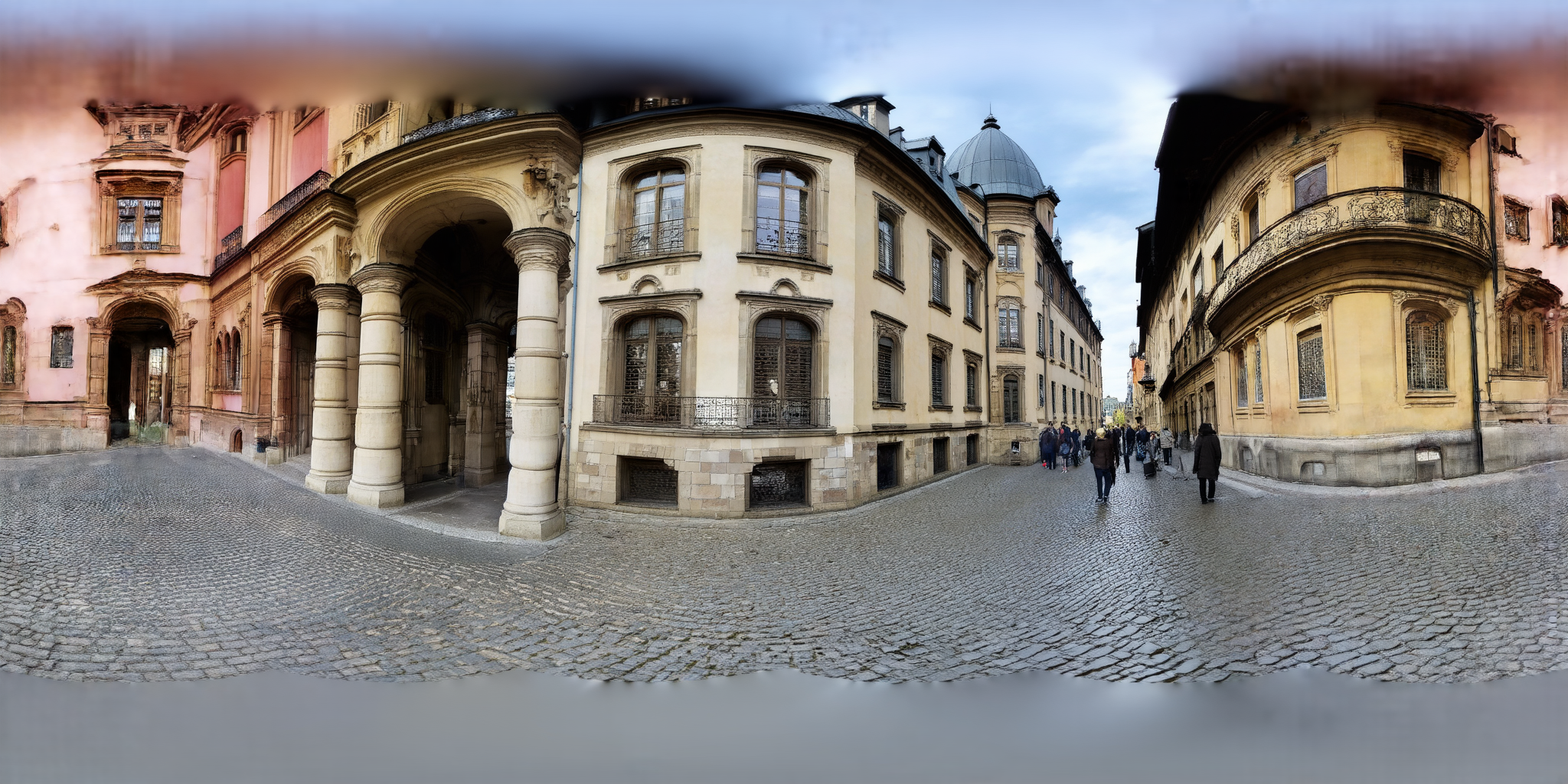} \\
        \\
        \\

        \emph{``A traditional Italian trattoria, where locals gather for hearty meals, laughter, and the warmth of shared conversation.''} \\
        \includegraphics[width=0.99\linewidth]{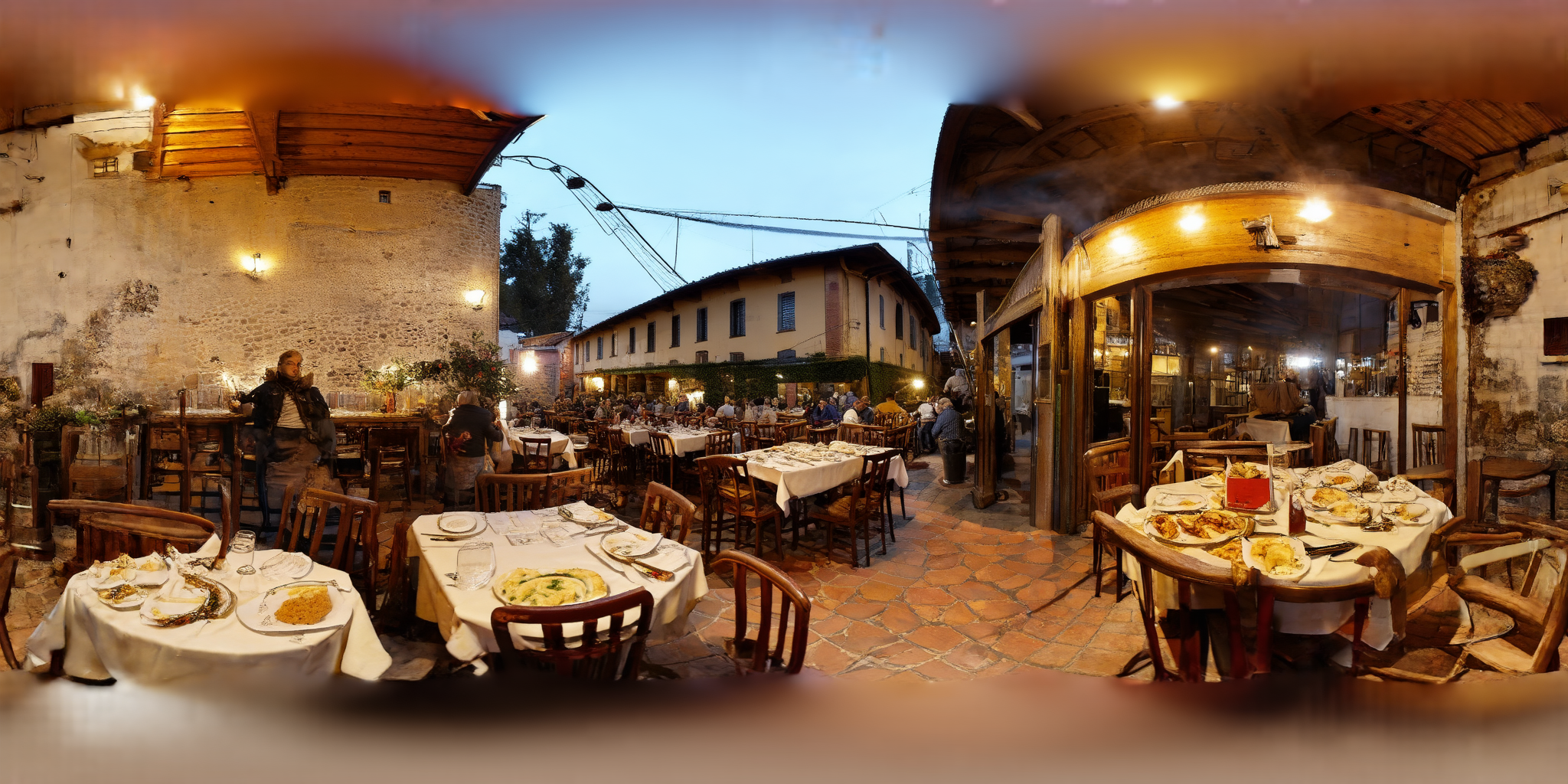}
    \end{tabular}
    \caption{{Additional high-resolution results for out-of-distribution prompts}.}
    \label{fig:highres2-ood}
\end{figure*}

\begin{figure*}[!th]
    \centering
    \setlength{\tabcolsep}{1pt}
    \def\arraystretch{0.3}
    \begin{tabular}{c}
           \emph{``Alpine village, snow-covered rooftops, nestled between majestic peaks—a picture-perfect scene of winter tranquility.''} \\
        \includegraphics[width=0.99\linewidth]{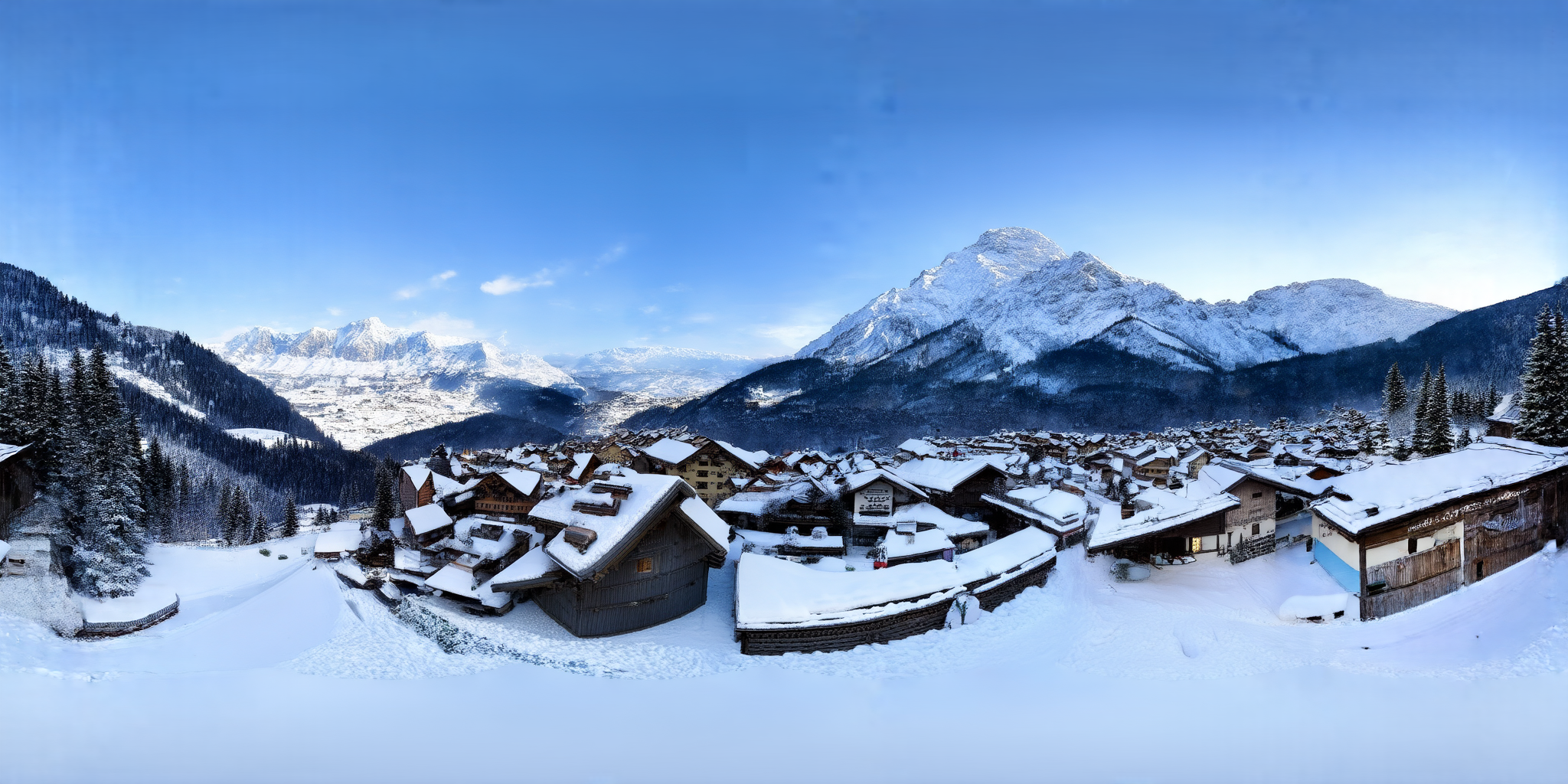} \\
        \\
        \\

        \emph{``Exploring an abandoned underwater city, where sunken buildings are now home to schools of bioluminescent fish.''} \\
        \includegraphics[width=0.99\linewidth]{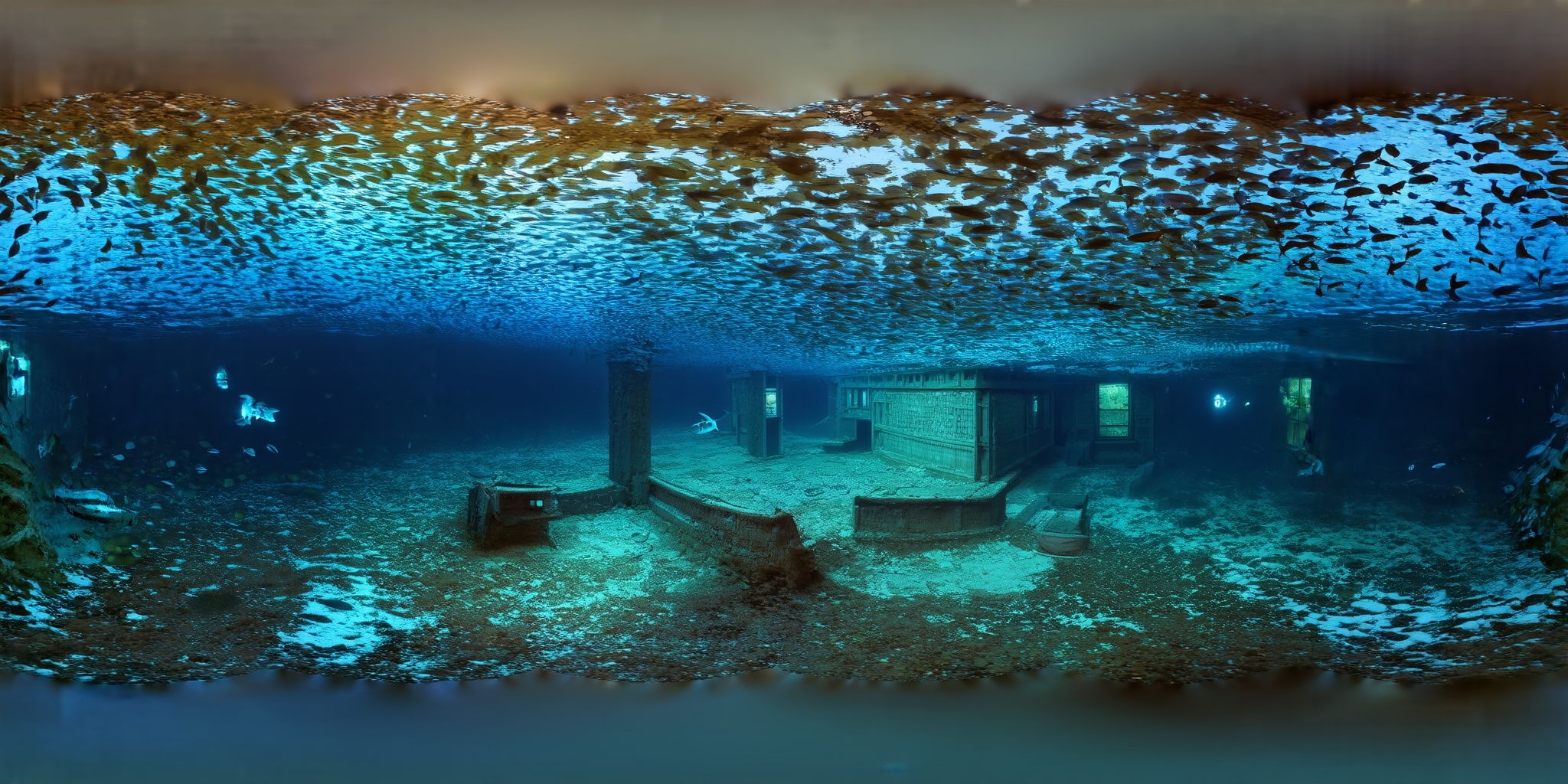}
    \end{tabular}
    \caption{{Additional high-resolution results for out-of-distribution prompts}.}
    \label{fig:highres3-ood}
\end{figure*}

% \begin{figure*}[!th]
%     \centering
%     \setlength{\tabcolsep}{1pt}
%     \def\arraystretch{0.3}
%     \begin{tabular}{c}
%            \emph{``Futuristic cityscape, skyscrapers piercing the sky, neon lights painting the streets—an electric dreamscape alive with } \\ \emph{urban energy.''} \\
%         \includegraphics[width=0.99\linewidth]{figs/high_resolution/highres_example7.png} \\
%         \\
%         \\

%         \emph{``On the surface of a distant planet, a landscape of alien rock formations and swirling, multicolored gases.''} \\
%         \includegraphics[width=0.99\linewidth]{figs/high_resolution/highres_example8.png}
%     \end{tabular}
%     \caption{{Additional high-resolution results for out-of-distribution prompts}.}
%     \label{fig:highres4-ood}
% \end{figure*}

% \begin{figure*}[!th]
%     \centering
%     \setlength{\tabcolsep}{1pt}
%     \def\arraystretch{0.3}
%     \begin{tabular}{c}
%            \emph{``Desert oasis, palm trees surrounding a pristine pool, an emerald jewel amid golden sands—an Arabian mirage.''} \\
%         \includegraphics[width=0.99\linewidth]{figs/high_resolution/highres_example9.png} \\
%         \\
%         \\

%         \emph{``Coastal cliffside, waves crashing on rugged rocks, seagulls soaring in the salty breeze—a dramatic meeting of land and sea.''} \\
%         \includegraphics[width=0.99\linewidth]{figs/high_resolution/highres_example10.png}
%     \end{tabular}
%     \caption{{Additional high-resolution results for out-of-distribution prompts}.}
%     \label{fig:highres5-ood}
% \end{figure*}

\begin{figure*}[!th]
    \centering
    \setlength{\tabcolsep}{1pt}
    \def\arraystretch{0.3}
    \begin{tabular}{c}
           \emph{``Rain-soaked city streets, glistening reflections.''} \\
        \includegraphics[width=0.99\linewidth]{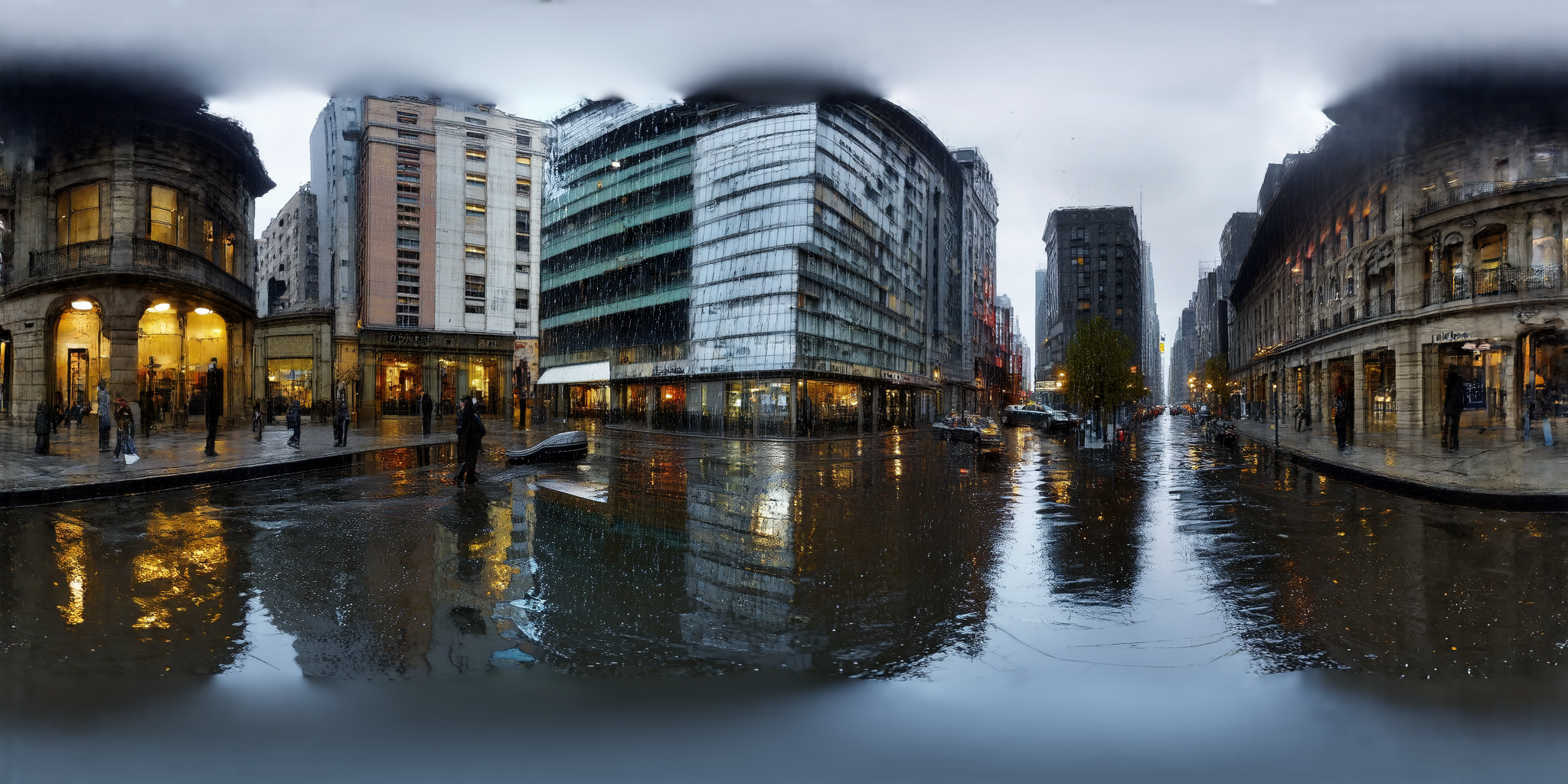} \\
        \\
        \\

        \emph{``Cobblestone alley, historic architecture bathed in soft morning light.''} \\
        \includegraphics[width=0.99\linewidth]{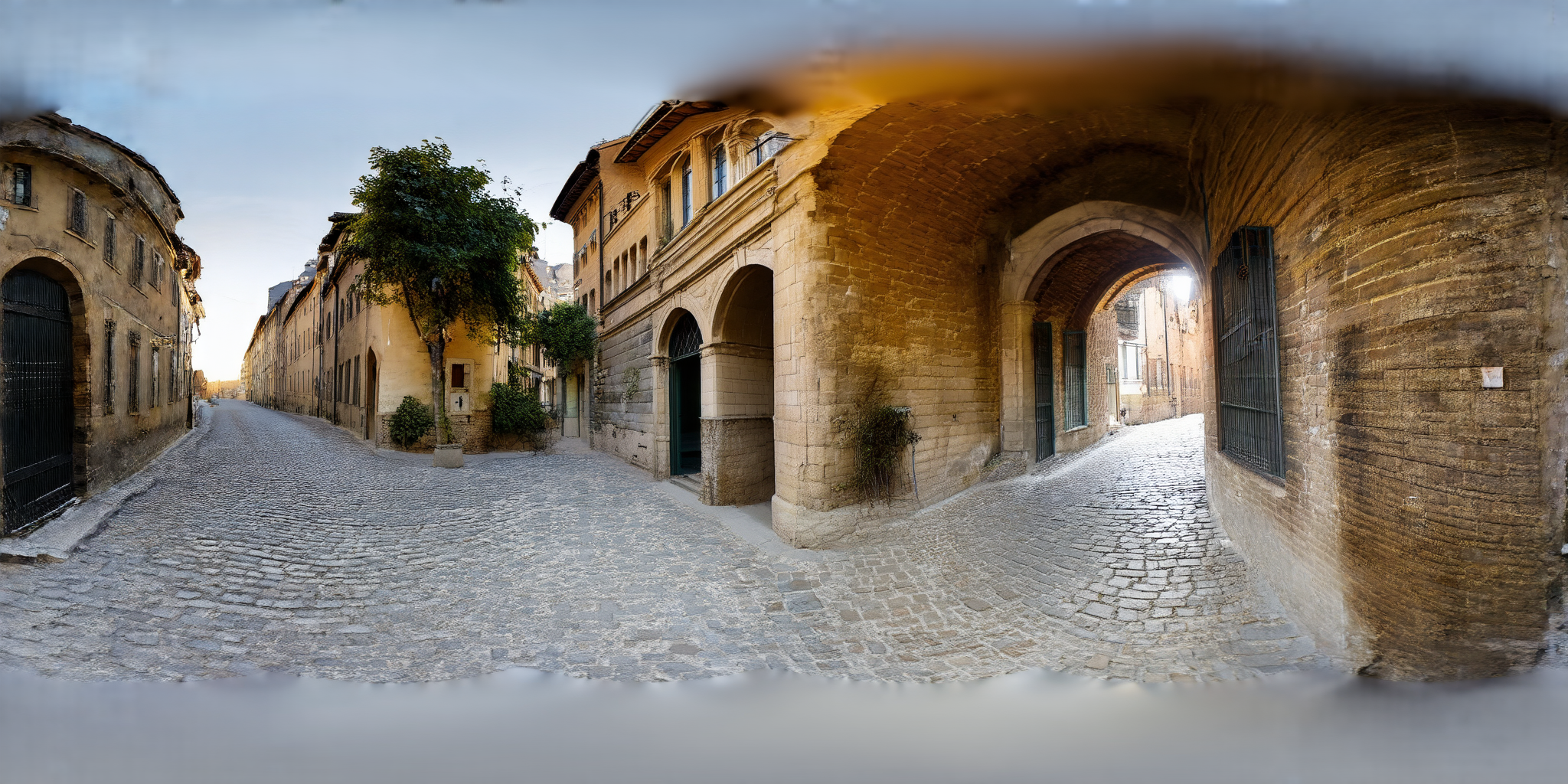}
    \end{tabular}
    \caption{{Additional high-resolution results for out-of-distribution prompts}.}
    \label{fig:highres6-ood}
\end{figure*}

\begin{figure*}[!th]
    \centering
    \setlength{\tabcolsep}{1pt}
    \def\arraystretch{0.3}
    \begin{tabular}{c}
           \emph{``On a distant planet's surface, towering crystalline structures rise against an alien sky. The landscape is surreal, with} \\ \emph{bioluminescent flora casting an otherworldly glow. Strange creatures move gracefully through the phosphorescent mist,}\\ \emph{creating an ethereal scene that defies earthly imagination.''} \\
        \includegraphics[width=0.99\linewidth]{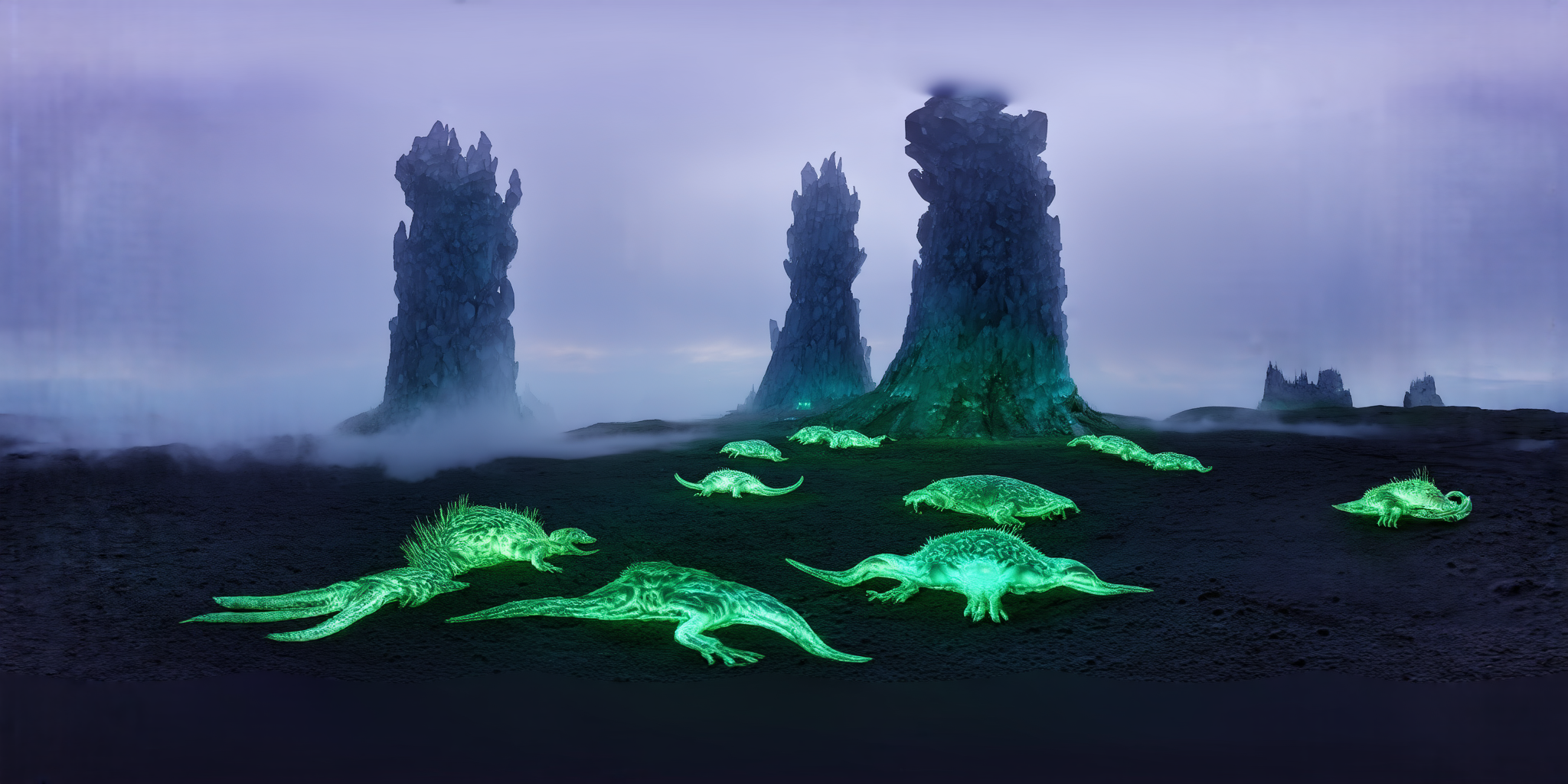} \\
        \\
        \\

        \emph{``Amidst the bustling energy of a busy market, vendors peddle their wares with animated fervor. A kaleidoscope of colors,}\\ \emph{from fresh produce to woven textiles, creates a vibrant tapestry. The air is thick with the mingling scents of spices, street} \\ \emph{food, and the lively chatter of buyers and sellers.''} \\
        \includegraphics[width=0.99\linewidth]{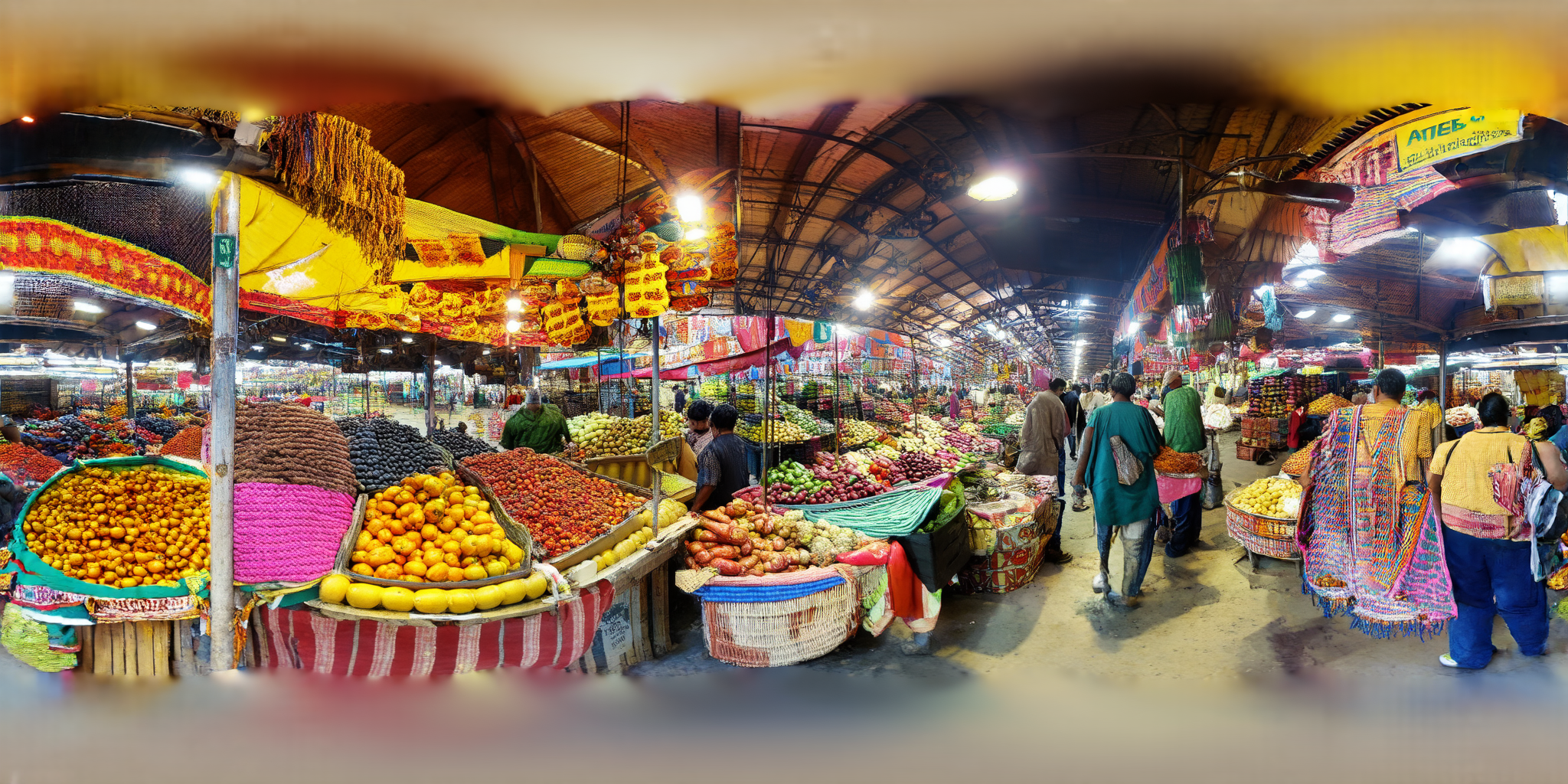}
    \end{tabular}
    \caption{{Additional high-resolution results with complex prompts}.}
    \label{fig:highres1-complex}
\end{figure*}

% \begin{figure*}[!th]
%     \centering
%     \setlength{\tabcolsep}{1pt}
%     \def\arraystretch{0.3}
%     \begin{tabular}{c}
%            \emph{``In a quaint coastal village, weathered cottages line the shore, their pastel hues blending with the colors of the sea. Fishing} \\ \emph{boats bob gently in the harbor, and the air is tinged with the scent of saltwater and freshly caught fish. Seagulls circle } \\ \emph{overhead, adding to the maritime chorus.''} \\
%         \includegraphics[width=0.99\linewidth]{figs/high_resolution/highres_ood_example3.png} \\
%         \\
%         \\

%         \emph{``Nestled in a canyon, a pueblo village stands against the red earth. Adobe homes with turquoise accents blend seamlessly } \\ \emph{into the landscape. The sun sets, casting a warm glow on the cliffs, and the rhythmic beats of traditional drums resonate} \\ \emph{ in the stillness, echoing ancient tales.''} \\
%         \includegraphics[width=0.99\linewidth]{figs/high_resolution/highres_ood_example4.png}
%     \end{tabular}
%     \caption{{Additional high-resolution results with complex prompts}.}
%     \label{fig:highres2-complex}
% \end{figure*}

\subsection{Additional High-resolution Results}
We present the qualitative results for scaling UniPano to generate $1024\times 2048$ panoramic images. We provide more qualitative results in~\cref{fig:highres-in-dist}. To illustrate the power of implementing UniPano on a more powerful base model, we showcase the results with out-of-distribution text prompts in~\cref{fig:highres2-ood,fig:highres3-ood,fig:highres6-ood} and with extremely long and complex text prompts in~\cref{fig:highres1-complex}. We refer the reader to the semantic class distribution of Matterport3D in~\citep[Fig 5]{Matterport3D} for the definition of in- and out-of-distribution.

%% file: main.bbl
\begin{thebibliography}{59}
\providecommand{\natexlab}[1]{#1}
\providecommand{\url}[1]{\texttt{#1}}
\expandafter\ifx\csname urlstyle\endcsname\relax
  \providecommand{\doi}[1]{doi: #1}\else
  \providecommand{\doi}{doi: \begingroup \urlstyle{rm}\Url}\fi

\bibitem[car()]{carenderer}
Chief architect 360° panorama renderings.
\newblock \url{https://www.chiefarchitect.com/products/360-panorama-viewer}.
\newblock Accessed: 2025-03-03.

\bibitem[ren()]{renderer}
Online 360° panorama viewer vr.
\newblock \url{https://renderstuff.com/tools/360-panorama-web-viewer}.
\newblock Accessed: 2025-03-03.

\bibitem[Akimoto et~al.(2022)Akimoto, Matsuo, and Aoki]{akimoto2022diverse}
Naofumi Akimoto, Yuhi Matsuo, and Yoshimitsu Aoki.
\newblock Diverse plausible 360-degree image outpainting for efficient 3dcg background creation.
\newblock In \emph{CVPR}, 2022.

\bibitem[Bar-Tal et~al.(2023)Bar-Tal, Yariv, Lipman, and Dekel]{multidiffusion}
Omer Bar-Tal, Lior Yariv, Yaron Lipman, and Tali Dekel.
\newblock Multidiffusion: Fusing diffusion paths for controlled image generation.
\newblock In \emph{ICML}, 2023.

\bibitem[Chang et~al.(2017)Chang, Dai, Funkhouser, Halber, Niessner, Savva, Song, Zeng, and Zhang]{Matterport3D}
Angel Chang, Angela Dai, Thomas Funkhouser, Maciej Halber, Matthias Niessner, Manolis Savva, Shuran Song, Andy Zeng, and Yinda Zhang.
\newblock Matterport3d: Learning from rgb-d data in indoor environments.
\newblock In \emph{3DV}, 2017.

\bibitem[Chen et~al.(2023)Chen, Sun, Song, and Luo]{diffusiondet}
Shoufa Chen, Peize Sun, Yibing Song, and Ping Luo.
\newblock Diffusiondet: Diffusion model for object detection.
\newblock In \emph{ICCV}, 2023.

\bibitem[Chen et~al.(2022)Chen, Wang, and Liu]{chen2022text2light}
Zhaoxi Chen, Guangcong Wang, and Ziwei Liu.
\newblock Text2light: Zero-shot text-driven hdr panorama generation.
\newblock \emph{TOG}, 2022.

\bibitem[Chen et~al.(2024)Chen, Wang, Wang, and Liu]{chen2024text}
Zilong Chen, Feng Wang, Yikai Wang, and Huaping Liu.
\newblock Text-to-3d using gaussian splatting.
\newblock In \emph{CVPR}, 2024.

\bibitem[Esser et~al.(2024)Esser, Kulal, Blattmann, Entezari, M{\"u}ller, Saini, Levi, Lorenz, Sauer, Boesel, et~al.]{sd3}
Patrick Esser, Sumith Kulal, Andreas Blattmann, Rahim Entezari, Jonas M{\"u}ller, Harry Saini, Yam Levi, Dominik Lorenz, Axel Sauer, Frederic Boesel, et~al.
\newblock Scaling rectified flow transformers for high-resolution image synthesis.
\newblock In \emph{ICML}, 2024.

\bibitem[Fei et~al.(2024)Fei, Fan, Yu, Li, and Huang]{ditmoe}
Zhengcong Fei, Mingyuan Fan, Changqian Yu, Debang Li, and Junshi Huang.
\newblock Scaling diffusion transformers to 16 billion parameters.
\newblock \emph{arXiv preprint arXiv:2407.11633}, 2024.

\bibitem[Gal et~al.(2023)Gal, Alaluf, Atzmon, Patashnik, Bermano, Chechik, and Cohen-Or]{textualinv}
Rinon Gal, Yuval Alaluf, Yuval Atzmon, Or Patashnik, Amit~H Bermano, Gal Chechik, and Daniel Cohen-Or.
\newblock An image is worth one word: Personalizing text-to-image generation using textual inversion.
\newblock In \emph{ICLR}, 2023.

\bibitem[Hassani et~al.(2023)Hassani, Walton, Li, Li, and Shi]{nat}
Ali Hassani, Steven Walton, Jiachen Li, Shen Li, and Humphrey Shi.
\newblock Neighborhood attention transformer.
\newblock In \emph{CVPR}, 2023.

\bibitem[Ho and Salimans(2022)]{cfg}
Jonathan Ho and Tim Salimans.
\newblock Classifier-free diffusion guidance.
\newblock \emph{arXiv preprint arXiv:2207.12598}, 2022.

\bibitem[Ho et~al.(2020)Ho, Jain, and Abbeel]{ddpm}
Jonathan Ho, Ajay Jain, and Pieter Abbeel.
\newblock Denoising diffusion probabilistic models.
\newblock In \emph{NeurIPS}, 2020.

\bibitem[Hu et~al.(2022)Hu, Shen, Wallis, Allen-Zhu, Li, Wang, Wang, and Chen]{lora}
Edward~J Hu, Yelong Shen, Phillip Wallis, Zeyuan Allen-Zhu, Yuanzhi Li, Shean Wang, Lu Wang, and Weizhu Chen.
\newblock Lo{RA}: Low-rank adaptation of large language models.
\newblock In \emph{ICLR}, 2022.

\bibitem[Hu et~al.(2018)Hu, Shen, and Sun]{senet}
Jie Hu, Li Shen, and Gang Sun.
\newblock Squeeze-and-excitation networks.
\newblock In \emph{CVPR}, 2018.

\bibitem[Kalischek et~al.(2025)Kalischek, Oechsle, Manhardt, Henzler, Schindler, and Tombari]{cubediff}
Nikolai Kalischek, Michael Oechsle, Fabian Manhardt, Philipp Henzler, Konrad Schindler, and Federico Tombari.
\newblock Cubediff: Repurposing diffusion-based image models for panorama generation.
\newblock In \emph{ICLR}, 2025.

\bibitem[Ke et~al.(2024)Ke, Obukhov, Huang, Metzger, Daudt, and Schindler]{marigold}
Bingxin Ke, Anton Obukhov, Shengyu Huang, Nando Metzger, Rodrigo~Caye Daudt, and Konrad Schindler.
\newblock Repurposing diffusion-based image generators for monocular depth estimation.
\newblock In \emph{CVPR}, 2024.

\bibitem[Li and Bansal(2023)]{panogen}
Jialu Li and Mohit Bansal.
\newblock Panogen: Text-conditioned panoramic environment generation for vision-and-language navigation.
\newblock In \emph{NeurIPS}, 2023.

\bibitem[Li et~al.(2023)Li, Li, Savarese, and Hoi]{blip2}
Junnan Li, Dongxu Li, Silvio Savarese, and Steven Hoi.
\newblock Blip-2: Bootstrapping language-image pre-training with frozen image encoders and large language models.
\newblock In \emph{ICML}, 2023.

\bibitem[Li et~al.(2024)Li, Pan, Yang, Xu, Zhou, Zhang, Li, Kadambi, Wang, Tu, et~al.]{li20244k4dgen}
Renjie Li, Panwang Pan, Bangbang Yang, Dejia Xu, Shijie Zhou, Xuanyang Zhang, Zeming Li, Achuta Kadambi, Zhangyang Wang, Zhengzhong Tu, et~al.
\newblock 4k4dgen: Panoramic 4d generation at 4k resolution.
\newblock \emph{arXiv preprint arXiv:2406.13527}, 2024.

\bibitem[Liu et~al.(2024)Liu, Li, Chen, Li, Xu, and Plummer]{panofree}
Aoming Liu, Zhong Li, Zhang Chen, Nannan Li, Yi Xu, and Bryan~A Plummer.
\newblock Panofree: Tuning-free holistic multi-view image generation with cross-view self-guidance.
\newblock In \emph{ECCV}, 2024.

\bibitem[Liu et~al.(2023)Liu, Gong, and Liu]{rf}
Xingchao Liu, Chengyue Gong, and Qiang Liu.
\newblock Flow straight and fast: Learning to generate and transfer data with rectified flow.
\newblock In \emph{ICLR}, 2023.

\bibitem[Loshchilov and Hutter(2019)]{adamw}
Ilya Loshchilov and Frank Hutter.
\newblock Decoupled weight decay regularization.
\newblock In \emph{ICLR}, 2019.

\bibitem[Lu et~al.(2024)Lu, Hu, Wang, Bai, and Wang]{lu2024autoregressive}
Zhuqiang Lu, Kun Hu, Chaoyue Wang, Lei Bai, and Zhiyong Wang.
\newblock Autoregressive omni-aware outpainting for open-vocabulary 360-degree image generation.
\newblock In \emph{AAAI}, 2024.

\bibitem[Lugmayr et~al.(2022)Lugmayr, Danelljan, Romero, Yu, Timofte, and Van~Gool]{lugmayr2022repaint}
Andreas Lugmayr, Martin Danelljan, Andres Romero, Fisher Yu, Radu Timofte, and Luc Van~Gool.
\newblock Repaint: Inpainting using denoising diffusion probabilistic models.
\newblock In \emph{CVPR}, 2022.

\bibitem[Meng et~al.(2023)Meng, Rombach, Gao, Kingma, Ermon, Ho, and Salimans]{meng2023on}
Chenlin Meng, Robin Rombach, Ruiqi Gao, Diederik Kingma, Stefano Ermon, Jonathan Ho, and Tim Salimans.
\newblock On distillation of guided diffusion models.
\newblock In \emph{CVPR}, 2023.

\bibitem[Oh et~al.(2022)Oh, Cho, Chae, Park, Wang, and Yoon]{faed}
Changgyoon Oh, Wonjune Cho, Yujeong Chae, Daehee Park, Lin Wang, and Kuk-Jin Yoon.
\newblock Bips: Bi-modal indoor panorama synthesis via residual depth-aided adversarial learning.
\newblock In \emph{ECCV}, 2022.

\bibitem[Peebles and Xie(2023)]{dit}
William Peebles and Saining Xie.
\newblock Scalable diffusion models with transformers.
\newblock In \emph{ICCV}, 2023.

\bibitem[Podell et~al.(2024)Podell, English, Lacey, Blattmann, Dockhorn, M{\"u}ller, Penna, and Rombach]{sdxl}
Dustin Podell, Zion English, Kyle Lacey, Andreas Blattmann, Tim Dockhorn, Jonas M{\"u}ller, Joe Penna, and Robin Rombach.
\newblock Sdxl: Improving latent diffusion models for high-resolution image synthesis.
\newblock In \emph{ICLR}, 2024.

\bibitem[Poole et~al.(2023)Poole, Jain, Barron, and Mildenhall]{dreamfusion}
Ben Poole, Ajay Jain, Jonathan~T Barron, and Ben Mildenhall.
\newblock Dreamfusion: Text-to-3d using 2d diffusion.
\newblock In \emph{ICLR}, 2023.

\bibitem[Rae and Razavi(2020)]{localattn}
Jack Rae and Ali Razavi.
\newblock Do transformers need deep long-range memory?
\newblock In \emph{ACL}, 2020.

\bibitem[Ramesh et~al.(2021)Ramesh, Pavlov, Goh, Gray, Voss, Radford, Chen, and Sutskever]{dalle}
Aditya Ramesh, Mikhail Pavlov, Gabriel Goh, Scott Gray, Chelsea Voss, Alec Radford, Mark Chen, and Ilya Sutskever.
\newblock Zero-shot text-to-image generation.
\newblock In \emph{ICML}, 2021.

\bibitem[Ramesh et~al.(2022)Ramesh, Dhariwal, Nichol, Chu, and Chen]{dalle2}
Aditya Ramesh, Prafulla Dhariwal, Alex Nichol, Casey Chu, and Mark Chen.
\newblock Hierarchical text-conditional image generation with clip latents.
\newblock \emph{arXiv preprint arXiv:2204.06125}, 1\penalty0 (2):\penalty0 3, 2022.

\bibitem[Rombach et~al.(2022)Rombach, Blattmann, Lorenz, Esser, and Ommer]{ldm}
Robin Rombach, Andreas Blattmann, Dominik Lorenz, Patrick Esser, and Bj{\"o}rn Ommer.
\newblock High-resolution image synthesis with latent diffusion models.
\newblock In \emph{CVPR}, 2022.

\bibitem[Ronneberger et~al.(2015)Ronneberger, Fischer, and Brox]{unet}
Olaf Ronneberger, Philipp Fischer, and Thomas Brox.
\newblock U-net: Convolutional networks for biomedical image segmentation.
\newblock In \emph{MICCAI}, 2015.

\bibitem[Ruiz et~al.(2023)Ruiz, Li, Jampani, Pritch, Rubinstein, and Aberman]{dreambooth}
Nataniel Ruiz, Yuanzhen Li, Varun Jampani, Yael Pritch, Michael Rubinstein, and Kfir Aberman.
\newblock Dreambooth: Fine tuning text-to-image diffusion models for subject-driven generation.
\newblock In \emph{CVPR}, 2023.

\bibitem[Schuhmann et~al.(2022)Schuhmann, Beaumont, Vencu, Gordon, Wightman, Cherti, Coombes, Katta, Mullis, Wortsman, et~al.]{laion}
Christoph Schuhmann, Romain Beaumont, Richard Vencu, Cade Gordon, Ross Wightman, Mehdi Cherti, Theo Coombes, Aarush Katta, Clayton Mullis, Mitchell Wortsman, et~al.
\newblock Laion-5b: An open large-scale dataset for training next generation image-text models.
\newblock In \emph{NeurIPS}, 2022.

\bibitem[Shazeer et~al.(2017)Shazeer, Mirhoseini, Maziarz, Davis, Le, Hinton, and Dean]{moe}
Noam Shazeer, Azalia Mirhoseini, Krzysztof Maziarz, Andy Davis, Quoc Le, Geoffrey Hinton, and Jeff Dean.
\newblock Outrageously large neural networks: The sparsely-gated mixture-of-experts layer.
\newblock In \emph{ICLR}, 2017.

\bibitem[Song et~al.(2020)Song, Meng, and Ermon]{ddim}
Jiaming Song, Chenlin Meng, and Stefano Ermon.
\newblock Denoising diffusion implicit models.
\newblock In \emph{ICLR}, 2020.

\bibitem[Song et~al.(2021)Song, Sohl-Dickstein, Kingma, Kumar, Ermon, and Poole]{scoresde}
Yang Song, Jascha Sohl-Dickstein, Diederik~P Kingma, Abhishek Kumar, Stefano Ermon, and Ben Poole.
\newblock Score-based generative modeling through stochastic differential equations.
\newblock In \emph{ICLR}, 2021.

\bibitem[Song et~al.(2023)Song, Dhariwal, Chen, and Sutskever]{cm}
Yang Song, Prafulla Dhariwal, Mark Chen, and Ilya Sutskever.
\newblock Consistency models.
\newblock In \emph{ICML}, 2023.

\bibitem[Szegedy et~al.(2016)Szegedy, Vanhoucke, Ioffe, Shlens, and Wojna]{inceptionnet}
Christian Szegedy, Vincent Vanhoucke, Sergey Ioffe, Jon Shlens, and Zbigniew Wojna.
\newblock Rethinking the inception architecture for computer vision.
\newblock In \emph{CVPR}, 2016.

\bibitem[Tang et~al.(2023)Tang, Zhang, Chen, Wang, and Furukawa]{mvdiff}
Shitao Tang, Fuyang Zhang, Jiacheng Chen, Peng Wang, and Yasutaka Furukawa.
\newblock Mvdiffusion: Enabling holistic multi-view image generation with correspondence-aware diffusion.
\newblock In \emph{NeurIPS}, 2023.

\bibitem[von Platen et~al.(2022)von Platen, Patil, Lozhkov, Cuenca, Lambert, Rasul, Davaadorj, Nair, Paul, Berman, Xu, Liu, and Wolf]{diffusers}
Patrick von Platen, Suraj Patil, Anton Lozhkov, Pedro Cuenca, Nathan Lambert, Kashif Rasul, Mishig Davaadorj, Dhruv Nair, Sayak Paul, William Berman, Yiyi Xu, Steven Liu, and Thomas Wolf.
\newblock Diffusers: State-of-the-art diffusion models.
\newblock \url{https://github.com/huggingface/diffusers}, 2022.

\bibitem[Wang et~al.(2022)Wang, Yang, Loy, and Liu]{wang2022stylelight}
Guangcong Wang, Yinuo Yang, Chen~Change Loy, and Ziwei Liu.
\newblock Stylelight: Hdr panorama generation for lighting estimation and editing.
\newblock In \emph{ECCV}, 2022.

\bibitem[Wang et~al.(2024)Wang, Xiang, Fan, and Xue]{stitchdiffusion}
Hai Wang, Xiaoyu Xiang, Yuchen Fan, and Jing-Hao Xue.
\newblock Customizing 360-degree panoramas through text-to-image diffusion models.
\newblock In \emph{WACV}, 2024.

\bibitem[Wu et~al.(2024{\natexlab{a}})Wu, Zheng, and Cham]{panodiffusion}
Tianhao Wu, Chuanxia Zheng, and Tat-Jen Cham.
\newblock Panodiffusion: 360-degree panorama outpainting via diffusion.
\newblock In \emph{ICLR}, 2024{\natexlab{a}}.

\bibitem[Wu et~al.(2024{\natexlab{b}})Wu, Huang, and Wei]{mole}
Xun Wu, Shaohan Huang, and Furu Wei.
\newblock Mixture of lora experts.
\newblock In \emph{ICLR}, 2024{\natexlab{b}}.

\bibitem[Xia et~al.(2022)Xia, Pan, Song, Li, and Huang]{deformableattn}
Zhuofan Xia, Xuran Pan, Shiji Song, Li~Erran Li, and Gao Huang.
\newblock Vision transformer with deformable attention.
\newblock In \emph{CVPR}, 2022.

\bibitem[Xie et~al.(2023)Xie, Zhang, Lin, Hinz, and Zhang]{xie2023smartbrush}
Shaoan Xie, Zhifei Zhang, Zhe Lin, Tobias Hinz, and Kun Zhang.
\newblock Smartbrush: Text and shape guided object inpainting with diffusion model.
\newblock In \emph{CVPR}, 2023.

\bibitem[Xu et~al.(2023)Xu, Liu, Vahdat, Byeon, Wang, and De~Mello]{xu2023open}
Jiarui Xu, Sifei Liu, Arash Vahdat, Wonmin Byeon, Xiaolong Wang, and Shalini De~Mello.
\newblock Open-vocabulary panoptic segmentation with text-to-image diffusion models.
\newblock In \emph{CVPR}, 2023.

\bibitem[Yang et~al.(2024)Yang, Dong, Ma, Hu, Liu, Cui, and Ma]{yang2023dreamspace}
Bangbang Yang, Wenqi Dong, Lin Ma, Wenbo Hu, Xiao Liu, Zhaopeng Cui, and Yuewen Ma.
\newblock Dreamspace: Dreaming your room space with text-driven panoramic texture propagation.
\newblock In \emph{IEEE VR}, 2024.

\bibitem[Ye et~al.(2024)Ye, Ji, Chen, Gao, Huang, Zhang, Ouyang, He, Zhao, and Zhang]{diffpano}
Weicai Ye, Chenhao Ji, Zheng Chen, Junyao Gao, Xiaoshui Huang, Song-Hai Zhang, Wanli Ouyang, Tong He, Cairong Zhao, and Guofeng Zhang.
\newblock Diffpano: Scalable and consistent text to panorama generation with spherical epipolar-aware diffusion.
\newblock In \emph{NeurIPS}, 2024.

\bibitem[Yu et~al.(2023)Yu, Forghani, Derpanis, and Brubaker]{yu2023long}
Jason~J Yu, Fereshteh Forghani, Konstantinos~G Derpanis, and Marcus~A Brubaker.
\newblock Long-term photometric consistent novel view synthesis with diffusion models.
\newblock In \emph{ICCV}, 2023.

\bibitem[Zhang et~al.(2024)Zhang, Wu, Gambardella, Huang, Phung, Ouyang, and Cai]{panfusion}
Cheng Zhang, Qianyi Wu, Camilo~Cruz Gambardella, Xiaoshui Huang, Dinh Phung, Wanli Ouyang, and Jianfei Cai.
\newblock Taming stable diffusion for text to 360° panorama image generation.
\newblock In \emph{CVPR}, 2024.

\bibitem[Zhang et~al.(2023)Zhang, Rao, and Agrawala]{controlnet}
Lvmin Zhang, Anyi Rao, and Maneesh Agrawala.
\newblock Adding conditional control to text-to-image diffusion models.
\newblock In \emph{ICCV}, 2023.

\bibitem[Zhao et~al.(2023)Zhao, Bai, Rao, Zhou, and Lu]{unipc}
Wenliang Zhao, Lujia Bai, Yongming Rao, Jie Zhou, and Jiwen Lu.
\newblock Unipc: A unified predictor-corrector framework for fast sampling of diffusion models.
\newblock In \emph{NeurIPS}, 2023.

\bibitem[Zheng et~al.(2023)Zheng, Lu, Chen, and Zhu]{dpmv3}
Kaiwen Zheng, Cheng Lu, Jianfei Chen, and Jun Zhu.
\newblock Dpm-solver-v3: Improved diffusion ode solver with empirical model statistics.
\newblock In \emph{NeurIPS}, 2023.

\end{thebibliography}
